\documentclass[10pt,journal,compsoc]{IEEEtran}
\usepackage[colorlinks=true,linkcolor=magenta,citecolor=green,urlcolor=magenta,]{hyperref}
\usepackage{url}
\usepackage{ragged2e}
\usepackage{makecell}
\usepackage{amsfonts}
\usepackage{utfsym}
\usepackage{amsmath}
\usepackage{bbding}
\usepackage{amsthm}
\usepackage{graphicx}
\usepackage{tikz}
\usepackage[edges]{forest}
\usepackage{tabularx}
\usepackage{array}
\usepackage{longtable}
\usepackage{ltxtable}
\definecolor{hidden-draw}{RGB}{20,68,106}
\definecolor{hidden-pink}{RGB}{255,245,247}
\usepackage{amsmath}
\usepackage{setspace}
\usepackage{cleveref} 
\usepackage{makecell}
\usepackage{colortbl}

\usepackage{subfig}
\usepackage{multirow}
\usepackage{booktabs}
\usepackage{stfloats}
\usepackage{color}
\usepackage{colortbl}
\usepackage{xcolor}
\usepackage{cleveref}
\usepackage{amssymb}
\usepackage{pifont}
\usepackage{amsmath}
\usepackage[ruled, vlined, linesnumbered]{algorithm2e}
\SetKwInput{KwInput}{Input}
\SetKwInput{KwOutput}{Output}
\SetKwInput{KwRequire}{Require}
\SetKwInput{KwModel}{Model}
\SetKwInput{KwInit}{Initialization} 

\definecolor{drp-blue}{HTML}{1f77b4}
\definecolor{pretty-blue}{RGB}{0, 113, 188}
\definecolor{kaiming-green}{RGB}{57,181,74} 
\definecolor{mypurple}{RGB}{55,0,168} 
\definecolor{icmlblue}{rgb}{0,0.08,0.45} 
\definecolor{mygreen}{HTML}{4FC978}
\definecolor{linecolor1}{RGB}{246, 248, 239}
\definecolor{linecolor2}{RGB}{230, 234, 217}
\definecolor{linecolor3}{RGB}{211, 222, 190}
\definecolor{line-blue}{RGB}{243, 248, 252}
\definecolor{reconcolor}{HTML}{412F8A}
\definecolor{runpei-orange}{HTML}{F35F27}
\definecolor{runpei_blue}{HTML}{14294B}
\definecolor{datacolor}{HTML}{0009BF}
\definecolor{vitcolor}{HTML}{fc8e62}
\definecolor{cvprblue}{rgb}{0.21,0.49,0.74}
\definecolor{myblue}{rgb}{.39,.58,.93}
\definecolor{my_green}{RGB}{51,102,0}
\definecolor{my_red}{RGB}{204, 0, 0}
\definecolor{my_orange}{RGB}{240, 188, 124}
\definecolor{my_light_blue}{RGB}{181, 212, 216}

\def\eg{\emph{e.g., }}

\usepackage[utf8]{inputenc} 
\usepackage[T1]{fontenc}    
\usepackage{amsfonts}       
\usepackage{nicefrac}       
\usepackage{xcolor}         
\usepackage{graphicx}
\usepackage{multirow}
\usepackage{tikz}
\usepackage[edges]{forest}
\definecolor{hidden-draw}{RGB}{20,68,106}
\definecolor{hidden-pink}{RGB}{255,245,247}
\usepackage{amsmath}
\usepackage{setspace}
\usepackage{cleveref} 
\usepackage{makecell}
\usepackage{colortbl}

\usepackage{booktabs}
\usepackage{array}
\usepackage{hyperref}
\usepackage{graphicx}

\usepackage[nocompress]{cite}

\definecolor{drp-blue}{HTML}{1f77b4}
\definecolor{pretty-blue}{RGB}{0, 113, 188}
\definecolor{kaiming-green}{RGB}{57,181,74} 
\definecolor{mypurple}{RGB}{55,0,168} 
\definecolor{icmlblue}{rgb}{0,0.08,0.45} 
\definecolor{mygreen}{HTML}{4FC978}
\definecolor{linecolor1}{RGB}{246, 248, 239}
\definecolor{linecolor2}{RGB}{230, 234, 217}
\definecolor{linecolor3}{RGB}{211, 222, 190}
\definecolor{line-blue}{RGB}{243, 248, 252}
\definecolor{reconcolor}{HTML}{412F8A}
\definecolor{runpei-orange}{HTML}{F35F27}
\definecolor{runpei_blue}{HTML}{14294B}
\definecolor{datacolor}{HTML}{0009BF}
\definecolor{vitcolor}{HTML}{fc8e62}
\definecolor{cvprblue}{rgb}{0.21,0.49,0.74}
\definecolor{myblue}{rgb}{.39,.58,.93}
\crefname{table}{Tab.}{Tab.}
\crefname{figure}{Fig.}{Fig.}
\crefname{equation}{Eq.}{Eq.}

\ifCLASSINFOpdf
\else
\fi
\begin{document}
%
\title{A Survey on Efficient Vision-Language-Action Models}

%
%
%

\author{Zhaoshu~Yu,
        Bo~Wang,
        Pengpeng~Zeng,
        Haonan~Zhang,
        Ji~Zhang,
        Zheng~Wang, \\
        Lianli~Gao,
        Jingkuan~Song,
        Nicu~Sebe,
        and Heng~Tao~Shen, \textit{IEEE Fellow}
\thanks{This study is supported by grant from the Fundamental and Interdisciplinary Disciplines Breakthrough Plan of the Ministry of Education of China (Grant No. JYB2025XDXM116).}
\IEEEcompsocitemizethanks
{\IEEEcompsocthanksitem Zhaoshu Yu, Bo Wang, Pengpeng Zeng, Haonan Zhang, Zheng~Wang, and Heng Tao Shen are with the School of Computer Science and Technology, Tongji University, China.
\IEEEcompsocthanksitem Ji Zhang is with the School of Computing and Artificial Intelligence, Southwest Jiaotong University, China.
\IEEEcompsocthanksitem Lianli Gao is with the School of Computer Science and Engineering, University of Electronic Science and Technology of China, China. 
\IEEEcompsocthanksitem Jingkuan Song is with the School of Computer Science and Technology, Tongji University, China, and Shanghai Innovation Institute, China.
\IEEEcompsocthanksitem Nicu Sebe is with the Department of Information Engineering and Computer Science, University of Trento, Italy.
\IEEEcompsocthanksitem Corresponding author: Pengpeng Zeng. (is.pengpengzeng@gmail.com) 
}
}

%
%

\markboth{JOURNAL OF \LaTeX\ CLASS FILES,~VOL.~14, NO.~8, AUGUST~2021}%
{Shell \MakeLowercase{\textit{et al.}}: A Sample Article Using IEEEtran.cls for IEEE Journals}

\IEEEtitleabstractindextext{%
\begin{abstract}
  \justifying
Vision-Language-Action models (VLAs) represent a significant frontier in embodied intelligence, aiming to bridge digital knowledge with physical-world interaction. Despite their remarkable performance, foundational VLAs are hindered by the prohibitive computational and data demands inherent to their large-scale architectures. 
To this end, recent studies improve VLA efficiency from different views, \eg real-time inference, training computation, and scalable data collection. However, these efforts are mostly studied separately. A unified view is still missing for understanding how efficiency should be optimized across the full VLA lifecycle.
To bridge this gap, this survey presents the first comprehensive review of Efficient Vision-Language-Action models (\textbf{Efficient VLAs}) across the entire model-training-data pipeline.
Specifically, we introduce a unified taxonomy to systematically organize the disparate efforts in this domain, categorizing current techniques into three core pillars: \textit{(1) Efficient Model Design}, focusing on efficient architectures and model compression; \textit{(2) Efficient Training}, which reduces computational burdens during model learning; and \textit{(3) Efficient Data Collection}, which addresses the bottlenecks in acquiring and utilizing robotic data. Through a critical review of state-of-the-art methods within this framework, this survey provides an organized reference for the community and summarizes representative applications, delineates key challenges, and charts a roadmap for future research. We maintain a continuously updated project page to track our latest developments: \url{https://evla-survey.github.io/}.

\end{abstract}

}

\maketitle

\IEEEdisplaynontitleabstractindextext

%
\IEEEpeerreviewmaketitle

\IEEEraisesectionheading{\section{Introduction}\label{sec:introduction}}


\IEEEPARstart{V}{ision-Language-Action} Models (VLAs)~\cite{kim2024openvla,black2024pi_0,zitkovich2023rt,li2024cogact} signify a breakthrough in the evolution of artificial intelligence, endowing interactive physical entities with the capacity for perception and decision-making. These models ground the semantic reasoning of Large Language Models (LLMs) and the cross-modal perception of Vision-Language Models (VLMs) into action spaces and show strong dexterity and generalization in complex environments. The potential of this paradigm is already manifest across diverse high-stakes domains, ranging from the real-time interaction in autonomous driving~\cite{cao2025fastdrivevla, zhou2025opendrivevla} and industrial manufacturing~\cite{margadji2025hybrid} to the high-precision requirements of medical robotics~\cite{li2024robonurse} and laboratory automation~\cite{zhang2025robochemist}.

\begin{figure}[]
  \centering
\includegraphics[width=\linewidth]{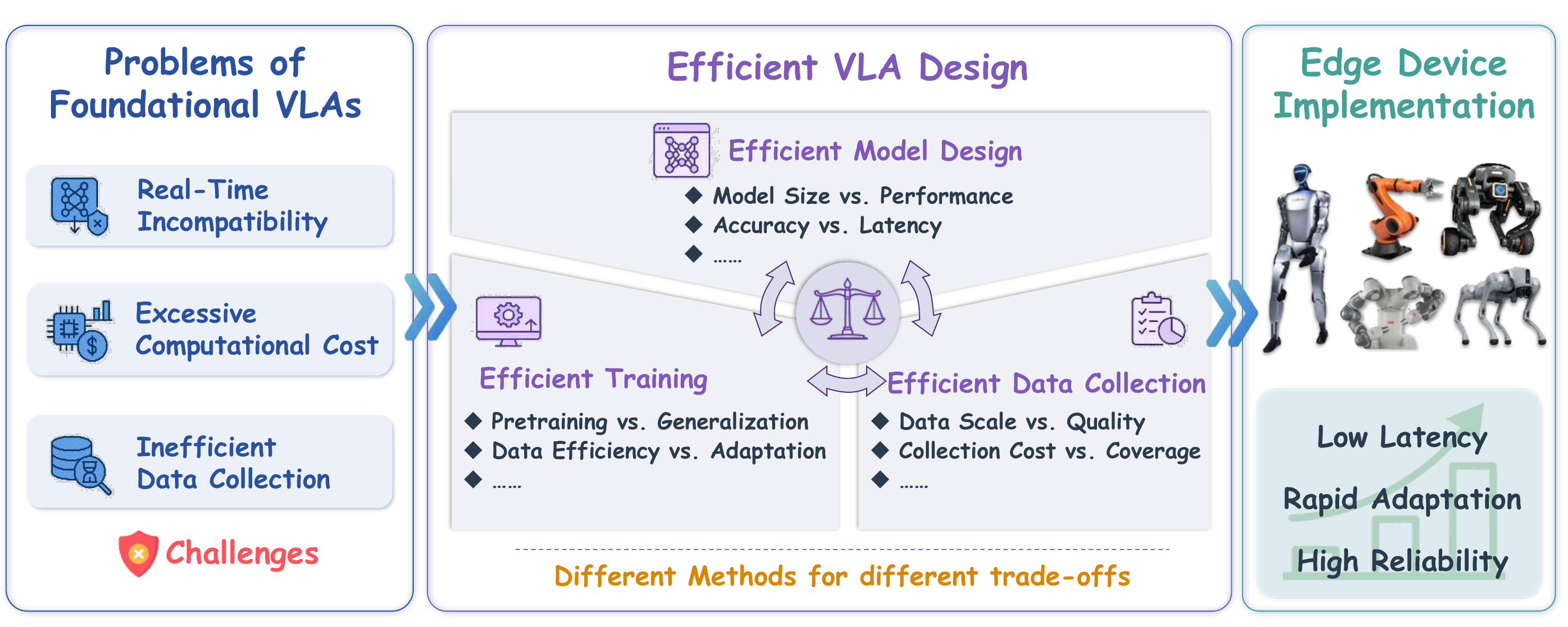}
\caption{\textbf{The transition from foundational VLAs to efficient VLAs.} Foundational VLAs are bottlenecked by real-time incompatibility, excessive computational costs, and inefficient data collection. To enable deployment on edge devices, Efficient VLAs navigate a design space characterized by diverse trade-offs across three core pillars. By strategically balancing these competing objectives, different methods achieve low latency, rapid adaptation, and high reliability across diverse edge device implementations.}
  \label{fig: roadmap towards Efficient VLAs}
\end{figure}

Recent VLA research is organized around three technical axes: architecture design, training procedures, and data curation. Architectural research focuses on coupling multimodal backbones~\cite{beyer2024paligemma, karamcheti2024prismatic} with diverse action decoders to facilitate end-to-end mapping from perception to execution, spanning autoregressive policies~\cite{kim2024openvla} for sequential prediction to diffusion-based generators~\cite{black2024pi_0} for trajectory synthesis. Training methodologies have evolved toward large-scale cross-embodiment pre-training on heterogeneous datasets~\cite{bjorck2025gr00t}, progressively integrating reinforcement learning~\cite{li2025simplevla} and supervised fine-tuning~\cite{kim2025fine} to improve downstream adaptation. Meanwhile, data-centric strategies prioritize high-fidelity trajectories through expert demonstrations~\cite{bu2025agibot}, realistic simulation~\cite{chen2025robotwin}, and generative augmentation~\cite{li2025llara} to improve transfer across diverse tasks. These multi-dimensional advancements constitute a cohesive research ecosystem that continually pushes the frontiers of autonomous agency in the physical world.




Although these models show strong generalist behavior, foundational VLAs inherit large computational and data footprints from scaling-centric LLM and VLM backbones, which often conflict with latency and power budgets on physical platforms. This mismatch complicates real-time edge deployment on robotic manipulators and mobile agents, and appears through three bottlenecks:
(1) \textbf{Real-Time Incompatibility}: current VLAs suffer from high inference latency and insufficient control frequency~\cite{kim2024openvla}, making them incompatible with the sub-second control cycles required for responsive and adaptive robotic manipulation; 
(2) \textbf{Excessive Computational Cost}: the pre-training requirements are prohibitively large, as exemplified by OpenVLA~\cite{kim2024openvla} consuming 21,500 A100-GPU hours on a 64-GPU cluster, posing significant barriers to reproducibility and scalability; and 
(3) \textbf{Inefficient Data Collection}: The reliance on large-scale datasets, with examples like $\pi_0$~\cite{black2024pi_0} requiring over 10,000 hours of robotic trajectories, results in a time-consuming and labor-intensive data collection pipeline that severely limits the applicability of such approaches. 

To address these challenges, we introduce the \textbf{Efficient VLA} paradigm. 
An efficient VLA is not merely a compact model, but a resource-aware embodied system jointly optimized across model design, training, and data collection. 
It aims to reduce resource cost and inference latency while preserving real-world embodied capability.
As illustrated in~\cref{fig: roadmap towards Efficient VLAs}, rather than isolated optimizations, these designs~\cite{kim2025fine, zhang2025mole, liu2024robomamba, shukor2025smolvla, pertsch2025fast, yang2025efficientvla} navigate a co-dependent space of trade-offs across the three pillars. By forming a synergistic, closed-loop ecosystem, they narrow the gap between heavyweight foundations and edge deployment, targeting high-frequency control under constrained robotic platforms.

\begin{figure*}
  \centering
  \includegraphics[width=\linewidth]{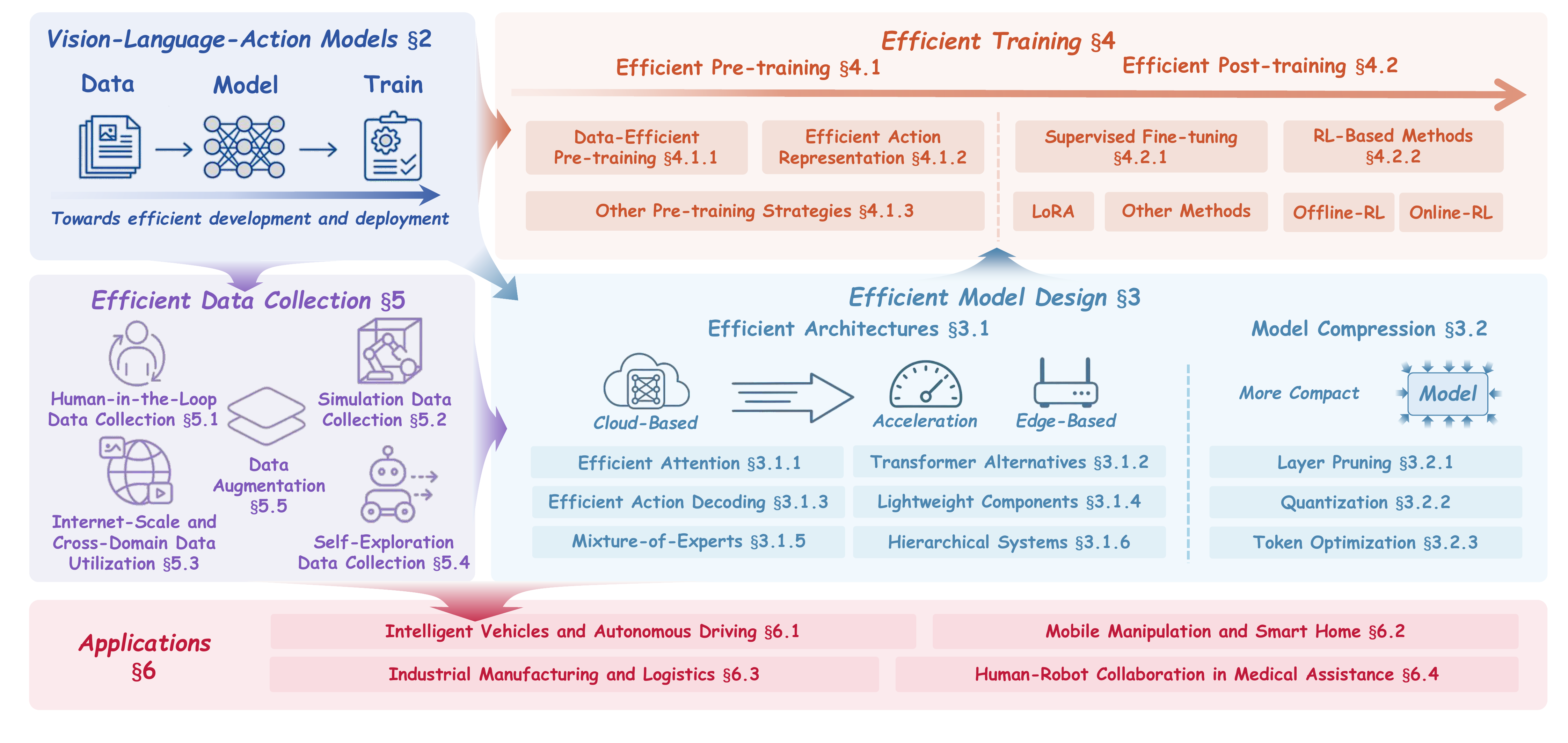}
  \caption{\textbf{The organization of our survey.} We systematically categorize efficient VLAs into three core pillars: (1) \textbf{Efficient Model Design}, encompassing efficient architectures and model compression techniques, (2) \textbf{Efficient Training}, covering efficient pre-training and post-training strategies, and (3) \textbf{Efficient Data Collection}, including efficient data collection and augmentation methods. The framework also reviews VLA foundations and key applications.
  }
  \label{fig:TOC}
\end{figure*}

The overarching goal of this survey is to provide a comprehensive view of the technological advances in efficient VLAs.
As illustrated in~\cref{fig:TOC}, we organize the literature in a taxonomy comprising three main categories: 
(1) \textbf{Efficient Model Design}: encompassing innovative strategies to optimize the architectural and inference efficiency of VLAs. This includes \textit{Efficient Architectures}, which offers innovative structural optimizations alongside scalable frameworks, and \textit{Model Compression}, which involves techniques to compress model parameters and refine token handling. 
(2) \textbf{Efficient Training}:  incorporating advanced approaches to reduce the computational and data burdens during VLAs training, containing \textit{Efficient Pre-training/Post-training}. The former improves efficiency through optimized data and action modeling, while the latter enhances adaptability and performance via customized learning strategies, \eg supervised fine-tuning and reinforcement learning.  
(3) \textbf{Efficient Data Collection}: focusing on cutting-edge methods to improve data collection and enrichment for VLAs development, using interactive, simulated, reusable, and self-driven strategies to make dataset acquisition more scalable and efficient.
Through this exploration, we provide a comprehensive understanding of the current state of efficient VLAs, thereby highlighting key challenges and advancements in embodied intelligence and robotic learning.




While existing literature provides extensive reviews of the general VLA landscape, ranging from architectural overviews~\cite{ma2024survey, din2025vision, zhang2025pure} and VLM-driven advancements~\cite{shao2025large} to specific mechanisms such as action tokenization~\cite{zhong2025survey} and motor learning~\cite{xiang2025parallels}, these surveys primarily focus on conventional, performance-oriented paradigms. To address their fragmented treatment of efficiency and the lack of a unified taxonomy, this paper presents the first systematic review of efficient VLAs, synthesizing model design, training, and data collection strategies to advance scalable embodied AI. Through this comprehensive synthesis, we bridge the gap between theoretical development and real-world deployment in resource-constrained scenarios. Accompanied by a dedicated GitHub repository, this survey is designed to serve as both a valuable reference for navigating the literature and a foundation for inspiring future innovations.



\IEEEpubidadjcol

The primary contributions of this survey are summarized as follows:
\begin{itemize}
  \item \textbf{Pioneering Survey:} To the best of our knowledge, this work presents the first comprehensive survey specifically dedicated to the realm of efficient VLAs that covers the entire ``model-training-data'' process. It fills a critical gap in the literature and aims to establish an organized reference for the research community.
  \item \textbf{Novel Taxonomy:} We introduce a novel and systematically structured taxonomy that organizes the core technical landscape for building efficient VLAs into three interconnected pillars: Efficient Model Design, Efficient Training, and Efficient Data Collection.
  \item \textbf{Future Roadmap:} We critically distill the key challenges and current limitations plaguing the field, thereby outlining promising and forward-looking research directions to inspire and guide future endeavors in scalable embodied intelligence.
  
\end{itemize}


The remaining sections of this survey are organized as illustrated in~\cref{fig:TOC}. We begin in Sec.~\ref{sec:vision language action models} by introducing the foundational concepts of VLAs, delineating the computational challenges that necessitate the pursuit of efficiency. Sec.~\ref{sec:efficientmodeldesign} delves into efficient model design, systematically categorizing techniques that span efficient architectural innovations and model compression.  Sec.~\ref{sec:efficienttraining} investigates efficient training paradigms, encompassing strategies to reduce the computational overhead and dataset demands during both pre-training and post-training stages. Sec.~\ref{sec:efficientdatacollection} examines the crucial role of efficient data collection, exploring methodologies for scalable data collection and effective augmentation to maximize data utility. We then survey key applications in Sec.~\ref{sec:applications} where efficient VLAs have demonstrated significant potential, highlighting their practical utility in real-world scenarios. Finally, Sec.~\ref{sec:conclusion} summarizes the survey.

\section{Vision-Language-Action Models}
\label{sec:vision language action models}


In this section, we first review the basic VLA pipeline and identify the key efficiency bottlenecks behind resource-aware VLA design. 
We then present the taxonomy principle and highlight the unique positioning of this survey as the first systematic review of efficient VLAs in embodied AI.


\subsection{An Overview of VLAs}  


\begin{figure}
  \centering
  \includegraphics[width=\linewidth]{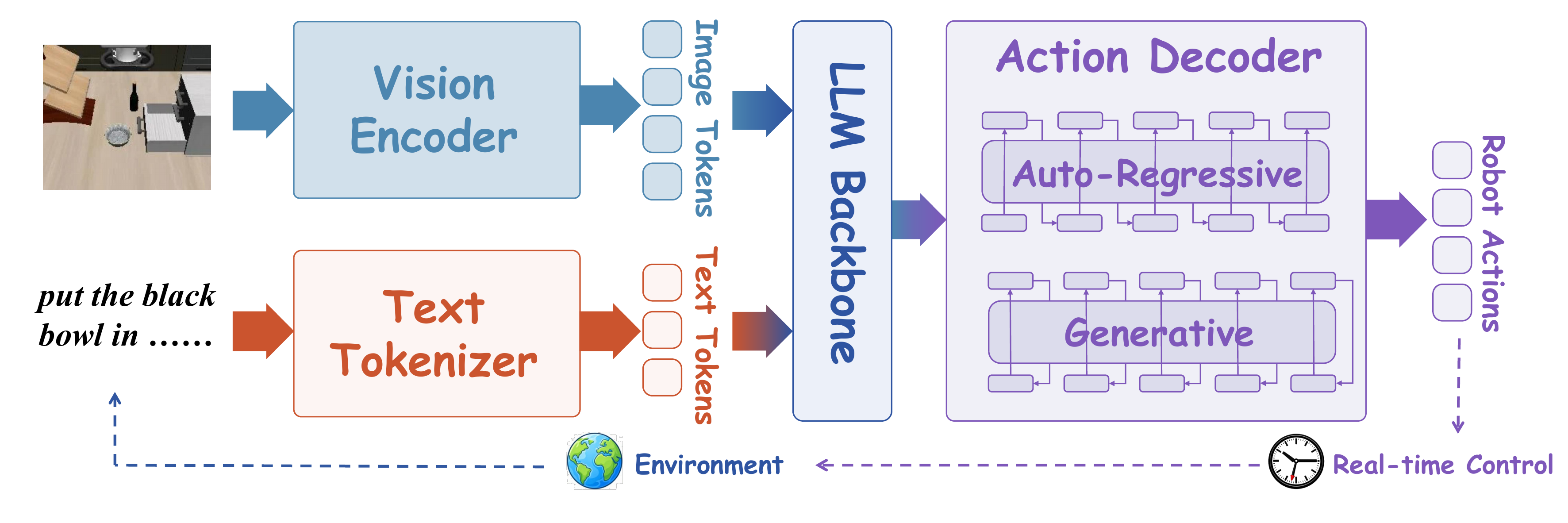}
  \caption{\textbf{An overview of VLAs.} VLAs integrate vision encoders to extract visual features, LLM backbones to fuse multimodal inputs, and action decoders (autoregressive or generative) to produce robotic control signals, enabling end-to-end vision-language-action reasoning for embodied manipulation tasks.}
  \label{fig:overview of vlas}
\end{figure}

\subsubsection{Foundational Pipeline}
As depicted in~\cref{fig:overview of vlas}, VLAs' foundational pipeline divides multimodal inference into three synergistic modules: a vision encoder that processes scene images into patch embeddings, a LLM backbone augmented for vision-language fusion that orchestrates high-level reasoning, and an action decoder that yields precise control trajectories. Pre-trained components allow these pipelines to inherit visual-language priors before adaptation to embodied action generation.

\noindent\textbf{Vision Encoder.}
At the ingress, the vision encoder $E_{img}(\cdot)$ ingests RGB observations and extracts their hierarchical features. Prevailing choices include vision transformers like ViT~\cite{dosovitskiy2020image}, SigLIP~\cite{zhai2023sigmoid}, DINOv2~\cite{oquab2023dinov2}, and CLIP~\cite{radford2021learning}, whose contrastive or self-supervised pre-training on vast corpora imparts zero-shot versatility. Abstractly, this module maps raw imagery to semantic tokens:
\begin{equation}
    \begin{aligned}
    \mathbf{v} = E_{img}(I;\theta_{img}),
    \end{aligned}
    \label{eq1}
\end{equation}
where $I \in \mathbb{R}^{H \times W \times 3}$ denotes the input images, $\mathbf{v} \in \mathbb{R}^{N_v \times D_v}$ represents the vision tokens, and $\theta_{img}$ indicates the encoder parameters.

\noindent\textbf{LLM Backbone.}
After a projector $P(\cdot)$ that ensures seamless alignment and mitigates modality gaps, the multimodal sequence then feeds into the LLM backbone $LLM(\cdot)$—the center for semantic reasoning. Spanning Qwen~\cite{bai2023qwen}, PaLM~\cite{chowdhery2023palm}, Gemma~\cite{team2024gemma}, Llama~\cite{touvron2023llama}, and Mamba~\cite{gu2024mamba}, pre-trained LLMs process the fused embeddings for task planning, which can be formulated as follows:
\begin{equation}
    \begin{aligned}
    \mathbf{h} = LLM(P(\mathbf{v},\mathbf{l});\theta_{LLM}),
    \end{aligned}
    \label{eq2}
\end{equation}
where $\mathbf{l} \in \mathbb{R}^{N_l \times D_l}$ is language tokens, $\mathbf{h} \in \mathbb{R}^{N_h \times D}$ denotes the hidden states and $\theta_{LLM}$ represents the LLM parameters.

\noindent\textbf{Action Decoder.}
Culminating the flow, the action decoder transmutes latents into robot-ready outputs, such as end-effector poses and gripper commands. Common instantiations include diffusion/flow matching~\cite{ho2020denoising, peebles2023scalable, lipman2022flow} for stochastic refinement of trajectories and autoregressive decoding for sequential action prediction. They all generate action sequences conditioned on reasoning:
\begin{equation}
    \begin{aligned}
    \mathbf{a}_{1:T} = D_{act}(\mathbf{h};\theta_{act}),
    \end{aligned}
    \label{eq3}
\end{equation}
where $\mathbf{a}_{1:T} \in \mathbb{R}^{T \times D_a}$ denotes the action chunk over $T$ timesteps and $\theta_{act}$ indicates the decoder parameters.

Overall, this tripartite architecture couples perception, language-conditioned reasoning, and action prediction within an end-to-end policy pipeline.





\begin{figure*}[!htbp]
  \centering
  \includegraphics[width=\textwidth]{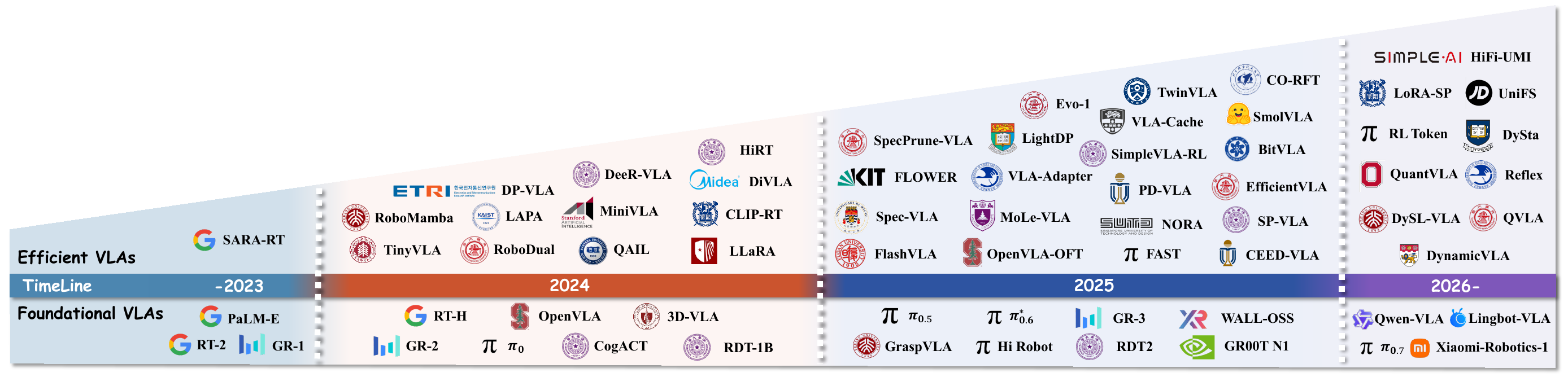}
  \caption{\textbf{Timeline of foundational and efficient VLAs.} This timeline charts the progression toward resource-conscious embodied intelligence. Inclusion in the ``Efficient VLA'' trajectory strictly requires explicit contributions addressing foundational bottlenecks via at least one of our core taxonomy pillars: mitigating computational footprint and inference latency (Efficient Model Design), minimizing adaptation overhead (Efficient Training), or reducing data acquisition costs (Efficient Data Collection).}
  \label{fig:timelineofvlafoundationmodelsandefficientvlas}
\end{figure*}

\subsubsection{Training Recipes}

Following the practice of LLMs and VLMs, VLAs are typically trained in two stages: pre-training and post-training. In this survey, we distinguish these stages by their functions, since they are often mixed in existing literature.
We define \textbf{Pre-training} as the stage that gives a VLM the basic capacity to act. It moves a general-purpose vision-language model into the embodied domain by aligning visual-language understanding with robot-executable actions. \textbf{Post-training}, in contrast, starts from this generalist policy and adapts it to specific tasks, environments, or resource budgets.
For instance, RT-2~\cite{zitkovich2023rt} integrates web-scale vision-language data with robotic trajectories, while OpenVLA~\cite{kim2024openvla} and $\pi_0$~\cite{black2024pi_0} leverage the massive Open X-Embodiment dataset~\cite{o2024open} to establish generalized behavioral priors, which are subsequently specialized through task-specific post-training to align these foundations with environmental constraints for manipulative dexterity.

Beyond these data-intensive pre-training paradigms, certain emerging works explore data-efficient~\cite{fan2025diffusion,ye2025latent} adaptation or parameter-efficient fine-tuning~\cite{zhang2025hirt, wen2025tinyvla} to swiftly bridge the gap between frozen VLM backbones and robotic agency. Consequently, this survey encompasses a diverse methodological spectrum within the pre-training stage in Sec.~\ref{subsec:efficientpretraining}, including techniques traditionally regarded as ``post-training'' in other domains~\cite{hu2022lora, luo2024moil, yu2025moe, zhang2024vision, xu2026parameter, zhang2026survey}. The key distinction is not the method itself, but its objective: \textit{if the goal is to create the first version of the action-capable model from a VLM, we classify it as pre-training.} This approach offers a more fundamental and logically consistent framework that aligns with the practical development lifecycle of VLAs.

\subsubsection{Data Collection Paradigms}
VLA data mainly comes from two sources: \textbf{real-world data} and \textbf{simulation data}. 
Real-world data is collected on physical robots through teleoperation. Representative datasets such as Open X-Embodiment~\cite{o2024open}, BridgeData V2~\cite{walke2023bridgedata}, and DROID~\cite{khazatsky2024droid} further standardize heterogeneous embodiments, tasks, and camera setups. In parallel, simulation data is generated in virtual environments by instantiating tasks, objects, and robot embodiments in physics engines. Benchmarks such as RLBench~\cite{james2020rlbench}, RoboCasa~\cite{nasiriany2024robocasa}, RoboGen~\cite{wang2024robogen}, and LIBERO~\cite{liu2023libero} are simulation data. Together, these two sources constitute the basic training substrate for modern VLAs.

\subsection{The Need for Efficient VLAs}

Despite the formidable capabilities demonstrated by VLA foundational models, their adoption beyond research prototypes faces significant barriers due to profound inefficiencies. These challenges stem from the "triple-bottleneck" across the model-training-data lifecycle.

\begin{table}[t]  
\scriptsize
\renewcommand{\arraystretch}{1.1} 
\newcolumntype{C}[1]{>{\centering\arraybackslash}p{#1}} 
\newcolumntype{L}[1]{>{\raggedright\arraybackslash}p{#1}} 
\caption{\textbf{Efficiency-related metrics of representative VLAs.} The table reports the number of parameters (\textbf{Params.}), inference latency (\textbf{I.L.}), operating frequency (\textbf{Freq.}), and accelerator (\textbf{Accelerator}) of representative VLAs to contextualize cross-paper efficiency comparisons. ``$\downarrow$'' indicates that lower values are better, and ``$\uparrow$'' indicates that higher values are better. ``/'' denotes that the model can operate at two different frequencies.}
\label{table:efficiency_metrics}
\resizebox{\linewidth}{!}{
\begin{tabular}{C{1.7cm} C{1.0cm} C{1.3cm} C{1.5cm} C{1.9cm}}
    \toprule
    \textbf{Models} &
    \textbf{Params.}\textbf{($\downarrow$)} &
    \textbf{I.L.}\textbf{(ms) ($\downarrow$)} &
    \textbf{Freq.}\textbf{(Hz) ($\uparrow$)} &
    \textbf{Accelerator} \\
    \midrule
    RT-2-PaLI-X~\cite{zitkovich2023rt} & 55B  & 330-1000  & 1-3  & Multi-TPU cloud \\
    RT-2-PaLI-X~\cite{zitkovich2023rt} & 5B  & 200 & 5 & Multi-TPU cloud \\
    OpenVLA~\cite{kim2024openvla} & 7B  & 166 & 6 & RTX 4090 \\
    $\pi_0$~\cite{black2024pi_0} & 3.3B & 73  & 20/50 & RTX 4090 \\
    Hi Robot~\cite{shi2025hi} & 3B   & 73 & 10/50 & 1--2 RTX 4090 \\
    GR00T N1~\cite{bjorck2025gr00t} & 2.2B   & 63.9 & - & L40 GPU \\
    
    \bottomrule
    \end{tabular}
}
\end{table}

\noindent\textbf{Architectural Incompatibility.}
From an architectural perspective, the primary impediment resides in the massive computational footprint of transformer-based VLM backbones. Characterized by billions of parameters~\cite{beyer2024paligemma, karamcheti2024prismatic, driess2023palm, chen2023pali}, these models incur prohibitive inference latency, resulting in control frequencies that fail to meet the stringent real-time requirements of dynamic robotic actuation, as evidenced in~\cref{table:efficiency_metrics}. 
This inefficiency is further amplified by visual signals~\cite{xu2025vla,wang2025specprune}, which produce dense token sequences that must be processed through deep layers.

\noindent\textbf{Training Overhead.}
The current training paradigm is burdened by the exorbitant computational budgets and energy consumption required for large-scale pre-training. Recent works~\cite{guo2026thousand,yang2025instructvla} illustrate that foundational VLA pre-training is becoming a thousand-GPU-hours effort, requiring large accelerator clusters, distributed optimization, and high-throughput infrastructure. Achieving robust generalization across heterogeneous embodiments demands massive compute resources, creating a significant barrier to entry.

\noindent\textbf{Data Scarcity.}
VLA performance is strongly constrained by the quality and diversity of embodied datasets. Whether through labor-intensive real-world demonstrations~\cite{o2024open, walke2023bridgedata, wu2026pragmatic} or expert-intensive simulation design~\cite{wang2024robogen, tian2025interndata}, curating diverse datasets remains time-consuming and difficult to scale, which slows embodied-policy scaling relative to data-rich vision-language domains.



These profound inefficiencies collectively culminate in a critical deployment bottleneck for applications, particularly in resource-constrained settings such as autonomous vehicles and affordable consumer robotics. Therefore, the pursuit of efficiency is not merely an optional optimization but a fundamental prerequisite for unlocking the transformative potential of VLAs across the broader robotics landscape. 
Among the three, real-time inference is the most direct requirement for physical interaction. 
Training and data efficiency are less visible at deployment time, but they decide whether such systems can be built and scaled.
Addressing this imperative, the field of \textbf{Efficient VLAs} has emerged. As illustrated in~\cref{fig:timelineofvlafoundationmodelsandefficientvlas}, the developmental trajectories of foundational VLAs and efficient VLAs are progressing in parallel. Notably, the timeline reveals a distinct acceleration in the emergence of efficient VLAs, particularly from late 2024 onwards. This proliferation underscores a rapidly intensifying research focus on resource-efficient models, likely driven by the demands of real-world robotic deployment.

\noindent\textbf{Glossary of Efficiency Metrics.}
~\cref{table:efficiency_glossary} outlines a comprehensive glossary of VLA efficiency terms in this survey, grounding theoretical computational metrics into operational physical constraints within the embodied lifecycle.

\begin{table}[t]
    \scriptsize
    \renewcommand{\arraystretch}{1.1} 
    \newcolumntype{L}[1]{>{\raggedright\arraybackslash}m{#1}} 
    \caption{\textbf{Glossary of VLA efficiency metrics.} Standardized definitions of core efficiency dimensions across the model-training-data lifecycle.}
    \label{table:efficiency_glossary}
    \resizebox{\linewidth}{!}{
    \begin{tabular}{
        L{2.1cm}  
        L{6.0cm}  
    }
    \toprule
    \textbf{Terms} & \textbf{Definition} \\
    \midrule
    Inference Latency & End-to-end time from input query to model output. \\
    Control Frequency & The frequency at which a controller outputs control commands or action signals, typically measured in Hz. \\
    Memory Footprint & Runtime or storage memory required by parameters, activations, and cached states. \\
    Training Compute & Total computation consumed during training, typically measured by FLOPs or GPU-hours. \\
    Data Efficiency & The data requirement for achieving target performance level; greater data efficiency corresponds to lower data requirements for comparable performance. \\
    Deployment Efficiency & Practical efficiency under real deployment constraints, including latency, memory, and energy. \\
    \bottomrule
    \end{tabular}
    }
\end{table}

\subsection{Taxonomy Principle}
As discussed above, current VLAs are constrained by a "triple-bottleneck" across the model-training-data lifecycle.
Accordingly, this survey reviews the efficiency of VLAs through the lens of three foundational pillars: model architecture, training paradigm, and data acquisition.
It should be emphasized that these pillars do not represent isolated dimensions.
Instead, they are closely connected and jointly define the efficiency landscape of VLAs:
\\
(1) \textbf{Model $\leftrightarrow$ Training}: Efficient architectures (\eg Mamba backbones~\cite{gu2024mamba,liu2024robomamba} or MoE~\cite{jacobs1991adaptive}) can reduce per-token or activated computation, lowering training and inference cost. Conversely, efficient training strategies (like PEFT~\cite{xu2026parameter,kim2024openvla,kim2025fine}) allow massive foundational models to be adapted without retraining the entire parameter set.
\\
(2) \textbf{Training $\leftrightarrow$ Data}: Data-efficient pre-training focuses on maximizing the information gain per trajectory, reducing the total volume of data required~\cite{ye2025latent}. In parallel, efficient data collection (\eg simulation synthesis) provides the high-throughput fuel necessary for large-scale training regimes~\cite{wang2024robogen,tian2025interndata}.
\\
(3) \textbf{Data $\leftrightarrow$ Model}: Compact model designs often require higher-quality, task-specific data to maintain performance~\cite{kim2024openvla}, whereas scalable data pipelines enable models to tolerate architectural sparsity through robust exposure to diverse scenarios~\cite{o2024open,khazatsky2024droid,wang2024robogen}.

This tripartite synergy binds model, training, and data into a co-dependent ecosystem. 
Based on this foundation, we provide an in-depth survey of these three pillars in Sec.~\ref{sec:efficientmodeldesign}-\ref{sec:efficientdatacollection}, examining how they jointly shape the efficiency and practicality of VLAs.


\begin{table}[t]  
\centering
\scriptsize
\renewcommand{\arraystretch}{1.1} 
\newcolumntype{C}[1]{>{\centering\arraybackslash}m{#1}} 
\newcolumntype{L}[1]{>{\raggedright\arraybackslash}m{#1}} 
\caption{\textbf{Comparison of related VLA surveys.} Prior works predominantly focus on general taxonomies or specific mechanisms, leaving a noticeable gap in systematic efficiency analysis.}
\label{table:relevant_surveys}
\resizebox{\linewidth}{!}{
\begin{tabular}{  
    C{1.8cm}  
    C{0.4cm}  
    C{3.0cm}  
    C{3.0cm}  
}
\toprule
\textbf{Survey} & \textbf{Year} & \textbf{Main Topic} & \textbf{Limitations} \\
\midrule
Ma et al.~\cite{ma2024survey} & 2024  & General VLA taxonomy for embodied AI  & Efficient VLAs scarcely discussed  \\
Shao et al.~\cite{shao2025large} & 2025  & Large VLM-driven robotic manipulation VLAs & Only a few efficient fine-tuning methods mentioned \\
Xiang et al.~\cite{xiang2025parallels} & 2025  & Human motor learning inspired VLA post-training & No efficient solution listed  \\
Zhong et al.~\cite{zhong2025survey} & 2025 & Action tokenization methods in VLAs  & Efficiency limited to tokenization dimension  \\
Din et al.~\cite{din2025vision} & 2025 & Robotic manipulation oriented VLA review & Rarely discuss efficient solutions  \\
Zhang et al.~\cite{zhang2025pure} & 2025 & Pure end-to-end VLA architectures &  Only discuss part of efficient inference \\

\bottomrule
\end{tabular}
}
\end{table}

\subsection{Relevant Surveys}

The rapid growth of VLA research has made it a central paradigm in embodied AI. Several recent surveys have reviewed this field from different perspectives~\cite{ma2024survey, shao2025large, xiang2025parallels, zhong2025survey, din2025vision, zhang2025pure}.
As summarized in~\cref{table:relevant_surveys}, existing surveys mostly present VLAs from a general or capability-centric perspective.
For instance, Ma et al.~\cite{ma2024survey} provide a holistic overview of VLAs in embodied AI, spanning low-level control, high-level task planning, and diverse datasets and benchmarks. Other surveys delve into specific paradigms: Shao et al.~\cite{shao2025large} elucidate VLA evolution through a VLM-centric lens, Xiang et al.~\cite{xiang2025parallels} frame VLA post-training around human motor learning mechanisms, and Zhong et al.~\cite{zhong2025survey} isolate action tokenization as the architectural core, providing a systematic taxonomy of mainstream designs.
Although~\cite{ma2024survey} and~\cite{shao2025large} briefly touch upon works related to efficient VLAs in certain fields, none of them specifically address the efficiency problems from both model, training, and data view. To the best of our knowledge, this survey is the first to specifically focus on the VLA efficiency across the full lifecycle. We hope this survey will help establish efficiency as a central principle of VLA research, guiding the development of scalable and deployable embodied intelligence.

\section{Efficient Model Design}
\label{sec:efficientmodeldesign}


This section studies efficient model design for VLAs under resource constraints. 
A central goal is to improve training and inference efficiency by reducing runtime latency, memory usage, and computational overhead while preserving model capability. 
We organize this section into two complementary directions: 
(1) \textbf{Efficient Architectures}, which improve model structures through modular design, accelerated execution, and sparse computation; and 
(2) \textbf{Model Compression}, which removes redundancy through pruning, quantization, and token optimization.
Representative works on efficient model design are summarized in~\cref{table:efficientmodeldesign}.

\begin{figure*}[!htbp]
  \centering
  \includegraphics[width=\textwidth]{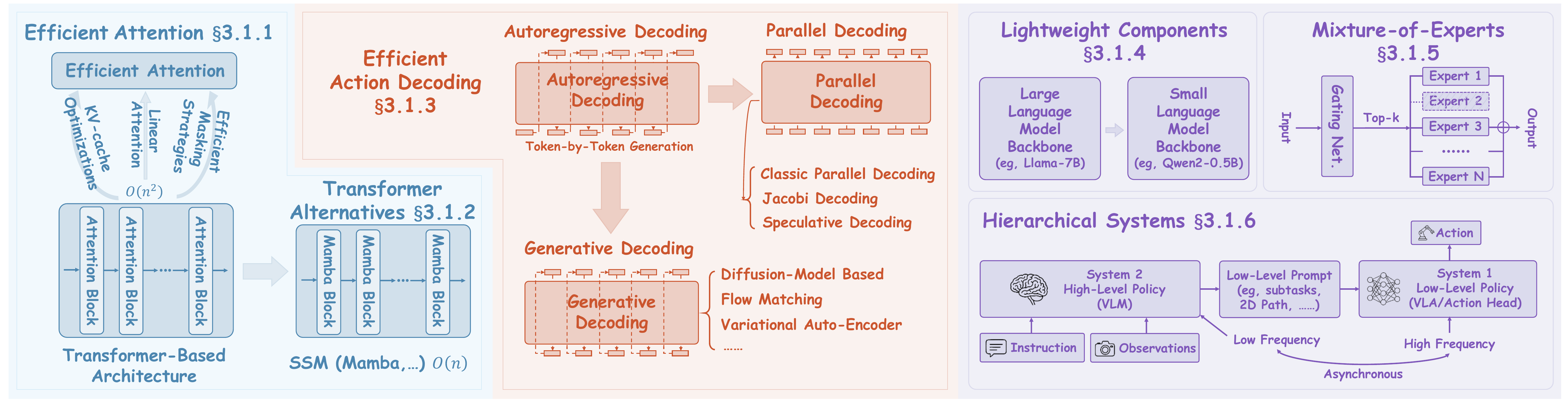}
  \caption{Key strategies for \textbf{Efficient Architectures} (Sec.~\ref{subsec:efficientarchitectures}), including Efficient Attention, Transformer Alternatives, Efficient Action Decoding, Lightweight Components, Mixture-of-Experts, and Hierarchical Systems.}
  \label{fig:efficientarchitectures}
\end{figure*}

\subsection{Efficient Architectures}
\label{subsec:efficientarchitectures}

As illustrated in~\cref{fig:efficientarchitectures}, this subsection reviews representative methods, including efficient attention, transformer alternatives, parallel and generative decoding, lightweight components, mixture-of-experts, and hierarchical processing.

\subsubsection{Efficient Attention}
\label{subsubsec:efficientattention}
Despite providing cross-modal synergy, the $O(L^2)$ complexity of attention mechanisms \cite{vaswani2017attention} hinders real-time, long-horizon robotic control. Recent innovations tackle this via structural linearization, strategic sparsification, and memory optimization. SARA-RT \cite{leal2024sara} utilizes up-training to convert standard Transformers into linear-attention counterparts while preserving model quality. Moving from architecture to interaction, Long-VLA \cite{fan2025long} reduces redundant attention via phase-aware masking, while dVLA \cite{wen2025dvla} enables partial KV caching with prefix masking. To further alleviate inference bottlenecks, RetoVLA \cite{koo2025retovla} repurposes register tokens for global context, while KV-Efficient VLA \cite{xu2025kv} leverages recurrent gating to compress historical KV caches into compact representations. Subsequently, BLURR \cite{ma2026blurr} amortizes repeated attention computation through instruction-prefix KV caching and FlashAttention. More recently, Reflex \cite{guo2026reflex} extends cache reuse to flow-matching VLAs by partitioning attention context into static, sliding, and dynamic regions, enabling $O(1)$ incremental cache updates. Together, these methods enable scalable, low-latency VLA reasoning.

\subsubsection{Transformer Alternatives}
\label{subsubsec:transformeralternatives}
To circumvent the quadratic bottlenecks of Transformers, State-Space Models (SSMs) like Mamba \cite{gu2024mamba} have emerged as high-efficiency alternatives with linear scaling. RoboMamba \cite{liu2024robomamba} pioneers this transition by adopting a Mamba backbone for streamlined embodied reasoning, while FlowRAM \cite{wang2025flowram} further integrates it with conditional flow matching to enhance manipulative precision. These structural pivots catalyze a shift toward leaner yet potent VLA foundations.


\begin{table*}[htbp]
\scriptsize
\renewcommand{\arraystretch}{1.2}
\renewcommand{\tabularxcolumn}[1]{m{#1}}
\newcolumntype{C}[1]{>{\centering\arraybackslash}m{#1}}    
\newcolumntype{L}[1]{>{\raggedright\arraybackslash}m{#1}}  
\newcolumntype{Y}{>{\raggedright\arraybackslash}X}          
\definecolor{lightgray}{gray}{0.9} 
\caption{\textbf{Representative works on Efficient Model Design.} "Categories" markers denote subcategories in Sec.~\ref{sec:efficientmodeldesign}: (A) Efficient Attention, (B) Transformer Alternatives, (C) Efficient Action Decoding, (D) Lightweight Component, (E) Mixture-of-Experts, (F) Hierarchical Systems, (G) Layer Pruning, (H) Quantization, and (I) Token Optimization. For general methods, the smallest experimental parameter scale is reported, with unavailable data denoted by ``$-$''.}
\label{table:efficientmodeldesign}
\resizebox{\textwidth}{!}{
\begin{tabular}{C{3.6cm} C{1.4cm} C{6.0cm} C{1.5cm} L{7.4cm}}
    \toprule
    \textbf{Method/Model}
    & \textbf{Categories}
    & \textbf{Architecture(V/L/A)}
    & \textbf{Param. Scale}
    & \textbf{Key Innovations in Efficient Model Design}\\
    
    \midrule

    SARA-RT~\cite{leal2024sara} & (A) & $-$/$-$/$-$ & \textasciitilde{}5B(PaLI-X) & Quadratic-to-linear transformer up-training. \\ \hline
    
    Long-VLA~\cite{fan2025long} & (A) & ResNet-18/CLIP/Diffusion Model(DDIM)& $-$ & Phase-aware input masking, long-horizon attention optimization. \\ \hline
    
    RetoVLA~\cite{koo2025retovla} & (A) & SigLIP/SmolLM2/Flow Matching Model & $-$ & Discarded register token reuse for spatial reasoning. \\ \hline
    
    KV-Efficient VLA~\cite{xu2025kv} & (A) (I) & DINOv2+SigLIP/LLaMA-2/$-$ & \textasciitilde{}7B & RNN-gated chunked KV caching, selective context pruning. \\ \hline
    
    dVLA~\cite{wen2025dvla} & (A) & MAGViT-v2/MMaDA/FAST& $-$ & Prefix attention masking, KV caching for diffusion speedup. \\ \hline
    
    RoboMamba~\cite{liu2024robomamba} & (B) (D) &  CLIP/Mamba/MLP & \textasciitilde{}3.2B & Linear-complexity Mamba reasoning, lightweight action decoding. \\ \hline
    
    FlowRAM~\cite{wang2025flowram} & (B) (C) & CLIP/Mamba/MLP & $-$ & Region-aware Mamba fusion, flow matching. \\ \hline
    
    TinyVLA~\cite{wen2025tinyvla} & (C) (D) & ViT/Pythia/Diffusion Policy & 1.3B & Compact VLM with a DP decoder for data-efficient adaptation. \\ \hline
    
    PD-VLA~\cite{song2025accelerating} & (C) & CLIP/Vicuna/$-$ & \textasciitilde{}7B & Parallel fixed-point iteration for action chunking. \\ \hline
    
    OpenVLA-OFT~\cite{kim2025fine} & (C) & DINOv2+SigLIP/LLaMA-2/MLP & \textasciitilde{}7B & Bidirectional attention for parallel continuous action decoding. \\ \hline
    
    HybridVLA~\cite{liu2025hybridvla} & (C) (I) & DINOv2+SigLIP/LLaMA-2/MLP & \textasciitilde{}7B & Diffusion-autoregression merger, pre-diffusion token KV caching. \\ \hline
    
    FreqPolicy~\cite{su2025freqpolicy} & (C) & ResNet-18+MLP/LLaMA-2/U-Net & \textasciitilde{}7B & Frequency consistency optimization for flow-based policies. \\ \hline
    
    CEED-VLA~\cite{song2025ceed} & (C) & $-$/$-$/$-$ & \textasciitilde{}7B & Consistency distillation, early-exit parallel Jacobi decoding. \\ \hline
    
    MinD~\cite{chi2025mind} & (C) (F) & CLIP/$-$/DiT & $-$ & Dual-system world models, asynchronous diffusion. \\ \hline
    
    VQ-VLA~\cite{wang2025vq} & (C) & DINOv2+SigLIP/LLaMA-2/VQ Decoder(VQ-VAE) & $-$ & Scaled quantized tokenizers for discrete action smoothing. \\ \hline
    
    Spec-VLA~\cite{wang2025spec} & (C) & DINOv2+SigLIP/LLaMA-2/$-$ & \textasciitilde{}7B & Relaxed speculative decoding, parallel token verification. \\ \hline
    
    AMS~\cite{zheng2025leveraging} & (C) (I) & $-$/$-$/$-$ & $-$ & OS-level management with two-phase hashing for action reuse. \\ \hline
    
    NinA~\cite{tarasov2025nina} & (C) & $-$/$-$/$-$ & $-$ & Normalizing flows for continuous action decoding. \\ \hline
    
    Discrete Diffusion VLA~\cite{liang2025discrete} & (C) & DINOv2+SigLIP/LLaMA-2/$-$ & \textasciitilde{}7B & Discrete diffusion embedding, unified action decoding. \\ \hline
    
    CLIP-RT~\cite{kang2024clip} & (D) & CLIP-ViT/CLIP/$-$ & \textasciitilde{}1B & Pre-trained CLIP adaptation, 7x parameter reduction. \\ \hline
    
    SVLR~\cite{samson2025scalable} & (D) & Mini-InternVL/Phi-3-mini/$-$ & $-$ & Modular small model integration, consumer GPU deployment. \\ \hline
    
    NORA~\cite{hung2025nora} & (D) & Qwen-2.5-VL(VLM)/FAST+ & \textasciitilde{}3B & Compact Qwen2.5-VL-3B backbone with FAST+ tokenization. \\ \hline
    
    SmolVLA~\cite{shukor2025smolvla} & (D) (G) (I) & SigLIP/SmolLM2/Flow Matching Transformer & 450M & $N=L/2$ layer pruning, pixel-shuffle compression. \\ \hline
    
    SP-VLA~\cite{li2025sp} & (D) (I) & DINOv2+SigLIP/LLaMA-2/$-$ & $-$ & Action-aware scheduling and spatio-semantic token pruning. \\ \hline
    
    EdgeVLA~\cite{budzianowski2025edgevla} & (D) & DINOv2+SigLIP/Qwen2-0.5B/$-$ & \textasciitilde{}1B & Non-autoregressive small language models for edge devices. \\ \hline
    
    MiniVLA~\cite{belkhale2024minivla} & (D) & DINOv2+SigLIP/Qwen2.5-0.5B/$-$ & \textasciitilde{}1B & Small LLM integration, 7x parameter reduction. \\ \hline
    
    GeRM~\cite{song2024germ} & (E) & $-$/$-$/$-$ & $-$ & MoE integration for fast inference, higher capacity. \\ \hline
    
    FedVLA~\cite{miao2025fedvla} & (E) & $-$/$-$/$-$ & $-$ & Dual-gating MoE, federated adaptive expert activation. \\ \hline
    
    TVVE~\cite{bai2025learning} & (E) & $-$/$-$/$-$ & $-$ & MoE visual encoder, task-disentangled feature extraction. \\ \hline
    
    HiRT~\cite{zhang2025hirt} & (F) & EfficientNet/InstructBLIP/MAP& $-$ & Hierarchical transformers, low-to-high frequency policy guidance. \\ \hline
    
    RoboDual~\cite{bu2024towards} & (F) & DINOv2+SigLIP/LLaMA-2/DiT & \textasciitilde{}7B & Generalist VLA, high frequency diffusion specialist. \\ \hline
    
    DP-VLA~\cite{han2024dual} & (F) & DINOv2+SigLIP/LLaMA-2/BC-Transformer & $-$ & Dual-process framework, reasoning-executor co-design. \\ \hline
    
    HAMSTER~\cite{li2025hamster} & (F) & SigLIP/Vicuna-13B-v1.5/3D-DA & $-$ & High-level VLM 2D paths, low-level policy guidance. \\ \hline
    
    FiS~\cite{chen2025fast} & (F) & DINOv2+SigLIP/LLaMA-2/$-$ & $-$ & Execution-in-reasoning parameter sharing with dual-system. \\ \hline
    
    
    MoLe-VLA~\cite{zhang2025mole} & (G) & DINOv2+SigLIP/LLaMA-2/DiT & $-$ & Spatio-temporal routing, selective LLM layer skipping. \\ \hline
    
    EfficientVLA~\cite{yang2025efficientvla} & (G) (I) & DINOv2+SigLIP/LLaMA-2/DiT & \textasciitilde{}5B & Training-free layer pruning, task-relevant token compression. \\ \hline
    
    RLRC~\cite{chen2025rlrc} & (G) (H) & $-$/$-$/$-$ & $-$ & Structured layer pruning, RL recovery, 4-bit quantization. \\ \hline
    
    
    FLOWER~\cite{reuss2025flower} & (G) & Florence-2 ViT/Florence-2 LLM/Flow Transformer & 950M & Intermediate modality fusion, 50\% LLM layer pruning. \\ \hline
    
    OpenVLA~\cite{kim2024openvla} & (H) & DINOv2+SigLIP/LLaMA-2/$-$ & \textasciitilde{}7B & 7B VLA quantization framework, zero success-rate degradation. \\ \hline
    
    QAIL~\cite{park2024quantization} & (H) & $-$/$-$/$-$ & $-$ & Quantization-aware fine-tuning, low-bit error mitigation. \\ \hline
    
    FAST~\cite{pertsch2025fast} & (H) (I) & $-$/$-$/$-$ & $-$ & DCT/BPE action tokenization for action sequences. \\ \hline
    
    SQIL~\cite{park2025saliency} & (H) & $-$/$-$/$-$ & $-$ & Saliency-based loss weighting, quantization-aware training. \\ \hline
    
    BitVLA~\cite{wang2025bitvla} & (H) & SigLIP-L/BitNet b1.58 2B4T/$-$ & \textasciitilde{}3B & 1-bit ternary quantization, vision encoder distillation alignment. \\ \hline
    
    SQAP-VLA~\cite{fang2025sqap} & (H) (I) & Prism-DinoSigLIP/Llama-2/DiT & $-$ & Quantization-aware token pruning with quantizer enhancement. \\ \hline
    
    VLA-Cache~\cite{xu2025vla} & (I) & $-$/$-$/$-$ & $-$ & Inter-frame change-aware token selection and KV-cache reuse. \\ \hline
    
    FlashVLA~\cite{tan2025think} & (I) & $-$/$-$/$-$ & $-$ & Visual token pruning via ICS, stability-based action reuse. \\ \hline
    
    Fast ECoT~\cite{duan2025fast} & (I) & DINOv2+SigLIP/LLaMA-2/$-$ & $-$ & High-level reasoning token caching, ECoT cross-timestep reuse. \\ \hline
    
    CronusVLA~\cite{li2025cronusvla} & (I) & DINOv2+SigLIP/LLaMA-2/DiT & \textasciitilde{}7B & FIFO-queued motion feature caching, multi-frame inference. \\ \hline
    
    VOTE~\cite{lin2025vote} & (I) & DINOv2+SigLIP/LLaMA-2/MLP & \textasciitilde{}7B & Action chunk compression into single <ACT> tokens. \\ \hline
    
    CogVLA~\cite{li2025cogvla} & (I) & DINOv2+SigLIP/LLaMA-2/$-$ & \textasciitilde{}7B & Instruction-driven token aggregation, FiLM routing, thresholds. \\ \hline
    
    SpecPrune-VLA~\cite{wang2025specprune} & (I) & $-$/$-$/$-$ & $-$ & Dynamic token pruning, global-local attention, velocity control. \\ \hline
    
    LightVLA~\cite{jiang2025better} & (I) & DINOv2+SigLIP/LLaMA-2/MLP & \textasciitilde{}7B & Gumbel-Softmax differentiable token pruning. \\ \hline
    
    ADP~\cite{pei2025action} & (I) & DINOv2+SigLIP/LLaMA-2/MLP & \textasciitilde{}7B & Trajectory-gated, text-guided vision token pruning. \\ \hline
    
    Oat-VLA~\cite{bendikas2025focusing} & (I) & DINOv2+SigLIP/LLaMA-2/$-$ & \textasciitilde{}7B & Gripper-guided object/agent-centric token compression. \\
    
    \bottomrule
\end{tabular}
}
\end{table*}

\subsubsection{Efficient Action Decoding}
\label{subsubsec:efficientactiondecoding}
Standard autoregressive (AR) decoding in VLAs \cite{kim2024openvla} treats actions as discrete tokens, which incurs $O(N)$ latency and hinders real-time control. Efficiency is enhanced by either parallelizing the AR stream or replacing it with holistic generative decoders.

\noindent\textbf{Parallel Decoding.}
To mitigate AR bottlenecks, OpenVLA-OFT \cite{kim2025fine}, EdgeVLA \cite{budzianowski2025edgevla}, and SemanticVLA \cite{li2026semanticvla} exploit bidirectional decoding for single-pass action chunk prediction. PD-VLA \cite{song2025accelerating} and its early-exit refinement CEED-VLA \cite{song2025ceed} recast decoding as nonlinear fixed-point equations solvable via parallel Jacobi iterations. Complementarily, Spec-VLA \cite{wang2025spec} accelerates speculative decoding via relaxed acceptance, while SV-VLA \cite{wang2026open} extends speculation to closed-loop verification with deviation-triggered replanning.

\noindent\textbf{Generative Decoding.}
Generative paradigms supplant token-by-token chains with trajectory synthesis. TinyVLA \cite{wen2025tinyvla} and HybridVLA \cite{liu2025hybridvla} integrate diffusion decoders for continuous actions, while FlowRAM \cite{wang2025flowram}, NinA \cite{tarasov2025nina}, and FreqPolicy \cite{su2025freqpolicy} leverage Flow Matching or Normalizing Flows to bypass iterative denoising for one-shot or few-step sampling. Further optimizations include MinD's \cite{chi2025mind} world-model latents, AMS's \cite{zheng2025leveraging} computation reuse, and VQ-VLA's \cite{wang2025vq} high-fidelity discretization. Finally, Discrete Diffusion VLA \cite{liang2025discrete} unifies these by combining discrete-token interfaces with adaptive re-masking for efficient error correction.

\subsubsection{Lightweight Components}
\label{subsubsec:lightweightcomponent}
Lightweight components offer a direct path to efficient VLAs by balancing parametric parsimony with operational efficacy. RoboMamba \cite{liu2024robomamba} exemplifies this with a 3.7M-parameter MLP head for 6-DoF pose prediction. Scaling this philosophy to the backbone, TinyVLA \cite{wen2025tinyvla}, EdgeVLA \cite{budzianowski2025edgevla}, and MiniVLA \cite{belkhale2024minivla} leverage sub-2B SLMs, while NORA \cite{hung2025nora} adopts a compact 3B backbone and Evo-1 \cite{lin2026evo} preserves semantic alignment at 0.77B. DynamicVLA \cite{xie2026dynamicvla} further pairs a 0.4B VLA with convolutional vision encoding for dynamic manipulation.

CLIP-RT \cite{kang2024clip} and DiVLA-2B \cite{wen2024diffusion} further show that 1--2B models can outperform 7B counterparts in throughput and task success. For edge deployment, SVLR \cite{samson2025scalable} combines retraining-free lightweight modules, while LiteVLA \cite{williams2026lite} enables 4-bit CPU-only deployment on Raspberry Pi 4. Finally, SP-VLA \cite{li2025sp} introduces an action-aware scheduler that dynamically alternates between heavy and light generators, achieving frequency-adaptive acceleration with negligible performance loss. Together, these designs reduce VLA size and inference cost for edge deployment.

\subsubsection{Mixture-of-Experts}
\label{subsubsec:moe}
Mixture-of-Experts (MoE) architectures decouple model capacity from computational cost by routing tokens to specialized subnetworks. GeRM \cite{song2024germ} pioneers sparse MoE in quadruped reinforcement learning, demonstrating that selective activation scales multi-task expressiveness while maintaining inference thrift. Extending this paradigm, FedVLA \cite{miao2025fedvla} proposes Dual Gating MoE (DGMoE), utilizing bidirectional token selection to dynamically sparsify computational graphs beyond standard top-$K$ mechanisms. Similarly, TVVE \cite{bai2025learning} introduces Task-Aware MoE (TaskMoE) to conditionally engage experts based on linguistic-visual signals, enabling cross-task interference mitigation through semantic routing. Collectively, these frameworks catalyze modular, adaptive VLAs that harmonize expansive generalization with pragmatic deployment.




\subsubsection{Hierarchical Systems}
\label{subsubsec:hierarchicalsystems}
Inspired by dual-process theories \cite{wason1974dual, kahneman2011thinking}, hierarchical VLAs bifurcate cognition into deliberate "System 2" reasoning and intuitive "System 1" execution. This decoupling lets compute-intensive VLMs guide latency-sensitive control asynchronously, reconciling semantic depth with real-time responsiveness.

HiRT \cite{zhang2025hirt} and DP-VLA \cite{han2024dual} establish low-frequency VLM guidance for high-frequency control, while SmolVLA \cite{shukor2025smolvla} decouples inference from action execution. RoboDual \cite{bu2024towards} and HAMSTER \cite{li2025hamster} condition low-level policies on generalist outputs or 2D paths. FiS \cite{chen2025fast} partially shares parameters across systems, Fast ECoT \cite{duan2025fast} asynchronously decouples reasoning from action decoding, and MinD \cite{chi2025mind} couples low-frequency foresight with high-frequency diffusion control. Recent variants refine the temporal interface: DuoCore-FS \cite{zou2025asynchronous} buffers slow-path latents for whole-body control, Yan et al. \cite{yan2026acting} caches semantics within existing VLAs, and UniFS \cite{sun2026unifs} stratifies VLM layers by update frequency. Collectively, these hierarchical architectures scale cognitive depth without compromising operational alacrity.

\subsection{Model Compression}
\label{sec:modelcompression}

\begin{figure*}[!htbp]
  \centering
  \includegraphics[width=\textwidth]{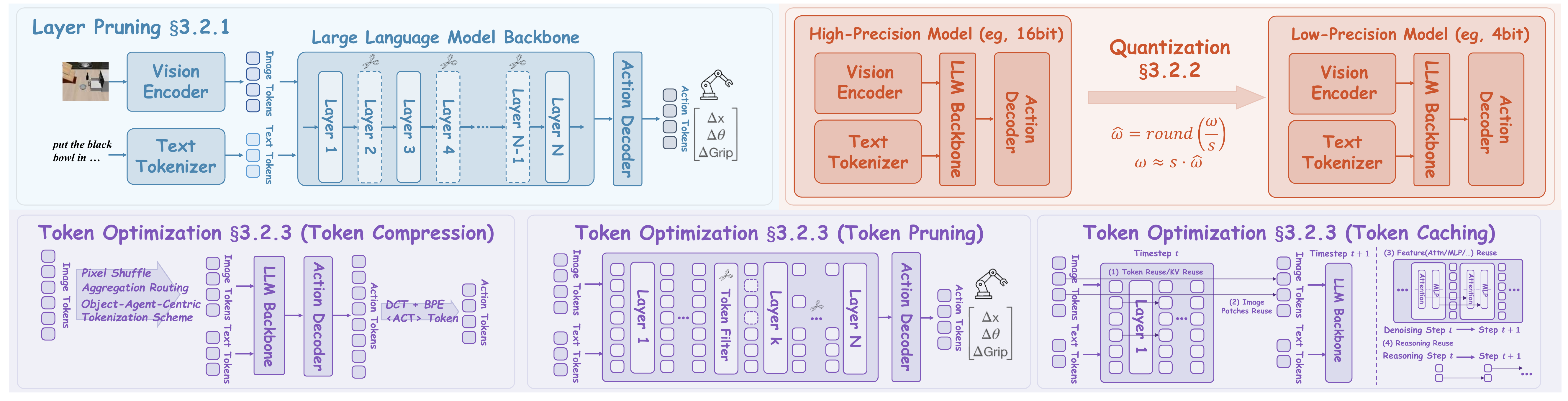}
  \caption{Key strategies for \textbf{Model Compression} (Sec.~\ref{sec:modelcompression}), including Layer Pruning, Quantization, and Token Optimization through compression, pruning, and caching.}
  \label{fig:modelcompression}
\end{figure*}

As illustrated in~\cref{fig:modelcompression}, this subsection systematically surveys pivotal techniques, from layer pruning and quantization to token optimization, elucidating their general schemes and how they benefit efficiency to cultivate robust, edge-compatible architectures.

\subsubsection{Layer Pruning}
\label{subsec:layerpruning}
Layer pruning strategically removes or bypasses redundant layers to reduce model size and latency, exploiting the high interlayer similarity of large-scale VLMs \cite{zhang2025mole, yang2025efficientvla}. These methods span fixed structural pruning and adaptive layer skipping.

Fixed schemes impose predetermined sparsity for efficient inference. SmolVLA \cite{shukor2025smolvla} truncates the VLM at half depth, EfficientVLA \cite{yang2025efficientvla} performs training-free importance-guided layer pruning, and RLRC \cite{chen2025rlrc} couples structured pruning with SFT/RL recovery. LightDP \cite{wu2025device} learns SVD-initialized DiT block pruning with consistency distillation, while FLOWER \cite{reuss2025flower} prunes architecture-specific VLM layers via intermediate fusion.

Adaptive schemes instead allocate depth online. DeeR-VLA \cite{yue2024deer} employs action-consistency-based early exits, while MoLe-VLA \cite{zhang2025mole} dynamically routes layers via STAR. DySL-VLA \cite{yang2026dysl} further retains informative layers statically while skipping incremental layers dynamically according to action importance.

\subsubsection{Quantization}
\label{subsec:quantization}
Quantization discretizes weights and activations into low-bit representations, drastically curtailing memory footprints while preserving the fidelity of multimodal action synthesis. OpenVLA \cite{kim2024openvla} shows that 4-bit PTQ can halve GPU memory with negligible performance loss. QAIL \cite{park2024quantization} introduces quantization-aware imitation learning, while SQIL \cite{park2025saliency} further emphasizes mission-critical states via saliency-weighted training. BitVLA \cite{wang2025bitvla} pushes VLAs to ternary 1-bit precision with distillation-aware vision compression, while RLRC \cite{chen2025rlrc} applies 4-bit quantization after SFT/RL recovery.

Beyond model weights, FAST \cite{pertsch2025fast} compresses actions through DCT/BPE tokenization, while SQAP-VLA \cite{fang2025sqap} co-designs quantization and token pruning with Hadamard stabilization. More recently, QVLA \cite{xu2026qvla} allocates channel-wise bits by action sensitivity, while QuantVLA \cite{zhang2026quantvla} performs scale-calibrated PTQ across language and DiT action modules. Collectively, these advancements sculpt VLAs for resource-frugal, ubiquitous embodiment.

\subsubsection{Token Optimization}
\label{subsec:tokenoptimization}
Token optimization strategically refines representational sequences in VLAs to mitigate the quadratic escalation of Transformer-based computations. By orchestrating compression, pruning, and caching, these paradigms curtail redundancy and expedite embodied reasoning on resource-constrained platforms.

\noindent\textbf{Token Compression.} Token compression aggregates voluminous streams into succinct representations to bolster inference velocity. For visual inputs, SmolVLA \cite{shukor2025smolvla} employs pixel shuffling to limit tokens per frame, while CogVLA \cite{li2025cogvla}, Oat-VLA \cite{bendikas2025focusing}, and Compressor-VLA \cite{gao2025compressor} compress task-relevant content via instruction-driven aggregation, object-agent-centric tokenization, and instruction-guided semantic-spatial compression, respectively. Action tokens are similarly condensed: FAST \cite{pertsch2025fast} combines DCT with BPE for action sequence compression, whereas VOTE \cite{lin2025vote} represents each action chunk with a single <ACT> token decoded by an action head.

\noindent\textbf{Token Pruning.} Pruning excises redundant tokens based on saliency metrics to streamline computational demands. FlashVLA \cite{tan2025think} and EfficientVLA \cite{yang2025efficientvla} prune visual tokens by information contribution and task-aware diversity, respectively, while SP-VLA \cite{li2025sp} adds spatio-semantic importance. SpecPrune-VLA \cite{wang2025specprune} exploits cross-step attention and action granularity; SQAP-VLA \cite{fang2025sqap} stabilizes pruning under quantization; and LightVLA \cite{jiang2025better} learns differentiable selection via Gumbel-softmax. KV-Efficient VLA \cite{xu2025kv} recurrently gates compressed historical KV chunks, whereas ADP \cite{pei2025action} adapts token retention to recent action dynamics. More recently, VLA-Pruner \cite{liu2025bridging} jointly scores semantic and action relevance via prefill and temporally smoothed decode attention, while TEAM-VLA \cite{ye2026token} combines current-frame foreground expansion with action-guided token merging.

\noindent\textbf{Token Caching.} Token caching exploits temporal invariances to eliminate redundant computations across iterative steps. VLA-Cache \cite{xu2025vla} reuses static visual KV entries across frames, while HybridVLA \cite{liu2025hybridvla} extends KV caching to diffusion-based action generation. FlashVLA \cite{tan2025think} reuses stable actions, whereas Fast ECoT \cite{duan2025fast} caches high-level reasoning. Beyond KV reuse, EfficientVLA \cite{yang2025efficientvla} caches intermediate diffusion features, while CronusVLA \cite{li2025cronusvla} reuses compact motion features across frames. AMS \cite{zheng2025leveraging} generalizes reuse through a hardware-aware context pool spanning KV states, diffusion latents, and action outputs. More recently, DySta \cite{qiu2026efficient} disentangles multi-level static and dynamic visual tokens and learns when to refresh static KV caches, enabling efficient long-horizon multi-frame inference.

\subsection{Discussion}
\subsubsection{Innovations}
Current designs for efficient VLAs have shifted from a purely scaling-centric paradigm toward adaptive, lightweight architectures. Linear sequence alternatives, such as the Mamba backbone~\cite{liu2024robomamba,wang2025flowram}, alleviate the quadratic computational bottleneck of traditional Transformer architectures. Concurrently, parallel decoding strategies, such as bidirectional masked prediction~\cite{kim2025fine,budzianowski2025edgevla} and parallel Jacobi iteration~\cite{song2025accelerating,song2025ceed}, together with generative trajectory prediction~\cite{wen2025tinyvla,liu2025hybridvla,su2025freqpolicy}, reduce token-by-token autoregressive action decoding overhead. Furthermore, hierarchical system designs~\cite{zhang2025hirt,han2024dual} decouple high-level semantic planning from low-level real-time control, while dynamic token pruning~\cite{yang2025efficientvla,li2025sp}, compression~\cite{pertsch2025fast,li2025cogvla}, and caching~\cite{xu2025vla,tan2025think} techniques reduce redundant computation with limited performance loss in reported settings.

\subsubsection{Limitations, Challenges and Future Works} 
A fundamental trade-off exists between structural sparsification and high-fidelity behavioral representations. Extensive model pruning and low-bit quantization can compromise latent planning depth~\cite{yue2024deer,park2024quantization}, degrading fine-grained manipulation precision in complex, long-horizon tasks. Furthermore, asynchronous hierarchical systems and parallel decoding frameworks often suffer from spatiotemporal misalignment in dynamic environments~\cite{zhang2025hirt}, which induces control oscillations and jeopardizes hardware safety.

To address these bottlenecks, future architectures must prioritize intrinsic adaptability by coupling algorithmic design with hardware deployment. Through hardware-software co-design and system-level optimization~\cite{zheng2025leveraging}, models can be aligned with specific execution profiles, effectively mitigating computational constraints across heterogeneous edge devices. Ultimately, this co-optimization paradigm enables scalable deployment from consumer devices to industrial manipulators without compromising cross-modal reasoning capabilities.


\begin{figure*}[!htbp]
  \centering
  \includegraphics[width=\textwidth]{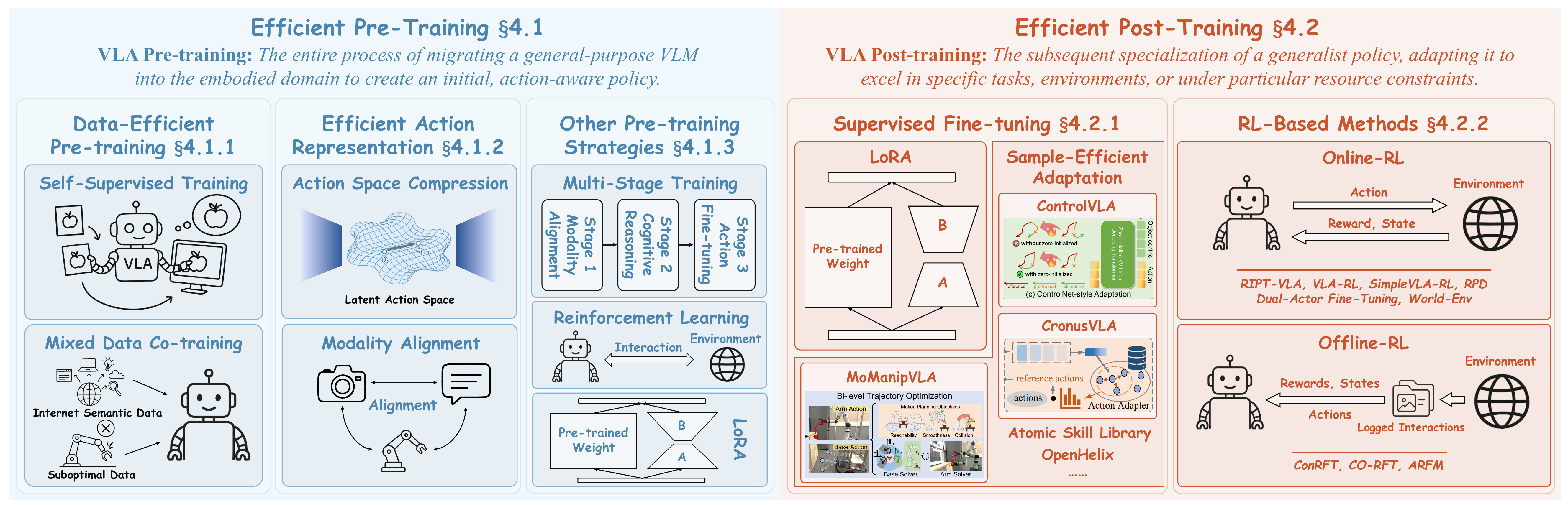}
  \caption{Key strategies for \textbf{Efficient Training} (Sec.~\ref{sec:efficienttraining}), comprising Efficient Pre-Training and Efficient Post-Training with their corresponding strategies.}
  \label{fig:efficienttraining}
\end{figure*}

\section{Efficient Training}
\label{sec:efficienttraining}

While foundational VLAs leverage pre-trained VLM backbones for robust multimodal reasoning, they inherit prohibitive computational and data overheads, necessitating the advancement of \textbf{Efficient Training} methodologies to minimize resource footprint while maintaining competitive performance. As illustrated in~\cref{fig:efficienttraining}, this chapter systematically explores the spectrum of these techniques, focusing on two pivotal stages: (1) \textbf{Efficient Pre-training}, instilling foundational action capabilities into pre-trained VLMs or training the entire VLA from scratch with minimal overhead, and (2) \textbf{Efficient Post-training}, enabling the swift and effective deployment of VLAs to specific downstream tasks. We list representative works in~\cref{table:efficienttraining}.

\subsection{Efficient Pre-Training}
\label{subsec:efficientpretraining}

As illustrated in~\cref{fig:efficienttraining} left, this section systematically examines advancements categorized into data-efficient pre-training, efficient action representation, and other efficient pre-training strategies. By highlighting these innovations, we outline the critical pathways for developing robust, general VLA policies without incurring the traditional burden of extensive, resource-intensive pre-training cycles.

\subsubsection{Data-Efficient Pre-training}
\label{subsubsec:dataefficientpretraining}
To mitigate the data scarcity inherent in embodied learning, data-efficient pre-training prioritizes the strategic utilization of scarce robotic trajectories alongside abundant non-robotic sources, such as ego-centric videos. This paradigm primarily evolves through two axes: (1) self-supervised objectives that extract intrinsic signals from unlabeled data, and (2) mixed-data co-training frameworks designed to bridge domain gaps.

\noindent\textbf{Self-Supervised Training.}
Self-supervised training synthesizes supervisory signals from unlabeled or passive datasets to alleviate the prohibitive cost of expert demonstrations. Current research focuses on two frontiers.

First, to maximize the utility of existing robotic data, DTP~\cite{fan2025diffusion} employs a diffusion-based trajectory prediction objective. By modeling future end-effector trajectories in the RGB domain, it bridges the vision-action modality gap and improves long-horizon generalization.

Second, a more prevalent direction leverages internet-scale ego-centric videos to bypass the need for explicit action labels. While LAPA~\cite{ye2025latent} pioneered learning discrete latent action spaces from action-less videos, subsequent works have addressed its susceptibility to task-irrelevant noise (\eg camera shake). For instance, Bu et al.~\cite{bu2025learning} introduced a task-centric VQ-VAE to decouple essential actions from environmental dynamics, while LAWM~\cite{tharwat2025latent} extended this principle to compact world-model architectures. Expanding beyond latent spaces, EgoVLA~\cite{yang2025egovla} and Being-H0~\cite{luo2025being} utilize human anatomical priors (\eg MANO parameters) to map human hand movements to robotic manipulation. Complementarily, RynnVLA-001~\cite{jiang2025rynnvla} and Wang et al.~\cite{wang2025unified} demonstrate that pre-training on environment dynamics via generative world models provides a robust foundation for sample-efficient policy fine-tuning.

\noindent\textbf{Mixed Data Co-training.}
Mixed-data co-training enhances VLA efficiency and transferability by unifying heterogeneous datasets of varying quality and modality. GeRM~\cite{song2024germ} utilizes a Conservative Q-Learning (CQL) framework to robustly ingest both expert and suboptimal data, mitigating overestimation on out-of-distribution actions. To manage cross-domain complexity, HAMSTER~\cite{li2025hamster} adopts a hierarchical decoupling strategy—predicting 2D trajectories to condition 3D control—thereby efficiently assimilating massive non-robotic data. Similarly, GraspVLA~\cite{deng2025graspvla} integrates autoregressive perception with flow-matching action generation via Chain-of-Thought (CoT) reasoning. Furthermore, AnyPos~\cite{tan2025anypos} uses arm-decoupled estimation to extract transferable motion primitives from task-agnostic data, while TwinVLA~\cite{im2026twinvla} composes pretrained single-arm policies for bimanual control without bimanual pre-training. Similarly, WSA$_1$~\cite{jiang2026wsa} combines heterogeneous data with 3D world-action joint modeling for data-efficient generalization.

\subsubsection{Efficient Action Representation}
\label{subsubsec:efficientactionrepresentation}
The high dimensionality and inherent noise of robot action spaces frequently impede VLA training efficiency and generalization. To address this, current research transitions from modeling raw control signals to learning compact, semantic motion primitives, primarily through action space compression and innovative structural modeling.

\noindent\textbf{Action Space Compression.}
A dominant strategy involves projecting continuous actions into distilled latent spaces to truncate the policy search space. While LAPA~\cite{ye2025latent} and Bu et al.~\cite{bu2025learning} utilize VQ-VAE~\cite{van2017neural} for its robust quantization benefits, RynnVLA-001~\cite{jiang2025rynnvla} and LAWM~\cite{tharwat2025latent} leverage standard VAE architectures or world-model priors to learn these embeddings. Moving beyond per-step encoding, FAST~\cite{pertsch2025fast} achieves significant pre-training speedups by applying sequence-level compression, such as DCT and BPE, thereby reducing dimensionality in the temporal domain. More recently, INN Adapter~\cite{zhang2026invertible} constrains one-step flow matching within an invertible latent action manifold, preserving information while avoiding iterative denoising.

\noindent\textbf{Modality Alignment.}
Complementarily, efficiency is enhanced by aligning action spaces with visual or anatomical priors. EgoVLA~\cite{yang2025egovla} bridges the human-robot domain gap through MANO-based shared action spaces, while VLA-Adapter~\cite{wang2025vla} employs a Bridge Attention module to facilitate cross-modal fusion. Furthermore, structural optimizations, such as cVLA’s~\cite{argus2025cvla} adoption of image-coordinate-based encoding or ReSET’s~\cite{dai2025prepare} restructuring states, simplify the complex mapping from vision to actuation. Ultimately, these advancements enable VLA models to bypass low-level control noise and focus on mastering transferable motion primitives.

\begin{table*}[htbp]
    \centering
    \scriptsize
    \renewcommand{\arraystretch}{1.2}
    \renewcommand{\tabularxcolumn}[1]{m{#1}}
    \newcolumntype{C}[1]{>{\centering\arraybackslash}m{#1}}  
    \newcolumntype{L}[1]{>{\raggedright\arraybackslash}m{#1}} 
    \newcolumntype{Y}{>{\raggedright\arraybackslash}X}          
    \definecolor{lightgray}{gray}{0.9} 
    
    \caption{\textbf{Representative works on Efficient Training.} The ``Training Phase'' column denotes the specific learning stage where efficient optimizations are targeted. The ``Strategies'' markers ``IL'', ``On-RL'', and ``Off-RL'' correspond to efficiency-focused Imitation Learning, Online Reinforcement Learning, and Offline Reinforcement Learning paradigms, respectively. ``Trainable modules'' column outlines the specific architectural components updated during training.}
    \label{table:efficienttraining}
    \resizebox{\textwidth}{!}{
        \begin{tabular}{C{3.3cm} C{2.3cm} C{1.3cm} C{6.0cm} p{9.5cm}}
            \toprule
            \textbf{Method/Model}
            & \textbf{Training Phase}
            & \textbf{Strategies}
            & \textbf{Trainable modules}
            & \textbf{Key Innovations in Efficient Training}\\
            
            \midrule
        
            GeRM
            \cite{song2024germ}
            & Pre-training
            & Off-RL
            & decoder-only Transformer + sparse MoE experts
            & Sparse MoE routing, offline RL over demonstration and sub-optimal data.  \\
            
            LAPA
            \cite{ye2025latent}
            & Pre-training
            & IL
            & VLM backbone + action quantization model
            & Unsupervised VQ-VAE quantization, label-free latent action extraction from video. \\
            
            HAMSTER
            \cite{li2025hamster}
            & Pre-training
            & IL
            & vision encoder + LLM backbone + action module
            & Hierarchical action modeling, embodiment-agnostic 2D path representation. \\
            
            DTP
            \cite{fan2025diffusion}
            & Multi-stage training
            & IL
            & diffusion trajectory model + action module
            & Diffusion-based trajectory synthesis, long-horizon task training configuration.  \\
            
            Humanoid-VLA
            \cite{ding2025humanoid}
            & Multi-stage training
            & IL
            & vision encoder + cross-attention module + LLM backbone + codebook
            & Language--motion pre-alignment, parameter-efficient egocentric visual adaptation. \\
            
            GraspVLA
            \cite{deng2025graspvla}
            & Multi-stage training
            & IL
            & LLM backbone + projector + flow action expert
            & Synthetic data scaling, dual-stage grasping pre-training pipeline. \\
            
            UniVLA (Bu et al. )
            \cite{bu2025learning}
            & Multi-stage training
            & IL
            & VLM backbone + action expert
            & Language-conditioned task-centric video latents for cross-embodiment policy learning. \\
            
            UniVLA (Wang et al.)
            \cite{wang2025unified}
            & Pre-training
            & IL
            & multimodal transformer
            & Multimodal tokenization, sparse data stream unification. \\
            
            EgoVLA
            \cite{yang2025egovla}
            & Pre-training
            & IL
            & VLM backbone + action head
            & MANO-shared action space for egocentric transfer, human-to-robot retargeting. \\
            
            AnyPos
            \cite{tan2025anypos}
            & Pre-training
            & IL
            & DINOv2-Reg + DAD + MLP
            & Inverse dynamics modeling, task-agnostic data utilization pipeline. \\
            
            Being-H0
            \cite{luo2025being}
            & Pre-training
            & IL
            & LMM backbone + motion tokenizer + action expert 
            & Video-derived hand motion priors, physical instruction tuning implementation. \\
            
            RynnVLA-001
            \cite{jiang2025rynnvla}
            & Multi-stage training
            & IL
            & AR Transformer + ActionVAE + action head
            & ActionVAE chunk compression and continuous mapping for multi-stage transfer.\\
            
            LAWM
            \cite{tharwat2025latent}
            & Pre-training
            & IL
            & imitation learning model + world model
            & Self-supervised imitation-world models for latent forecasts and action extraction. \\
            
            FAST
            \cite{pertsch2025fast}
            & Pre-training
            & IL
            & autoregressive VLA with FAST action tokenizer
            & DCT and BPE integration, action sequence compression via quantization.\\
            
            cVLA
            \cite{argus2025cvla}
            & Pre-training
            & IL
            & PaliGemma2 self-attention layers
            & Image-space discrete keypose prediction, simulation-based training structure. \\
            
            VLA-Adapter
            \cite{wang2025vla}
            & Pre-training
            & IL
            & ActionQuery + Bridge-Attention + action module
            & Bridge Attention mechanism, vision-language feature injection into actions. \\
            
            ReSET
            \cite{dai2025prepare}
            & Pre-training
            & IL
            & scoring network + flow generator + action policy
            & Human-video-guided initial-state restructuring with a flow-based reduction policy. \\
            
            RoboMamba
            \cite{liu2024robomamba}
            & Multi-stage training
            & IL
            & vision encoder + LLM backbone + action head
            & Mamba-based efficient alignment, followed by compact policy head adaptation. \\
            
            TinyVLA
            \cite{wen2025tinyvla}
            & Pre-training
            & IL
            & vision encoder + LLM backbone + policy head
            & LoRA-based compact VLM initialization, direct policy transfer bypassing data.\\
            
            TVVE
            \cite{bai2025learning}
            & Multi-stage training
            & IL + Off-RL
            & TaskMoE + MVEP + action module
            & Pseudo-environment deployment, proactive informational perspective capturing.\\
            
            OpenVLA
            \cite{kim2024openvla}
            & Post-training
            & IL
            & VLM backbone
            & Systematic PEFT evaluation, comparative analysis. \\
            
            Atomic Skill Library
            \cite{li2025atomic}
            & Post-training
            & IL
            & skill-specific VLA policies
            & Few-shot supervised fine-tuning, atomic skill VLA adaptation architecture. \\
            
            OpenVLA-OFT
            \cite{kim2025fine}
            & Post-training
            & IL
            & VLM backbone + action head
            & Parallel decoding implementation, action chunking integration, LoRA fine-tuning. \\
            
            MoManipVLA
            \cite{wu2025momanipvla}
            & Post-training
            & IL
            & VLM backbone
            & Waypoint-guided supervised fine-tuning, mobile manipulation adaptation. \\
            
            OpenHelix
            \cite{cui2025openhelix}
            & Post-training
            & IL
            & LLM backbone + \texttt{<ACT>} token + action expert
            & Target-specific <ACT> token refinement, selective parameter updating. \\
            
            ControlVLA
            \cite{li2025controlvla}
            & Post-training
            & IL
            & additional cross-attention layers
            & Object-centric representation injection, ControlNet-style adaptation architecture. \\
            
            CronusVLA
            \cite{li2025cronusvla}
            & Post-training
            & IL
            & VLM backbone + cross-frame decoder
            & Historical frame aggregation, single-to-multi-frame decoding transition design. \\
            
            InstructVLA
            \cite{yang2025instructvla}
            & Post-training
            & IL
            & MoE adapter
            & LoRA fine-tuning, MoE-adaptation module integration. \\
            
            RICL
            \cite{sridhar2025ricl}
            & Post-training
            & IL
            & LLM backbone
            & Distance-weighted action interpolation, in-context training mechanism. \\
            
            ATE
            \cite{zhang2025align}
            & Post-training
            & IL
            & ActionVAE + guidance module
            & Latent space fine-tuning, unified latent guidance mechanism. \\
            
            
            RPD
            \cite{julg2025refined}
            & Post-training
            & On-RL
            & action module
            & MSE-guided on-policy exploration, VLA distillation into compact structures. \\
            
            RIPT-VLA
            \cite{tan2025interactive}
            & Post-training
            & On-RL
            & action head
            & Dynamic rollout sampling, leave-one-out advantage estimation. \\
            
            VLA-RL
            \cite{lu2025vla}
            & Post-training
            & On-RL
            & VLA policy + value model
            & Trajectory-level RL, auto-regressive modeling, pseudo-labeled segment rewards. \\
            
            CO-RFT
            \cite{huang2025co}
            & Post-training
            & Off-RL
            & action expert + critic network
            & Action-chunk TD learning extension, offline RL policy fine-tuning. \\
            
            ARFM
            \cite{zhang2025balancing}
            & Post-training
            & Off-RL
            & VLM backbone + action expert
            & Loss scaling factor introduction, flow matching loss modification. \\
            
            SimpleVLA-RL
            \cite{li2025simplevla}
            & Post-training
            & On-RL
            & VLM backbone + action head
            & VLA-specific trajectory sampling, parallelized training execution. \\
            
            
            RehearseVLA
            \cite{xiao2026rehearsevla}
            & Post-training
            & On-RL
            & world model + VLA Policy + reflector
            & Video-based world model deployment, resettable virtual environment. \\
            
            \bottomrule
        \end{tabular}
    }
\end{table*}

\subsubsection{Other Pre-training Strategies}
\label{subsubsec:othoerpretrainingstrategies}
Complementing data and action-centric optimizations, several specialized paradigms further bolster VLA pre-training efficiency. \textbf{Multi-stage training}~\cite{liu2024robomamba,wen2025dexvla,fan2025diffusion,ding2025humanoid,deng2025graspvla,jiang2025rynnvla} in the pre-training phase partitions the learning pipeline into sequential phases, such as modality alignment, cognitive reasoning, and action fine-tuning. MotionWAM~\cite{zheng2026motionwam} stages egocentric dynamics pre-training, cross-embodiment grounding, and whole-body adaptation. This staged decoupling enables progressive representation transfer, allowing the model to consolidate foundational perception before mastering complex physical interactions, thereby reducing aggregate computational overhead.

Strategic integration of \textbf{reinforcement learning} (RL) and \textbf{modular adaptation} further catalyzes this efficiency. To maximize data utility, GeRM~\cite{song2024germ} employs Conservative Q-Learning to ingest suboptimal trajectories, while TVVE~\cite{bai2025learning} utilize pseudo-environments to accelerate convergence by bypassing real-world trial-and-error. Parallelly, parameter-efficient fine-tuning (PEFT) via Low-Rank Adaptation (LoRA~\cite{hu2022lora}) is adopted in~\cite{zhang2025hirt, wen2025tinyvla} to inject lightweight adapters into frozen VLM backbones. This approach preserves the potent multimodal priors of large-scale models while minimizing the gradient updates required for action integration, yielding resource-aware VLAs without compromising performance.

\subsection{Efficient Post-Training}
\label{subsec:efficientposttraining}

As illustrated in~\cref{fig:efficienttraining} right, these methods encompass supervised fine-tuning for targeted parameter updates and reinforcement learning-based approaches for policy optimization. Notably, efficient post-training adapts pretrained VLAs to downstream embodiments and tasks while reducing parameter, data, interaction, or optimization costs.

\subsubsection{Supervised Fine-tuning}
\label{subsubsec:supervisedfinetuning}
Supervised Fine-Tuning (SFT) specializes pre-trained VLAs for downstream tasks, balancing task-specific proficiency with the preservation of multimodal priors. Systematic benchmarking in OpenVLA PEFT~\cite{kim2024openvla} identifies LoRA as the optimal performance-compute trade-off. OpenVLA-OFT~\cite{kim2025fine} couples LoRA with parallel chunk decoding, InstructVLA~\cite{yang2025instructvla} with MoE adapters, and Omaisan et al.~\cite{omaisan2025towards} with 4-bit quantization for 8GB consumer GPUs; LoRA-SP~\cite{kim2026adaptive} further replaces fixed ranks with input- and layer-adaptive capacity.

To overcome the dependency on extensive labeled data, recent research emphasizes \textbf{sample-efficient adaptation} through structural and contextual priors. Methods such as ControlVLA~\cite{li2025controlvla} and MoManipVLA~\cite{wu2025momanipvla} utilize zero-initialized layers or bi-level trajectory optimization to enable successful transfer with as few as 10–50 samples. Complementarily, OpenHelix~\cite{cui2025openhelix} updates a learnable <ACT> token atop a frozen MLLM, while CronusVLA~\cite{li2025cronusvla} introduces lightweight cross-frame decoding for multi-frame adaptation. Beyond gradient-based tuning, RICL~\cite{sridhar2025ricl} adopts In-Context Learning (ICL) to retrieve task-relevant prompts, while ATE~\cite{zhang2025align} employs energy-guided latent alignment to steer policy distributions. Together, these strategies facilitate a transition from broad digital knowledge to specialized physical expertise with minimal overhead.

\subsubsection{RL-Based Methods}
\label{subsubsec:rlbasedmethod}
While SFT relies heavily on demonstration quality, reinforcement learning (RL)-based post-training provides a robust mechanism to overcome data scarcity and environmental variability. These methodologies span online paradigms that refine policies through interactive rollouts and offline frameworks that optimize policies and value functions from fixed trajectory datasets.

\noindent\textbf{Online Adaptation and Exploration.}
Online RL enhances VLA proficiency through real-time environment feedback and sophisticated reward engineering. To achieve extreme sample efficiency, RIPT-VLA~\cite{tan2025interactive} combines leave-one-out advantage estimation with PPO and sparse binary rewards, improving an SFT policy from 4\% to 97\% with only one demonstration. VLA-RL~\cite{lu2025vla} instead employs PPO with a VLM-based process reward model and a curriculum selection strategy, while SimpleVLA-RL~\cite{li2025simplevla} adopts GRPO with binary outcome rewards and exploration-enhancing rollout strategies. To reduce RL overhead, RehearseVLA~\cite{xiao2026rehearsevla} shifts exploration to virtual simulators, RPD~\cite{julg2025refined} distills VLA guidance into compact RL experts, and RL Token~\cite{xu2026rl} freezes the VLA while training lightweight actor--critic heads on its learned RL token. Furthermore, human-in-the-loop paradigms, such as the "Talk-and-Tweak" scheme~\cite{jin2025dual}, incorporate linguistic corrections into diffusion-based models, enabling rapid error recovery and multi-robot coordination.

\noindent\textbf{Offline Value Regularization.}
In contrast, offline RL focuses on maximizing the utility of archived trajectories while mitigating the distribution shift inherent in out-of-distribution (OOD) actions. Central to this axis, the offline stage of ConRFT~\cite{chen2025conrft} and CO-RFT~\cite{huang2025co} employ Cal-QL~\cite{nakamoto2023cal} to penalize overestimation of OOD actions, integrating consistency objectives or transformer-based critics to stabilize value-based learning. Building on these foundations, ARFM~\cite{zhang2025balancing} introduces adaptive scaling in flow-matching losses to curb variance and bolster perturbation robustness. Collectively, these RL-driven pipelines empower VLAs to transcend the limits of passive imitation, fostering autonomous prowess and data-thrifty adaptation.

\subsection{Discussion}
\subsubsection{Innovations}
The core of efficient training paradigms lies in establishing an embodied learning mechanism that balances training-resource efficiency with robust generalization performance across the model training lifecycle. During the pre-training stage, efficiency is primarily driven by mitigating the embodied data-silo effect through the extraction of latent skills from unannotated internet videos~\cite{ye2025latent,ding2025humanoid,wang2025unified} or anatomical priors~\cite{yang2025egovla}, combined with multi-dimensional action-space compression to shrink the policy search space and accelerate convergence. In the subsequent post-training phase, downstream adaptation is highly optimized through parameter- and sample-efficient supervised fine-tuning that drastically reduces training overhead via structural or contextual priors~\cite{kim2025fine,yang2025instructvla}, seamlessly coupled with interactive online exploration~\cite{jin2025dual} and conservative offline reinforcement learning~\cite{chen2025conrft, huang2025co} to achieve rapid task specialization while robustly safeguarding against out-of-distribution policy collapse.

\subsubsection{Limitations, Challenges and Future Works}
Despite these advances, kinematic and dynamic discrepancies between human egocentric data and robotic platforms introduce significant domain noise, degrading the transferability of cross-embodied action priors~\cite{yang2025egovla}. Furthermore, while PEFT methods (\eg LoRA) can limit adaptation capacity under restrictive trainable-parameter budgets, particularly when substantial policy reconfiguration is required~\cite{jiang2026vla}. During post-training, RL-based online methods often suffer from sample inefficiency and high gradient variance under sparse rewards~\cite{zhong2026vla}. Conversely, offline RL is susceptible to action-distribution shift and Q-value overestimation on OOD actions~\cite{huang2025co,chen2025conrft}.

Future research should prioritize physics-informed, closed-loop training paradigms. To better assimilate human egocentric data, these paradigms could explore explicitly incorporating geometric, kinematic, or physical consistency constraints into training objectives, such as a kinematic consistency loss $\mathcal{L}_{\text{kinematics}}$. Furthermore, to reduce reliance on task-specific reward engineering, future RL post-training may exploit learned progress signals, adaptive multi-teacher guidance~\cite{yuan2025more}, and hierarchical RL frameworks to systematically address the challenge of reward sparsity.



\begin{figure}
  \centering
  \includegraphics[width=\linewidth]{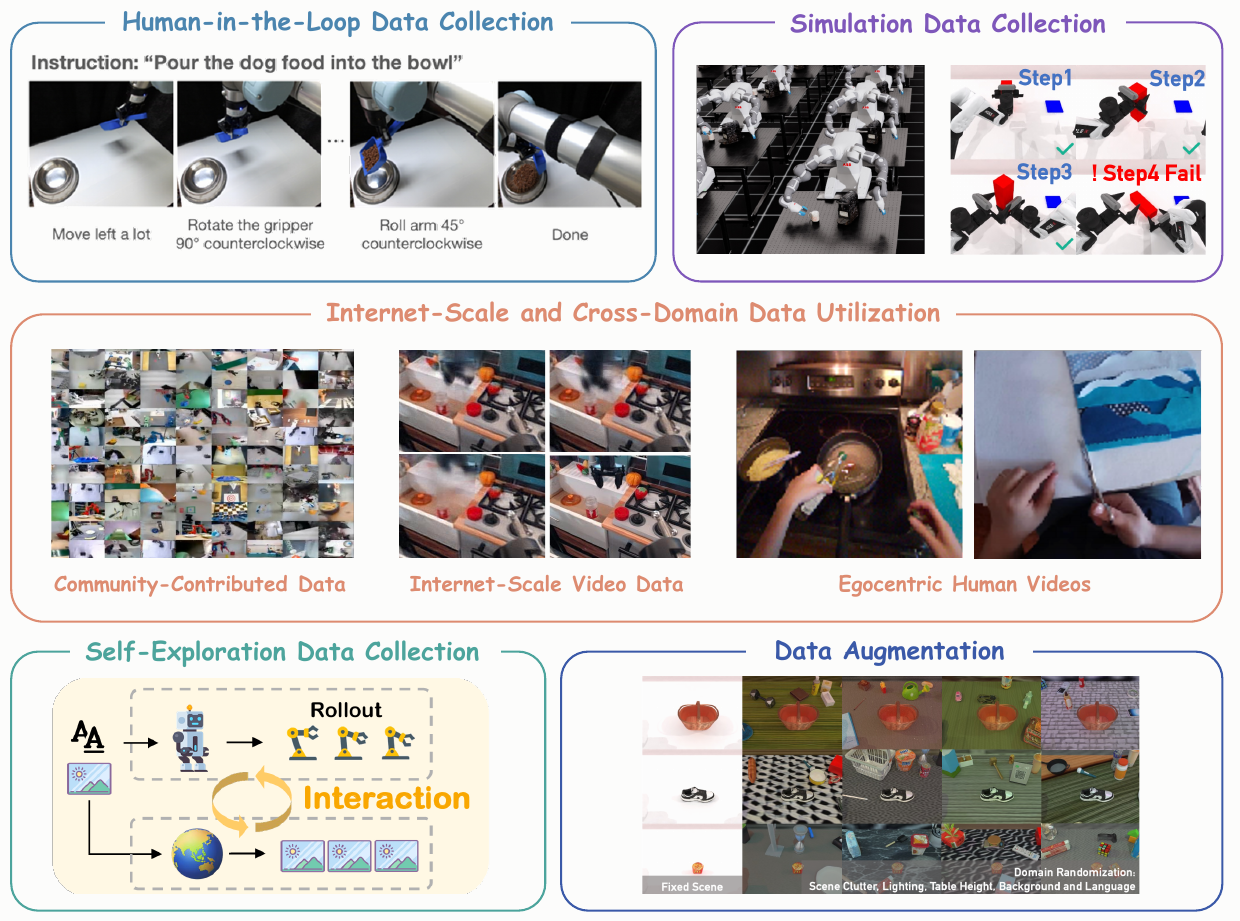}
  \caption{\textbf{Taxonomy of Efficient Data Collection strategies in VLAs.
 }This figure illustrates the primary approaches under Sec.~\ref{sec:efficientdatacollection}, encompassing human-in-the-loop, simulated, reusability-oriented,  self-driven, and augmentative techniques aimed at scaling dataset acquisition and improving data collection efficiency.}
  \label{fig:efficient data collection strategies}
\end{figure}

\section{Efficient Data Collection}
\label{sec:efficientdatacollection}

VLA performance depends on diverse, high-quality demonstrations, but data collection via human teleoperation is costly and hard to scale. \textbf{Efficient Data Collection} strives to scale up high-quality datasets at significantly reduced time and resource expenditure. As illustrated in~\cref{fig:efficient data collection strategies}, Recent efforts to overcome these challenges have pursued several strategies, including human-in-the-loop data collection (Sec.~\ref{subsec:humanintheloopdatacollection}), simulation data collection (Sec.~\ref{subsec:simulationdatacollection}), Internet-scale and cross-domain data utilization (Sec.~\ref{subsec:internetscaleandcrossdomaindataulilization}), self-exploration data collection (Sec.~\ref{subsec:selfexplorationdatacollection}), and data augmentation (Sec.~\ref{subsec:dataaugmentation}). We systematically review these strategies, analyzing their core principles and representative methods, with a concise summary provided in~\cref{table:efficiendatacollection}.

\begin{table*}[htbp]
    \scriptsize
    \setlength{\tabcolsep}{1.8pt}
    \renewcommand{\arraystretch}{1.2}
    \renewcommand{\tabularxcolumn}[1]{m{#1}}
    \newcolumntype{C}[1]{>{\centering\arraybackslash}m{#1}}
    \newcolumntype{L}[1]{>{\raggedright\arraybackslash}m{#1}}
    \newcolumntype{Y}{>{\raggedright\arraybackslash}X}
    \caption{\textbf{Representative works on Efficient Data Collection.} Categories follow Sec.~\ref{sec:efficientdatacollection}: (A) Human-in-the-loop Data Collection, (B) Simulation Data Collection, (C) Internet-Scale and Cross-Domain Data Utilization, (D) Self-Exploration Data Collection, and (E) Data Augmentation. ``Dataset Composition'' reports quantity only, and ``$-$'' means not specified. ``Quality Control'' summarizes data validation or correction. ``Collection'' Acceleration summarizes reductions in real-robot data collection.}
    \label{table:efficiendatacollection}
    \resizebox{\textwidth}{!}{
    \begin{tabular}{C{2.4cm} C{1cm} L{6.4cm} L{5cm} L{5.4cm}}
        \toprule
        \textbf{Method/Dataset}
        & \textbf{Category}
        & \textbf{Dataset Composition}
        & \textbf{Quality Control}
        & \textbf{Collection Acceleration}\\
    
        \midrule
    
        CLIP-RT~\cite{kang2024clip} & (A) & 18.1M transitions,21K in-domain transitions & STA checks; heuristic labels. & Low-skill teleop + STA expansion. \\ \hline

    
        GCENT~\cite{wang2025genie} & (A) & 48K--62K real frames. & HDF5 validation; mode labels; Task Sentinel. & Failure rewind/refine; one-to-many supervision. \\ \hline
    
        GeRM~\cite{song2024germ} & (B) & 257K simulation trajectories. & Success/failure labels; CQL penalty. & Automatic Isaac Gym rollouts. \\ \hline
    
        GraspVLA~\cite{deng2025graspvla} & (B) & 1B grasp frames. & CuRobo success; ray-traced randomization. & Parallel synthetic grasp generation. \\ \hline
    
    
        cVLA~\cite{argus2025cvla} & (B) & 150k synthetic training samples. & Execution success; visibility/manual filters. & Randomized simulation pre-training. \\ \hline
    
        RoboTwin 2.0~\cite{chen2025robotwin} & (B) & 100K+ expert trajectories. & 10 trials; log/VLM diagnosis; $\leq$5 repairs. & Code synthesis + simulation feedback. \\ \hline
    
    
        ReBot~\cite{fang2025rebot} & (B) & 220 real episodes. & Replay validation; VBench check. & Simulation replay + background inpainting. \\ \hline
    
        R2R2R~\cite{yu2025real2render2real} & (B) & 1 scan + 1 demo $\rightarrow$ 50--1000 renders/task. & 3DGS tracking; interpolation; real eval. & Single scan/demo to many renders. \\ \hline
    
        RealMirror~\cite{tai2025realmirror} & (B) & 5 scenarios; 1K+ trajectories. & Jump/jitter filter; temporal ensembling. & VR/Isaac reusable simulation demos. \\ \hline
    
        SmolVLA~\cite{shukor2025smolvla} & (C) & 481 sets; 22.9K episodes. & VLM labels; camera-view normalization. & Community-data reuse. \\ \hline
    
        EgoVLA~\cite{yang2025egovla} & (C) & $\sim$500K image-action pairs. & Wrist/MANO retargeting; downsampling. & Ego-video pre-training; few robot demos. \\ \hline
    
        RynnVLA-001~\cite{jiang2025rynnvla} & (C) & 12M egocentric + 244K robot videos; 798 demonstrations. & Keypoint filtering; Qwen2-VL captions. & Web ego-video pre-training; small teleop set. \\ \hline
    
        EgoScaler~\cite{yoshida2025developing} & (C) & 45,157 egocentric episodes; 800 robot episodes. & GPT-4o parsing; BGTS/distance filters. & Annotation-free ego-video extraction. \\ \hline
    
        Being-H0~\cite{luo2025being} & (C) & 150M+ samples; 1,000+ hours. & MANO alignment; GRQ tokens. & Hand-video physical tuning. \\ \hline
    

        EMMA~\cite{dong2025emma} & (C) & 3 tasks; 20--50 seed demos. & Depth/view/text filters; AdaMix. & Text-guided video variant synthesis. \\ \hline
    
        Humanoid-VLA~\cite{ding2025humanoid} & (C) & 0.929M clips; 557.3M frames; 7,790 hours. & Mocap quality; self-supervision. & Self-supervised motion expansion. \\ \hline

        AnyPos~\cite{tan2025anypos} & (D) & 610K pairs / 638 trajectories. & Uniform coverage; PPO feasibility. & 10h autonomous exploration. \\ \hline
    
        SimpleVLA-RL~\cite{li2025simplevla} & (D) & 1--1000 demonstrations/task; simulation rollouts. & Binary reward; dynamic sampling. & Simulation RL rollout generation. \\ \hline
    
        DiffusionRLVLA~\cite{yang2025beyond} & (D) & 130 tasks; 50 trajectories/task. & Efficiency, smoothness, variance checks. & BC-warmed diffusion RL demos. \\ \hline

        World-Env~\cite{xiao2025world} & (D) & 5 demonstrations/task + self-explored rollouts. & VLM reflector; & World-model virtual exploration. \\ \hline
    
        VLA-RFT~\cite{li2025vla} & (D) & - 16 rollouts/update. & L1+LPIPS reward verification. & World-model reward rollouts. \\ \hline
    
        LLaRA~\cite{li2025llara} & (E) & 0.8K/8K/80K demonstrations + six auxiliary tasks. & Rules; GPT-4 rephrasing; oracle boxes. & Conversational/auxiliary relabeling. \\ \hline
    
        InstructVLA~\cite{yang2025instructvla} & (E) & 650K annotated interactions. & GPT-4o/GT instruction; 95.4\% success. & Automatic instruction annotation. \\ \hline
    
        ReconVLA~\cite{song2025reconvla} & (E) & 100K+ trajectories; 2M samples. & Grounding DINO gaze; recon/action loss. & Gaze reconstruction on open data. \\ \hline
    
    
        ERMV~\cite{nie2025ermv} & (E) & $-$. & EMA-Attn/SST; expert mask correction. & 4D sequence editing; selective masks. \\
    
        \bottomrule
    \end{tabular}
    }
    \end{table*}

\subsection{Human-in-the-Loop Data Collection}
\label{subsec:humanintheloopdatacollection}

Human-in-the-loop data collection is limited by expert-dependent, one-to-one demonstrations, leading to data scarcity. Recent work shifts humans to high-level supervision to improve scalability. CLIP-RT \cite{kang2024clip} facilitates collection via natural language interfaces, yet still requires constant human presence. HiFi-UMI~\cite{simple2026hifi} improves robot-free UMI capture fidelity for deployable demonstrations. Moving toward greater autonomy, GCENT \cite{wang2025genie} employs a "guardian" strategy, where humans intervene only via rewind-and-correction during failures. This selective intervention allows a single operator to manage multiple robots, significantly enhancing collection throughput.

\subsection{Simulation Data Collection}
\label{subsec:simulationdatacollection}

Human-in-the-loop optimizations improve efficiency but still yield limited, homogeneous data. In contrast, simulation enables automated, parallel generation of diverse large-scale datasets across varied settings and embodiments at far lower cost.

To bypass the constraints of physical collection, recent research leverages massively parallel simulation for data synthesis. GraspVLA \cite{deng2025graspvla}, GeRM \cite{song2024germ}, and cVLA \cite{argus2025cvla} utilize modular experts, pre-trained policies, or privileged analytical models to generate billion-frame or multi-modal datasets autonomously. InternData-A1~\cite{tian2025interndata} extends this idea to generalist pre-training with high-fidelity synthetic trajectories across skills, scenes, and embodiments. Advancing this, RoboTwin 2.0 \cite{chen2025robotwin} employs a code-generation agent for closed-loop task synthesis and robot-specific motion planning. DynamicVLA~\cite{xie2026dynamicvla} further targets dynamic-object manipulation through automatic synthetic collection plus a small non-teleoperated real set. To mitigate the sim-to-real gap, these works employ domain randomization and photorealistic rendering.

Further fidelity is achieved via real-anchored paradigms: ReBot \cite{fang2025rebot} replays authentic trajectories in simulated scenes with real-background inpainting, while R2R2R \cite{yu2025real2render2real} converts human videos into 3D assets to track complex trajectories. Alternatively, RealMirror \cite{tai2025realmirror} bridges the gap online by integrating WebXR teleoperation with physics-filtered simulation. Together, these methods transition from pure synthesis to reality-grounded pipelines, enabling the scale required for robust VLA training.

\subsection{Internet-Scale and Cross-Domain Data Utilization}
\label{subsec:internetscaleandcrossdomaindataulilization}

Recent work leverages internet-scale robot and egocentric video data to reduce sim-to-real gaps and dataset costs. SmolVLA \cite{shukor2025smolvla} shows that standardizing noisy multi-source robot data with VLM-generated descriptions and viewpoint normalization enables state-of-the-art performance using much less data.

Beyond robotic data, leveraging human egocentric videos offers a more abundant resource. EgoVLA \cite{yang2025egovla} treats humans as a specific "embodiment," while RynnVLA-001 \cite{jiang2025rynnvla} automates this acquisition through pose-estimation-based curation. To bridge the embodiment gap, EgoScaler \cite{yoshida2025developing} and Being-H0 \cite{luo2025being} extract 6-DoF trajectories and fine-grained hand poses to convert passive observations into actionable signals. RoboWheel~\cite{zhang2025robowheel} adds contact-rich reconstruction and cross-embodiment retargeting to turn human videos into robotic supervision. Further, MimicDreamer \cite{li2025mimicdreamer} and DreamTransfer \cite{dong2025emma} employ diffusion models to synthesize photorealistic robot demonstrations from human videos, aiming to improve visual and kinematic alignment. Finally, HumanoidVLA \cite{ding2025humanoid} extends this scope to third-person videos by decomposing poses into structured tokens, further mitigating data scarcity through self-supervision.

\subsection{Self-Exploration Data Collection}
\label{subsec:selfexplorationdatacollection}

Beyond passive data acquisition, autonomous exploration transforms data collection into a self-improving process. Initially, AnyPos \cite{tan2025anypos} use RL to autonomously map the robot's reachable workspace before task-specific learning. Building on this, SimpleVLA-RL \cite{li2025simplevla} utilizes online RL to turn the VLA into a trajectory generator. To enhance exploration quality, Yang et al. \cite{yang2025beyond} replace standard policies with diffusion models, generating synthetic data that improves over original demonstrations. To further reduce interaction costs, exploration is migrating toward virtual environments.RehearseVLA \cite{xiao2026rehearsevla} leverages high-fidelity simulators for economic data augmentation; its generalizability is limited by predefined environments. Addressing this, VLA-RFT \cite{li2025vla} learns a data-driven world model from offline data. This evolution shifts VLA training from passive accumulation to fully active, scalable generation.

\subsection{Data Augmentation}
\label{subsec:dataaugmentation}

Data augmentation enhances VLA efficiency by maximizing the utility of existing trajectories through semantic, temporal, and visual transformations. Semantic enrichment methods, such as LLaRA \cite{li2025llara} and InstructVLA \cite{yang2025instructvla}, use LLMs (\eg GPT-4o) to automate hierarchical annotations and auxiliary tasks, preventing the degradation of pre-trained VLM capabilities. Similarly, RoboChemist \cite{zhang2025robochemist} diversifies instructions and visual prompts, while ReconVLA \cite{song2025reconvla} generates reconstruction objectives via automated region segmentation.

Beyond semantics, augmentation targets state-space diversity and visual appearance. Seeing Realism from Simulation~\cite{hui2026seeing} narrows the visual gap of simulated trajectories via realistic video transfer while preserving task and action information. Finally, ERMV \cite{nie2025ermv} manipulates the visual modality by propagating targeted frame edits across 4D sequences using motion-aware attention, synthesizing new training views without requiring new physical demonstrations.

\subsection{Discussion}
\label{subsec:discussion}

\subsubsection{Innovations}
Efficient data collection for VLAs is moving from one-to-one teleoperation toward reusable data engines that combine human supervision, simulation, cross-domain videos, autonomous exploration, and augmentation. Human-in-the-loop systems reduce expert burden by replacing continuous teleoperation with language-conditioned interaction or selective rewind-and-refine correction~\cite{kang2024clip,wang2025genie}. Simulation-based pipelines then scale collection through parallel expert generation, code-driven task synthesis, and robot-specific motion planning~\cite{deng2025graspvla,song2024germ,argus2025cvla,chen2025robotwin}, while real-anchored pipelines use real trajectories, human videos, or online teleoperation to improve visual and physical fidelity~\cite{fang2025rebot,yu2025real2render2real,tai2025realmirror}. A complementary line standardizes heterogeneous robot data or converts human and egocentric videos into action-relevant supervision~\cite{shukor2025smolvla,yang2025egovla,jiang2025rynnvla,yoshida2025developing,luo2025being,li2025mimicdreamer,ding2025humanoid}. Finally, self-exploration and augmentation close the loop by collecting new trajectories from policy interaction and enriching existing data with semantic, spatial, temporal, or visual transformations~\cite{tan2025anypos,yang2025beyond,li2025llara,yang2025instructvla,nie2025ermv}.

\subsubsection{Limitations, Challenges and Future Works}
Despite these advances, data acquisition remains constrained by a fundamental fidelity-scalability trade-off. Real-world demonstrations offer high-fidelity physical grounding at prohibitive costs, whereas scalable synthetic and video datasets suffer from domain shifts across embodiments, contact dynamics, and viewpoints~\cite{wang2026vision,o2024open,walke2023bridgedata,wu2026pragmatic,chen2025robotwin,hui2026seeing}. While internet and egocentric videos mitigate data scarcity, their absence of force, proprioception, and actionable annotations introduces kinematic ambiguities that generative datasets cannot fully resolve~\cite{yang2025egovla,yoshida2025developing,luo2025being,li2025mimicdreamer,ding2025humanoid}. Additionally, autonomous exploration introduces a safety-efficiency dilemma: maximizing state coverage risks hardware collisions and noisy reward signals, which degrade policy optimization~\cite{tan2025anypos,li2025simplevla,yang2025beyond}.

Therefore, future research must prioritize robust data infrastructures integrating standardized multimodal schemas, uncertainty-aware filtering, physics-grounded sim-to-real validation, and closed-loop engines that co-optimize collection efficiency, safety, and downstream policy performance~\cite{wang2026vision,wang2025genie,chen2025robotwin,tai2025realmirror,hui2026seeing}.



\section{Applications}
\label{sec:applications}
Efficiency bridges VLA intelligence and physical-world deployment. In latency-critical applications such as autonomous driving and mobile manipulation, \textbf{Efficient Model Design} reduces perception--reasoning--action overhead through compact architectures, token reduction, and fast decoding~\cite{cao2025fastdrivevla,zhou2025autovla,zhang2026mvlad,shukor2025smolvla,budzianowski2025edgevla}. In data-scarce applications such as household and industrial robotics, \textbf{Efficient Training} adapts pretrained VLAs to new tasks and embodiments with lower tuning cost~\cite{wu2025momanipvla,li2025controlvla,jiang2026vla}. When open-world deployment must cover diverse tasks, embodiments, and environments, \textbf{Efficient Data Collection} reduces costly teleoperation by expanding reusable supervision through standardized datasets, simulation, egocentric videos, self-exploration, and generative augmentation~\cite{o2024open,chen2025robotwin,yang2025egovla,deng2025graspvla,tan2025anypos}. We therefore select several representative application directions for more detailed discussion.


\subsection{Intelligent Vehicles and Autonomous Driving}
\label{subsec:intelligent vehicles and autonomous driving}
Autonomous driving uses VLAs to map multimodal traffic observations and high-level intent into trajectories or control actions. The central efficiency constraint is real-time reasoning under onboard compute limits, where dense scenes and long-tail hazards leave little tolerance for slow deliberation or unstable action decoding~\cite{zhou2025autovla,cao2025fastdrivevla,zhang2026mvlad}. Recent methods address this through adaptive reasoning and reinforcement fine-tuning~\cite{zhou2025autovla,luo2025adathinkdrive}, diffusion-based planning~\cite{jiang2025diffvla}, token pruning~\cite{cao2025fastdrivevla}, and masked VLA diffusion for efficient and explainable planning~\cite{zhang2026mvlad}.


\subsection{Mobile Manipulation and Smart Home}
\label{subsec:mobile manipulation and smart home}
Mobile manipulation and smart-home robotics require VLAs to navigate indoor spaces, manipulate objects, and follow open-vocabulary instructions. Efficiency is limited by edge deployment, privacy-preserving inference, long-horizon base-arm coordination, and sparse household demonstrations~\cite{miao2025fedvla}. Recent work responds with compact VLA backbones and asynchronous inference~\cite{shukor2025smolvla}, faster edge-oriented action decoding~\cite{budzianowski2025edgevla}, transfer to mobile manipulation~\cite{wu2025momanipvla}, and spatially grounded training with depth, multi-view, and temporal supervision~\cite{tu2026sgvla}.


\subsection{Industrial Manufacturing and Logistics}
\label{subsec:industrial manufacturing and logistics}
Industrial manufacturing and logistics involve VLA-based manipulation, inspection, assembly, disassembly, process control, and warehouse operation. The efficiency challenge shifts from single-task success to throughput, continuous-run reliability, precision placement, and contact-rich control~\cite{li2025transferring,liu2026selfvla,robotics2026habilis}. Recent studies evaluate VLA transfer under industrial constraints~\cite{li2025transferring}, use hybrid reasoning for adaptive manufacturing perception and process control~\cite{margadji2025hybrid}, add explicit skills for contact-rich disassembly~\cite{liu2026selfvla}, and report on-device humanoid logistics with productivity and reliability metrics~\cite{robotics2026habilis}.


\subsection{Human-Robot Collaboration in Medical Assistance}
\label{subsec:human-robot collaboration in medical assistance}
Human-robot collaboration and medical assistance require VLAs to infer intent, support context-aware multimodal interaction~\cite{zhang2026omnicharacter++}, and execute assistive actions such as tool handover or surgical support. Their efficiency bottlenecks include real-time interaction, limited user-specific data, and safety-critical recovery~\cite{li2024robonurse,sun2025collabvla}. Recent work studies real-time surgical instrument handover~\cite{li2024robonurse} and reflective collaboration with mixture-of-experts action generation~\cite{sun2025collabvla}.


\section{Conclusion}
\label{sec:conclusion}

This survey systematically charts the emergent field of \textbf{Efficient VLAs}, providing the first unified taxonomy focusing on the model-training-data loop to consolidate the fragmented efforts addressing their prohibitive computational and data bottlenecks. We demonstrate that current solutions converge upon three interdependent pillars, including \textbf{Efficient Model Design}, \textbf{Efficient Training}, and \textbf{Efficient Data Collection}. Our synthesis uncovers the intrinsic trade-offs between model compactness and multimodal expressivity, providing a structured perspective for understanding the design tensions that characterize this space.  The challenges and future directions delineated are a direct consequence of this synthesis, charting a necessary roadmap that moves beyond isolated optimizations toward adaptive, co-designed systems. This work thus serves to support this transition, contributing to the development from resource-bound prototypes toward deployable physical-world intelligence.

\bibliographystyle{IEEEtran}
\bibliography{ref}

@string(AAAI  = "{AAAI Conference on Artificial Intelligence}")

@string(CVPR  = "{IEEE Conference on Computer Vision and Pattern Recognition}")

@string(ICLR  = "{International Conference on Learning Representations}")

@string(AAAI  = "{AAAI}")

@string(CVPR  = "{CVPR}")

@string(ICLR  = "{ICLR}")

@article{ma2024survey,
  title={A survey on vision--language--action models for embodied ai},
  author={Ma, Yueen and Song, Zixing and Zhuang, Yuzheng and Hao, Jianye and King, Irwin},
  journal={IEEE Transactions on Neural Networks and Learning Systems},
  year={2026},
  publisher={IEEE}
}

@inproceedings{cao2025fastdrivevla,
  title={Fastdrivevla: Efficient end-to-end driving via plug-and-play reconstruction-based token pruning},
  author={Cao, Jiajun and Zhang, Qizhe and Jia, Peidong and Zhao, Xuhui and Lan, Bo and Zhang, Xiaoan and Wei, Xiaobao and Chen, Sixiang and Li, Liyun and Liu, Xianming and others},
  booktitle={Proceedings of the AAAI Conference on Artificial Intelligence},
  volume={40},
  number={4},
  pages={2571--2579},
  year={2026}
}

@inproceedings{zhou2025opendrivevla,
  title={Opendrivevla: Towards end-to-end autonomous driving with large vision language action model},
  author={Zhou, Xingcheng and Han, Xuyuan and Yang, Feng and Ma, Yunpu and Tresp, Volker and Knoll, Alois},
  booktitle={Proceedings of the AAAI Conference on Artificial Intelligence},
  volume={40},
  number={16},
  pages={13782--13790},
  year={2026}
}

@inproceedings{li2024robonurse,
  title={Robonurse-vla: Robotic scrub nurse system based on vision-language-action model},
  author={Li, Shunlei and Wang, Jin and Dai, Rui and Ma, Wanyu and Ng, Wing Yin and Hu, Yingbai and Li, Zheng},
  booktitle={2025 IEEE/RSJ International Conference on Intelligent Robots and Systems (IROS)},
  pages={3986--3993},
  year={2025},
  organization={IEEE}
}

@inproceedings{zhang2025robochemist,
  title={RoboChemist: Long-Horizon and Safety-Compliant Robotic Chemical Experimentation},
  author={Zhang, Zongzheng and Yue, Chenghao and Xu, Haobo and Liao, Minwen and Qi, Xianglin and Gao, Huan-ang and Wang, Ziwei and Zhao, Hao},
  booktitle={Conference on Robot Learning},
  pages={3537--3568},
  year={2025},
  organization={PMLR}
}

@article{black2024pi_0,
  title={$\pi_0 $: A Vision-Language-Action Flow Model for General Robot Control},
  author={Black, Kevin and Brown, Noah and Driess, Danny and Esmail, Adnan and Equi, Michael and Finn, Chelsea and Fusai, Niccolo and Groom, Lachy and Hausman, Karol and Ichter, Brian and others},
  journal={arXiv preprint arXiv:2410.24164},
  year={2024}
}

@inproceedings{kim2024openvla,
  title={OpenVLA: An Open-Source Vision-Language-Action Model},
  author={Kim, Moo Jin and Pertsch, Karl and Karamcheti, Siddharth and Xiao, Ted and Balakrishna, Ashwin and Nair, Suraj and Rafailov, Rafael and Foster, Ethan P and Sanketi, Pannag R and Vuong, Quan and others},
  booktitle={Conference on Robot Learning},
  pages={2679--2713},
  year={2025},
  organization={PMLR}
}

@article{li2024cogact,
  title={Cogact: A foundational vision-language-action model for synergizing cognition and action in robotic manipulation},
  author={Li, Qixiu and Liang, Yaobo and Wang, Zeyu and Luo, Lin and Chen, Xi and Liao, Mozheng and Wei, Fangyun and Deng, Yu and Xu, Sicheng and Zhang, Yizhong and others},
  journal={arXiv preprint arXiv:2411.19650},
  year={2024}
}

@inproceedings{zitkovich2023rt,
  title={Rt-2: Vision-language-action models transfer web knowledge to robotic control},
  author={Zitkovich, Brianna and Yu, Tianhe and Xu, Sichun and Xu, Peng and Xiao, Ted and Xia, Fei and Wu, Jialin and Wohlhart, Paul and Welker, Stefan and Wahid, Ayzaan and others},
  booktitle={Conference on Robot Learning},
  pages={2165--2183},
  year={2023},
  organization={PMLR}
}

@article{beyer2024paligemma,
  title={Paligemma: A versatile 3b vlm for transfer},
  author={Beyer, Lucas and Steiner, Andreas and Pinto, Andr{\'e} Susano and Kolesnikov, Alexander and Wang, Xiao and Salz, Daniel and Neumann, Maxim and Alabdulmohsin, Ibrahim and Tschannen, Michael and Bugliarello, Emanuele and others},
  journal={arXiv preprint arXiv:2407.07726},
  year={2024}
}

@inproceedings{karamcheti2024prismatic,
  title={Prismatic VLMs: investigating the design space of visually-conditioned language models},
  author={Karamcheti, Siddharth and Nair, Suraj and Balakrishna, Ashwin and Liang, Percy and Kollar, Thomas and Sadigh, Dorsa},
  booktitle={Proceedings of the 41st International Conference on Machine Learning},
  pages={23123--23144},
  year={2024}
}

@article{bjorck2025gr00t,
  title={Gr00t n1: An open foundation model for generalist humanoid robots},
  author={Bjorck, Johan and Casta{\~n}eda, Fernando and Cherniadev, Nikita and Da, Xingye and Ding, Runyu and Fan, Linxi and Fang, Yu and Fox, Dieter and Hu, Fengyuan and Huang, Spencer and others},
  journal={arXiv preprint arXiv:2503.14734},
  year={2025}
}

@article{guo2026thousand,
  title={Thousand-gpu large-scale training and optimization recipe for ai-native cloud embodied intelligence infrastructure},
  author={Guo, Yongjian and Ma, Yunxuan and Sun, Haoran and Guan, Zhong and Di, Shuai and Long, Jing and Xu, Wanting and Bai, Xiaodong and Huang, Wen and Guo, Yucheng and others},
  journal={arXiv preprint arXiv:2603.11101},
  year={2026}
}

@inproceedings{yang2025instructvla,
  title={Vision-language-action instruction tuning: From understanding to manipulation},
  author={Yang, Shuai and Li, Hao and Wang, Bin and Chen, Yilun and Tian, Yang and Wang, Tai and Wang, Hanqing and Zhao, Feng and Liao, Yiyi and Pang, Jiangmiao},
  booktitle={International Conference on Learning Representations},
  volume={2026},
  pages={152896--152943},
  year={2026}
}

@inproceedings{li2025simplevla,
  title={Simplevla-rl: Scaling vla training via reinforcement learning},
  author={Li, Haozhan and Zuo, Yuxin and Yu, Jiale and Zhang, Yuhao and Zhaohui, Yang and Zhang, Kaiyan and Zhu, Xuekai and Zhang, Yuchen and Chen, Tianxing and Cui, Ganqu and others},
  booktitle={International Conference on Learning Representations},
  volume={2026},
  pages={125846--125869},
  year={2026}
}

@article{bu2025agibot,
  title={Agibot world colosseo: A large-scale manipulation platform for scalable and intelligent embodied systems},
  author={Bu, Qingwen and Cai, Jisong and Chen, Li and Cui, Xiuqi and Ding, Yan and Feng, Siyuan and Gao, Shenyuan and He, Xindong and Hu, Xuan and Huang, Xu and others},
  journal={arXiv preprint arXiv:2503.06669},
  year={2025}
}

@inproceedings{zhang2025mole,
  title={Mole-vla: Dynamic layer-skipping vision language action model via mixture-of-layers for efficient robot manipulation},
  author={Zhang, Rongyu and Dong, Menghang and Zhang, Yuan and Heng, Liang and Chi, Xiaowei and Dai, Gaole and Du, Li and Wang, Dan and Du, Yuan and Zhang, Shanghang},
  booktitle={Proceedings of the AAAI Conference on Artificial Intelligence},
  volume={40},
  number={22},
  pages={18764--18772},
  year={2026}
}

@article{yang2025efficientvla,
  title={Efficientvla: Training-free acceleration and compression for vision-language-action models},
  author={Yang, Yantai and Wang, Yuhao and Wen, Zichen and Zhongwei, Luo and Zou, Chang and Zhang, Zhipeng and Wen, Chuan and Zhang, Linfeng},
  journal={Advances in Neural Information Processing Systems},
  volume={38},
  pages={40891--40914},
  year={2026}
}

@article{pertsch2025fast,
  title={Fast: Efficient action tokenization for vision-language-action models},
  author={Pertsch, Karl and Stachowicz, Kyle and Ichter, Brian and Driess, Danny and Nair, Suraj and Vuong, Quan and Mees, Oier and Finn, Chelsea and Levine, Sergey},
  journal={arXiv preprint arXiv:2501.09747},
  year={2025}
}

@article{shukor2025smolvla,
  title={Smolvla: A vision-language-action model for affordable and efficient robotics},
  author={Shukor, Mustafa and Aubakirova, Dana and Capuano, Francesco and Kooijmans, Pepijn and Palma, Steven and Zouitine, Adil and Aractingi, Michel and Pascal, Caroline and Russi, Martino and Marafioti, Andres and others},
  journal={arXiv preprint arXiv:2506.01844},
  year={2025}
}

@article{liu2024robomamba,
  title={Robomamba: Efficient vision-language-action model for robotic reasoning and manipulation},
  author={Liu, Jiaming and Liu, Mengzhen and Wang, Zhenyu and An, Pengju and Li, Xiaoqi and Zhou, Kaichen and Yang, Senqiao and Zhang, Renrui and Guo, Yandong and Zhang, Shanghang},
  journal={Advances in Neural Information Processing Systems},
  volume={37},
  pages={40085--40110},
  year={2024}
}

@article{kim2025fine,
  title={Fine-tuning vision-language-action models: Optimizing speed and success},
  author={Kim, Moo Jin and Finn, Chelsea and Liang, Percy},
  journal={arXiv preprint arXiv:2502.19645},
  year={2025}
}

@inproceedings{leal2024sara,
  title={Sara-rt: Scaling up robotics transformers with self-adaptive robust attention},
  author={Leal, Isabel and Choromanski, Krzysztof and Jain, Deepali and Dubey, Avinava and Varley, Jake and Ryoo, Michael and Lu, Yao and Liu, Frederick and Sindhwani, Vikas and Vuong, Quan and others},
  booktitle={2024 IEEE International Conference on Robotics and Automation (ICRA)},
  pages={6920--6927},
  year={2024},
  organization={IEEE}
}

@article{chen2023pali,
  title={Pali-x: On scaling up a multilingual vision and language model},
  author={Chen, Xi and Djolonga, Josip and Padlewski, Piotr and Mustafa, Basil and Changpinyo, Soravit and Wu, Jialin and Ruiz, Carlos Riquelme and Goodman, Sebastian and Wang, Xiao and Tay, Yi and others},
  journal={arXiv preprint arXiv:2305.18565},
  year={2023}
}

@inproceedings{driess2023palm,
  title={PaLM-E: An Embodied Multimodal Language Model},
  author={Driess, Danny and Xia, Fei and Sajjadi, Mehdi SM and Lynch, Corey and Chowdhery, Aakanksha and Ichter, Brian and Wahid, Ayzaan and Tompson, Jonathan and Vuong, Quan and Yu, Tianhe and others},
  booktitle={International Conference on Machine Learning},
  pages={8469--8488},
  year={2023},
  organization={PMLR}
}

@inproceedings{zhai2023sigmoid,
  title={Sigmoid loss for language image pre-training},
  author={Zhai, Xiaohua and Mustafa, Basil and Kolesnikov, Alexander and Beyer, Lucas},
  booktitle={Proceedings of the IEEE/CVF international conference on computer vision},
  pages={11975--11986},
  year={2023}
}

@article{oquab2023dinov2,
  title={DINOv2: Learning Robust Visual Features without Supervision},
  author={Oquab, Maxime and Darcet, Timoth{\'e}e and Moutakanni, Th{\'e}o and Vo, Huy and Szafraniec, Marc and Khalidov, Vasil and Fernandez, Pierre and Haziza, Daniel and Massa, Francisco and El-Nouby, Alaaeldin and others},
  journal={Transactions on Machine Learning Research Journal},
  year={2024}
}

@article{touvron2023llama,
  title={Llama 2: Open foundation and fine-tuned chat models},
  author={Touvron, Hugo and Martin, Louis and Stone, Kevin and Albert, Peter and Almahairi, Amjad and Babaei, Yasmine and Bashlykov, Nikolay and Batra, Soumya and Bhargava, Prajjwal and Bhosale, Shruti and others},
  journal={arXiv preprint arXiv:2307.09288},
  year={2023}
}

@inproceedings{o2024open,
  title={Open x-embodiment: Robotic learning datasets and rt-x models: Open x-embodiment collaboration 0},
  author={O’Neill, Abby and Rehman, Abdul and Maddukuri, Abhiram and Gupta, Abhishek and Padalkar, Abhishek and Lee, Abraham and Pooley, Acorn and Gupta, Agrim and Mandlekar, Ajay and Jain, Ajinkya and others},
  booktitle={2024 IEEE International Conference on Robotics and Automation (ICRA)},
  pages={6892--6903},
  year={2024},
  organization={IEEE}
}

@article{team2024gemma,
  title={Gemma: Open models based on gemini research and technology},
  author={Team, Gemma and Mesnard, Thomas and Hardin, Cassidy and Dadashi, Robert and Bhupatiraju, Surya and Pathak, Shreya and Sifre, Laurent and Rivi{\`e}re, Morgane and Kale, Mihir Sanjay and Love, Juliette and others},
  journal={arXiv preprint arXiv:2403.08295},
  year={2024}
}

@inproceedings{shi2025hi,
  title={Hi Robot: Open-Ended Instruction Following with Hierarchical Vision-Language-Action Models},
  author={Shi, Lucy Xiaoyang and Equi, Michael Robert and Ke, Liyiming and Pertsch, Karl and Vuong, Quan and Tanner, James and Walling, Anna and Wang, Haohuan and Fusai, Niccolo and Li-Bell, Adrian and others},
  booktitle={Forty-second International Conference on Machine Learning},
  year={2025}
}

@inproceedings{walke2023bridgedata,
  title={Bridgedata v2: A dataset for robot learning at scale},
  author={Walke, Homer Rich and Black, Kevin and Zhao, Tony Z and Vuong, Quan and Zheng, Chongyi and Hansen-Estruch, Philippe and He, Andre Wang and Myers, Vivek and Kim, Moo Jin and Du, Max and others},
  booktitle={Conference on Robot Learning},
  pages={1723--1736},
  year={2023},
  organization={PMLR}
}

@inproceedings{khazatsky2024droid,
  title={DROID: A Large-Scale In-The-Wild Robot Manipulation Dataset},
  author={Khazatsky, Alexander and Pertsch, Karl and Nair, Suraj and Balakrishna, Ashwin and Dasari, Sudeep and Karamcheti, Siddharth and Nasiriany, Soroush and Srirama, Mohan Kumar and Chen, Lawrence Yunliang and Ellis, Kirsty and others},
  booktitle={RSS 2024 Workshop: Data Generation for Robotics},
  year={2024}
}

@article{james2020rlbench,
  title={Rlbench: The robot learning benchmark \& learning environment},
  author={James, Stephen and Ma, Zicong and Arrojo, David Rovick and Davison, Andrew J},
  journal={IEEE Robotics and Automation Letters},
  volume={5},
  number={2},
  pages={3019--3026},
  year={2020},
  publisher={IEEE}
}

@inproceedings{nasiriany2024robocasa,
  title={RoboCasa: Large-Scale Simulation of Everyday Tasks for Generalist Robots},
  author={Nasiriany, Soroush and Maddukuri, Abhiram and Zhang, Lance and Parikh, Adeet and Lo, Aaron and Joshi, Abhishek and Mandlekar, Ajay and Zhu, Yuke},
  booktitle={RSS 2024 Workshop: Data Generation for Robotics},
  year={2024}
}

@inproceedings{wang2024robogen,
  title={RoboGen: towards unleashing infinite data for automated robot learning via generative simulation},
  author={Wang, Yufei and Xian, Zhou and Chen, Feng and Wang, Tsun-Hsuan and Wang, Yian and Fragkiadaki, Katerina and Erickson, Zackory and Held, David and Gan, Chuang},
  booktitle={Proceedings of the 41st International Conference on Machine Learning},
  pages={51936--51983},
  year={2024}
}

@article{liu2023libero,
  title={Libero: Benchmarking knowledge transfer for lifelong robot learning},
  author={Liu, Bo and Zhu, Yifeng and Gao, Chongkai and Feng, Yihao and Liu, Qiang and Zhu, Yuke and Stone, Peter},
  journal={Advances in Neural Information Processing Systems},
  volume={36},
  pages={44776--44791},
  year={2023}
}

@article{shao2025large,
  title={Large vlm-based vision-language-action models for robotic manipulation: A survey},
  author={Shao, Rui and Li, Wei and Zhang, Lingsen and Zhang, Renshan and Liu, Zhiyang and Chen, Ran and Nie, Liqiang},
  journal={arXiv preprint arXiv:2508.13073},
  year={2025}
}

@article{xiang2025parallels,
  title={Parallels Between VLA Model Post-Training and Human Motor Learning: Progress, Challenges, and Trends},
  author={Xiang, Tian-Yu and Jin, Ao-Qun and Zhou, Xiao-Hu and Gui, Mei-Jiang and Xie, Xiao-Liang and Liu, Shi-Qi and Wang, Shuang-Yi and Duan, Sheng-Bin and Xie, Fu-Chao and Wang, Wen-Kai and others},
  journal={arXiv preprint arXiv:2506.20966},
  year={2025}
}

@article{zhong2025survey,
  title={A Survey on Vision-Language-Action Models: An Action Tokenization Perspective},
  author={Zhong, Yifan and Bai, Fengshuo and Cai, Shaofei and Huang, Xuchuan and Chen, Zhang and Zhang, Xiaowei and Wang, Yuanfei and Guo, Shaoyang and Guan, Tianrui and Lui, Ka Nam and others},
  journal={arXiv preprint arXiv:2507.01925},
  year={2025}
}

@article{din2025vision,
  title={Vision language action models in robotic manipulation: A systematic review},
  author={Din, Muhayy Ud and Akram, Waseem and Saoud, Lyes Saad and Rosell, Jan and Hussain, Irfan},
  journal={arXiv preprint arXiv:2507.10672},
  year={2025}
}

@article{zhang2025pure,
  title={Pure Vision Language Action (VLA) Models: A Comprehensive Survey},
  author={Zhang, Dapeng and Sun, Jin and Hu, Chenghui and Wu, Xiaoyan and Yuan, Zhenlong and Zhou, Rui and Shen, Fei and Zhou, Qingguo},
  journal={arXiv preprint arXiv:2509.19012},
  year={2025}
}

@article{vaswani2017attention,
  title={Attention is all you need},
  author={Vaswani, Ashish and Shazeer, Noam and Parmar, Niki and Uszkoreit, Jakob and Jones, Llion and Gomez, Aidan N and Kaiser, {\L}ukasz and Polosukhin, Illia},
  journal={Advances in neural information processing systems},
  volume={30},
  year={2017}
}

@inproceedings{dosovitskiy2020image,
  title={An Image is Worth 16x16 Words: Transformers for Image Recognition at Scale},
  author={Dosovitskiy, Alexey and Beyer, Lucas and Kolesnikov, Alexander and Weissenborn, Dirk and Zhai, Xiaohua and Unterthiner, Thomas and Dehghani, Mostafa and Minderer, Matthias and Heigold, G and Gelly, S and others},
  booktitle={International Conference on Learning Representations},
  year={2020}
}

@inproceedings{radford2021learning,
  title={Learning transferable visual models from natural language supervision},
  author={Radford, Alec and Kim, Jong Wook and Hallacy, Chris and Ramesh, Aditya and Goh, Gabriel and Agarwal, Sandhini and Sastry, Girish and Askell, Amanda and Mishkin, Pamela and Clark, Jack and others},
  booktitle={International conference on machine learning},
  pages={8748--8763},
  year={2021},
  organization={PmLR}
}

@article{bai2023qwen,
  title={Qwen technical report},
  author={Bai, Jinze and Bai, Shuai and Chu, Yunfei and Cui, Zeyu and Dang, Kai and Deng, Xiaodong and Fan, Yang and Ge, Wenbin and Han, Yu and Huang, Fei and others},
  journal={arXiv preprint arXiv:2309.16609},
  year={2023}
}

@article{chowdhery2023palm,
  title={Palm: Scaling language modeling with pathways},
  author={Chowdhery, Aakanksha and Narang, Sharan and Devlin, Jacob and Bosma, Maarten and Mishra, Gaurav and Roberts, Adam and Barham, Paul and Chung, Hyung Won and Sutton, Charles and Gehrmann, Sebastian and others},
  journal={Journal of Machine Learning Research},
  volume={24},
  number={240},
  pages={1--113},
  year={2023}
}

@article{ho2020denoising,
  title={Denoising diffusion probabilistic models},
  author={Ho, Jonathan and Jain, Ajay and Abbeel, Pieter},
  journal={Advances in neural information processing systems},
  volume={33},
  pages={6840--6851},
  year={2020}
}

@inproceedings{peebles2023scalable,
  title={Scalable diffusion models with transformers},
  author={Peebles, William and Xie, Saining},
  booktitle={Proceedings of the IEEE/CVF international conference on computer vision},
  pages={4195--4205},
  year={2023}
}

@inproceedings{lipman2022flow,
  title={Flow Matching for Generative Modeling},
  author={Lipman, Yaron and Chen, Ricky TQ and Ben-Hamu, Heli and Nickel, Maximilian and Le, Matthew},
  booktitle={The Eleventh International Conference on Learning Representations},
  year={2023}
}

@inproceedings{fan2025long,
  title={Long-VLA: Unleashing Long-Horizon Capability of Vision Language Action Model for Robot Manipulation},
  author={Fan, Yiguo and Bai, Shuanghao and Tong, Xinyang and Ding, Pengxiang and Zhu, Yuyang and Lu, Hongchao and Dai, Fengqi and Zhao, Wei and Liu, Yang and Huang, Siteng and others},
  booktitle={9th Annual Conference on Robot Learning},
  year={2025}
}

@article{wen2025dvla,
  title={dVLA: Diffusion Vision-Language-Action Model with Multimodal Chain-of-Thought},
  author={Wen, Junjie and Zhu, Minjie and Liu, Jiaming and Liu, Zhiyuan and Yang, Yicun and Zhang, Linfeng and Zhang, Shanghang and Zhu, Yichen and Xu, Yi},
  journal={arXiv preprint arXiv:2509.25681},
  year={2025}
}

@article{koo2025retovla,
  title={RetoVLA: Reusing Register Tokens for Spatial Reasoning in Vision-Language-Action Models},
  author={Koo, Jiyeon and Cho, Taewan and Kang, Hyunjoon and Pyo, Eunseom and Oh, Tae Gyun and Kim, Taeryang and Choi, Andrew Jaeyong},
  journal={arXiv preprint arXiv:2509.21243},
  year={2025}
}

@article{xu2025kv,
  title={KV-Efficient VLA: A Method of Speed up Vision Language Model with RNN-Gated Chunked KV Cache},
  author={Xu, Wanshun and Zhuang, Long},
  journal={arXiv preprint arXiv:2509.21354},
  year={2025}
}

@inproceedings{gu2024mamba,
  title={Mamba: Linear-time sequence modeling with selective state spaces},
  author={Gu, Albert and Dao, Tri},
  booktitle={First Conference on Language Modeling},
  year={2024}
}

@article{budzianowski2025edgevla,
  title={Edgevla: Efficient vision-language-action models},
  author={Budzianowski, Pawe{\l} and Maa, Wesley and Freed, Matthew and Mo, Jingxiang and Hsiao, Winston and Xie, Aaron and M{\l}oduchowski, Tomasz and Tipnis, Viraj and Bolte, Benjamin},
  journal={arXiv preprint arXiv:2507.14049},
  year={2025}
}

@misc{belkhale2024minivla,
      title={MiniVLA: A Better VLA with a Smaller Footprint}, 
      author={Suneel Belkhale and Dorsa Sadigh},
      url={https://github.com/Stanford-ILIAD/openvla-mini},
      year={2024},
}

@inproceedings{song2025accelerating,
  title={Pd-vla: Accelerating vision-language-action model integrated with action chunking via parallel decoding},
  author={Song, Wenxuan and Chen, Jiayi and Ding, Pengxiang and Zhao, Han and Zhao, Wei and Zhong, Zhide and Ge, Zongyuan and Li, Zhijun and Wang, Donglin and Wang, Lujia and others},
  booktitle={2025 IEEE/RSJ International Conference on Intelligent Robots and Systems (IROS)},
  pages={13162--13169},
  year={2025},
  organization={IEEE}
}

@article{song2025ceed,
  title={CEED-VLA: Consistency Vision-Language-Action Model with Early-Exit Decoding},
  author={Song, Wenxuan and Chen, Jiayi and Ding, Pengxiang and Huang, Yuxin and Zhao, Han and Wang, Donglin and Li, Haoang},
  journal={arXiv preprint arXiv:2506.13725},
  year={2025}
}

@inproceedings{wang2025spec,
  title={Spec-vla: speculative decoding for vision-language-action models with relaxed acceptance},
  author={Wang, Songsheng and Yu, Rucheng and Yuan, Zhihang and Yu, Chao and Gao, Feng and Wang, Yu and Wong, Derek F},
  booktitle={Proceedings of the 2025 Conference on Empirical Methods in Natural Language Processing},
  pages={26916--26928},
  year={2025}
}

@article{wen2025tinyvla,
  title={Tinyvla: Towards fast, data-efficient vision-language-action models for robotic manipulation},
  author={Wen, Junjie and Zhu, Yichen and Li, Jinming and Zhu, Minjie and Tang, Zhibin and Wu, Kun and Xu, Zhiyuan and Liu, Ning and Cheng, Ran and Shen, Chaomin and others},
  journal={IEEE Robotics and Automation Letters},
  year={2025},
  publisher={IEEE}
}

@article{liu2025hybridvla,
  title={Hybridvla: Collaborative diffusion and autoregression in a unified vision-language-action model},
  author={Liu, Jiaming and Chen, Hao and An, Pengju and Liu, Zhuoyang and Zhang, Renrui and Gu, Chenyang and Li, Xiaoqi and Guo, Ziyu and Chen, Sixiang and Liu, Mengzhen and others},
  journal={arXiv preprint arXiv:2503.10631},
  year={2025}
}

@article{su2025freqpolicy,
  title={Freqpolicy: Efficient flow-based visuomotor policy via frequency consistency},
  author={Su, Yifei and Liu, Ning and Chen, Dong and Zhao, Zhen and Wu, Kun and Li, Meng and Xu, Zhiyuan and Che, Zhengping and Tang, Jian},
  journal={Advances in Neural Information Processing Systems},
  volume={38},
  pages={27769--27797},
  year={2026}
}

@inproceedings{wang2025flowram,
  title={FlowRAM: Grounding Flow Matching Policy with Region-Aware Mamba Framework for Robotic Manipulation},
  author={Wang, Sen and Wang, Le and Zhou, Sanping and Tian, Jingyi and Li, Jiayi and Sun, Haowen and Tang, Wei},
  booktitle={Proceedings of the Computer Vision and Pattern Recognition Conference},
  pages={12176--12186},
  year={2025}
}

@inproceedings{wang2025vq,
  title={Vq-vla: Improving vision-language-action models via scaling vector-quantized action tokenizers},
  author={Wang, Yating and Zhu, Haoyi and Liu, Mingyu and Yang, Jiange and Fang, Hao-Shu and He, Tong},
  booktitle={Proceedings of the IEEE/CVF International Conference on Computer Vision},
  pages={11089--11099},
  year={2025}
}

@article{van2017neural,
  title={Neural discrete representation learning},
  author={Van Den Oord, Aaron and Vinyals, Oriol and others},
  journal={Advances in neural information processing systems},
  volume={30},
  year={2017}
}

@article{zheng2025leveraging,
  title={Leveraging OS-Level Primitives for Robotic Action Management},
  author={Zheng, Wenxin and Li, Boyang and Xu, Bin and Feng, Erhu and Gu, Jinyu and Chen, Haibo},
  journal={arXiv preprint arXiv:2508.10259},
  year={2025}
}

@inproceedings{tarasov2025nina,
  title={NinA: Normalizing Flows in Action. Training VLA Models with Normalizing Flows},
  author={Tarasov, Denis and Nikulin, Alexander and Zisman, Ilya and Klepach, Albina and Nikita, Lyubaykin and Polubarov, Andrei and Derevyagin, Alexander and Kurenkov, Vladislav},
  booktitle={NeurIPS 2025 Workshop on Embodied World Models for Decision Making},
  year={2025}
}

@article{liang2025discrete,
  title={Discrete diffusion vla: Bringing discrete diffusion to action decoding in vision-language-action policies},
  author={Liang, Zhixuan and Li, Yizhuo and Yang, Tianshuo and Wu, Chengyue and Mao, Sitong and Pei, Liuao and Yang, Xiaokang and Pang, Jiangmiao and Mu, Yao and Luo, Ping},
  journal={arXiv preprint arXiv:2508.20072},
  year={2025}
}

@article{kang2024clip,
  title={CLIP-RT: Learning Language-Conditioned Robotic Policies from Natural Language Supervision},
  author={Kang, Gi-Cheon and Kim, Junghyun and Shim, Kyuhwan and Lee, Jun Ki and Zhang, Byoung-Tak},
  journal={arXiv preprint arXiv:2411.00508},
  year={2024}
}

@article{wen2024diffusion,
  title={Diffusion-VLA: Generalizable and Interpretable Robot Foundation Model via Self-Generated Reasoning},
  author={Wen, Junjie and Zhu, Minjie and Zhu, Yichen and Tang, Zhibin and Li, Jinming and Zhou, Zhongyi and Li, Chengmeng and Liu, Xiaoyu and Peng, Yaxin and Shen, Chaomin and others},
  journal={arXiv preprint arXiv:2412.03293},
  year={2024}
}

@inproceedings{samson2025scalable,
  title={Scalable, Training-Free Visual Language Robotics: a modular multi-model framework for consumer-grade GPUs},
  author={Samson, Marie and Muraccioli, Bastien and Kanehiro, Fumio},
  booktitle={2025 IEEE/SICE International Symposium on System Integration (SII)},
  pages={193--198},
  year={2025},
  organization={IEEE}
}

@article{hung2025nora,
  title={Nora: A small open-sourced generalist vision language action model for embodied tasks},
  author={Hung, Chia-Yu and Sun, Qi and Hong, Pengfei and Zadeh, Amir and Li, Chuan and Tan, U and Majumder, Navonil and Poria, Soujanya and others},
  journal={arXiv preprint arXiv:2504.19854},
  year={2025}
}

@inproceedings{li2025sp,
  title={Sp-vla: A joint model scheduling and token pruning approach for vla model acceleration},
  author={Li, Ye and Meng, Yuan and Sun, Zewen and Ji, Kangye and Tang, Chen and Fan, Jiajun and Ma, Xinzhu and Xia, Shu-Tao and Wang, Zhi and Zhu, Wenwu},
  booktitle={The Fourteenth International Conference on Learning Representations},
  year={2025}
}

@inproceedings{song2024germ,
  title={Germ: A generalist robotic model with mixture-of-experts for quadruped robot},
  author={Song, Wenxuan and Zhao, Han and Ding, Pengxiang and Cui, Can and Lyu, Shangke and Fan, Yaning and Wang, Donglin},
  booktitle={2024 IEEE/RSJ International Conference on Intelligent Robots and Systems (IROS)},
  pages={11879--11886},
  year={2024},
  organization={IEEE}
}

@inproceedings{miao2025fedvla,
  title={Fedvla: Federated vision-language-action learning with dual gating mixture-of-experts for robotic manipulation},
  author={Miao, Cui and Chang, Tao and Wu, Meihan and Xu, Hongbin and Li, Chun and Li, Ming and Wang, Xiaodong},
  booktitle={Proceedings of the IEEE/CVF International Conference on Computer Vision},
  pages={6904--6913},
  year={2025}
}

@inproceedings{bai2025learning,
  title={Learning to see and act: Task-aware virtual view exploration for robotic manipulation},
  author={Bai, Yongjie and Wang, Zhouxia and Liu, Yang and Luo, Kaijun and Wen, Yifan and Dai, Mingtong and Chen, Weixing and Chen, Ziliang and Liu, Lingbo and Li, Guanbin and others},
  booktitle={Proceedings of the IEEE/CVF Conference on Computer Vision and Pattern Recognition},
  pages={13386--13396},
  year={2026}
}

@article{wason1974dual,
  title={Dual processes in reasoning?},
  author={Wason, Peter C and Evans, J St BT},
  journal={Cognition},
  volume={3},
  number={2},
  pages={141--154},
  year={1974},
  publisher={Elsevier}
}

@book{kahneman2011thinking,
  title={Thinking, fast and slow},
  author={Kahneman, Daniel},
  year={2011},
  publisher={macmillan}
}

@inproceedings{zhang2025hirt,
  title={HiRT: Enhancing Robotic Control with Hierarchical Robot Transformers},
  author={Zhang, Jianke and Guo, Yanjiang and Chen, Xiaoyu and Wang, Yen-Jen and Hu, Yucheng and Shi, Chengming and Chen, Jianyu},
  booktitle={Conference on Robot Learning},
  pages={933--946},
  year={2025},
  organization={PMLR}
}

@article{han2024dual,
  title={A dual process vla: Efficient robotic manipulation leveraging vlm},
  author={Han, ByungOk and Kim, Jaehong and Jang, Jinhyeok},
  journal={arXiv preprint arXiv:2410.15549},
  year={2024}
}

@article{bu2024towards,
  title={Towards synergistic, generalized, and efficient dual-system for robotic manipulation},
  author={Bu, Qingwen and Li, Hongyang and Chen, Li and Cai, Jisong and Zeng, Jia and Cui, Heming and Yao, Maoqing and Qiao, Yu},
  journal={arXiv preprint arXiv:2410.08001},
  year={2024}
}

@inproceedings{li2025hamster,
  title={HAMSTER: Hierarchical Action Models for Open-World Robot Manipulation},
  author={Li, Yi and Deng, Yuquan and Zhang, Jesse and Jang, Joel and Memmel, Marius and Garrett, Caelan Reed and Ramos, Fabio and Fox, Dieter and Li, Anqi and Gupta, Abhishek and others},
  booktitle={The Thirteenth International Conference on Learning Representations},
  year={2025}
}

@article{chen2025fast,
  title={Fast-in-Slow: a dual-system VLA model unifying fast manipulation within slow reasoning},
  author={Chen, Hao and Liu, Jiaming and Gu, Chenyang and Liu, Zhuoyang and Zhang, Renrui and Li, Xiaoqi and He, Xiao and Guo, Yandong and Fu, Chi-Wing and Zhang, Shanghang and others},
  journal={Advances in Neural Information Processing Systems},
  volume={38},
  pages={98049--98083},
  year={2026}
}

@article{chi2025mind,
  title={MinD: Learning A Dual-System World Model for Real-Time Planning and Implicit Risk Analysis},
  author={Chi, Xiaowei and Ge, Kuangzhi and Liu, Jiaming and Zhou, Siyuan and Jia, Peidong and He, Zichen and Liu, Yuzhen and Li, Tingguang and Han, Lei and Han, Sirui and others},
  journal={arXiv preprint arXiv:2506.18897},
  year={2025}
}

@article{duan2025fast,
  title={Fast ECoT: Efficient Embodied Chain-of-Thought via Thoughts Reuse},
  author={Duan, Zhekai and Zhang, Yuan and Geng, Shikai and Liu, Gaowen and Boedecker, Joschka and Lu, Chris Xiaoxuan},
  journal={arXiv preprint arXiv:2506.07639},
  year={2025}
}

@article{yue2024deer,
  title={Deer-vla: Dynamic inference of multimodal large language models for efficient robot execution},
  author={Yue, Yang and Wang, Yulin and Kang, Bingyi and Han, Yizeng and Wang, Shenzhi and Song, Shiji and Feng, Jiashi and Huang, Gao},
  journal={Advances in Neural Information Processing Systems},
  volume={37},
  pages={56619--56643},
  year={2024}
}

@article{chen2025rlrc,
  title={Rlrc: Reinforcement learning-based recovery for compressed vision-language-action models},
  author={Chen, Yuxuan and Han, Yixin and Huang, Yize and Li, Xiao},
  journal={IEEE Robotics and Automation Letters},
  year={2026},
  publisher={IEEE}
}

@inproceedings{reuss2025flower,
  title={FLOWER: Democratizing Generalist Robot Policies with Efficient Vision-Language-Flow Models},
  author={Reuss, Moritz and Zhou, Hongyi and R{\"u}hle, Marcel and Ya{\u{g}}murlu, {\"O}mer Erdin{\c{c}} and Otto, Fabian and Lioutikov, Rudolf},
  booktitle={Conference on Robot Learning},
  pages={3736--3761},
  year={2025},
  organization={PMLR}
}

@inproceedings{wu2025device,
  title={On-device diffusion transformer policy for efficient robot manipulation},
  author={Wu, Yiming and Wang, Huan and Chen, Zhenghao and Pang, Jianxin and Xu, Dong},
  booktitle={Proceedings of the IEEE/CVF International Conference on Computer Vision},
  pages={14073--14083},
  year={2025}
}

@article{park2024quantization,
  title={Quantization-aware imitation-learning for resource-efficient robotic control},
  author={Park, Seongmin and Kim, Hyungmin and Jeon, Wonseok and Yang, Juyoung and Jeon, Byeongwook and Oh, Yoonseon and Choi, Jungwook},
  journal={arXiv preprint arXiv:2412.01034},
  year={2024}
}

@inproceedings{park2025saliency,
  title={Saliency-aware quantized imitation learning for efficient robotic control},
  author={Park, Seongmin and Kim, Hyungmin and Kim, Sangwoo and Jeon, Wonseok and Yang, Juyoung and Jeon, Byeongwook and Oh, Yoonseon and Choi, Jungwook},
  booktitle={Proceedings of the IEEE/CVF International Conference on Computer Vision},
  pages={13140--13150},
  year={2025}
}

@article{wang2025bitvla,
  title={BitVLA: 1-bit Vision-Language-Action Models for Robotics Manipulation},
  author={Wang, Hongyu and Xiong, Chuyan and Wang, Ruiping and Chen, Xilin},
  journal={arXiv preprint arXiv:2506.07530},
  year={2025}
}

@article{fang2025sqap,
  title={SQAP-VLA: A Synergistic Quantization-Aware Pruning Framework for High-Performance Vision-Language-Action Models},
  author={Fang, Hengyu and Liu, Yijiang and Du, Yuan and Du, Li and Yang, Huanrui},
  journal={arXiv preprint arXiv:2509.09090},
  year={2025}
}

@article{li2025cogvla,
  title={CogVLA: Cognition-aligned vision-language-action models via instruction-driven routing \& sparsification},
  author={Li, Wei and Zhang, Renshan and Shao, Rui and He, Jie and Nie, Liqiang},
  journal={Advances in neural information processing systems},
  volume={38},
  pages={137646--137675},
  year={2026}
}

@article{lin2025vote,
  title={Vote: Vision-language-action optimization with trajectory ensemble voting},
  author={Lin, Juyi and Taherin, Amir and Akbari, Arash and Akbari, Arman and Lu, Lei and Chen, Guangyu and Padir, Taskin and Yang, Xiaomeng and Chen, Weiwei and Li, Yiqian and others},
  journal={arXiv preprint arXiv:2507.05116},
  year={2025}
}

@article{tan2025think,
  title={Think Twice, Act Once: Token-Aware Compression and Action Reuse for Efficient Inference in Vision-Language-Action Models},
  author={Tan, Xudong and Yang, Yaoxin and Ye, Peng and Zheng, Jialin and Bai, Bizhe and Wang, Xinyi and Hao, Jia and Chen, Tao},
  journal={arXiv preprint arXiv:2505.21200},
  year={2025}
}

@article{wang2025specprune,
  title={SpecPrune-VLA: Accelerating Vision-Language-Action Models via Action-Aware Self-Speculative Pruning},
  author={Wang, Hanzhen and Xu, Jiaming and Pan, Jiayi and Zhou, Yongkang and Dai, Guohao},
  journal={arXiv preprint arXiv:2509.05614},
  year={2025}
}

@article{jiang2025better,
  title={The Better You Learn, The Smarter You Prune: Towards Efficient Vision-language-action Models via Differentiable Token Pruning},
  author={Jiang, Titong and Jiang, Xuefeng and Ma, Yuan and Wen, Xin and Li, Bailin and Zhan, Kun and Jia, Peng and Liu, Yahui and Sun, Sheng and Lang, Xianpeng},
  journal={arXiv preprint arXiv:2509.12594},
  year={2025}
}

@inproceedings{pei2025action,
  title={Action-aware dynamic pruning for efficient vision-language-action manipulation},
  author={Pei, Xiaohuan and Chen, Yuxing and Xu, Siyu and Wang, Yunke and Shi, Yuheng and Xu, Chang},
  booktitle={International Conference on Learning Representations},
  volume={2026},
  pages={10832--10851},
  year={2026}
}

@inproceedings{bendikas2025focusing,
  title={Focusing on What Matters: Object-Agent-centric Tokenization for Vision Language Action models},
  author={Bendikas, Rokas and Dijkman, Daniel and Peschl, Markus and Haresh, Sanjay and Mazzaglia, Pietro},
  booktitle={9th Annual Conference on Robot Learning},
  year={2025}
}

@article{xu2025vla,
  title={Vla-cache: Efficient vision-language-action manipulation via adaptive token caching},
  author={Xu, Siyu and Wang, Yunke and Xia, Chenghao and Zhu, Dihao and Huang, Tao and Xu, Chang},
  journal={Advances in Neural Information Processing Systems},
  volume={38},
  pages={164448--164473},
  year={2026}
}

@inproceedings{li2025cronusvla,
  title={Towards efficient and robust manipulation via multi-frame vision-language-action modeling},
  author={Li, Hao and Yang, Shuai and Chen, Yilun and Chen, Xinyi and Yang, Xiaoda and Tian, Yang and Wang, Hanqing and Wang, Tai and Lin, Dahua and Zhao, Feng and others},
  booktitle={Proceedings of the AAAI Conference on Artificial Intelligence},
  volume={40},
  number={22},
  pages={18388--18396},
  year={2026}
}

@article{hu2022lora,
  title={Lora: Low-rank adaptation of large language models.},
  author={Hu, Edward J and Shen, Yelong and Wallis, Phillip and Allen-Zhu, Zeyuan and Li, Yuanzhi and Wang, Shean and Wang, Lu and Chen, Weizhu and others},
  journal={ICLR},
  volume={1},
  number={2},
  pages={3},
  year={2022}
}

@article{luo2024moil,
  title={Moil: Momentum imitation learning for efficient vision-language adaptation},
  author={Luo, Gen and Zhou, Yiyi and Huang, Minglang and Ren, Tianhe and Sun, Xiaoshuai and Ji, Rongrong},
  journal={IEEE Transactions on Pattern Analysis and Machine Intelligence},
  year={2024},
  publisher={IEEE}
}

@article{yu2025moe,
  title={MoE-Adapters++: Towards More Efficient Continual Learning of Vision-Language Models via Dynamic Mixture-of-Experts Adapters},
  author={Yu, Jiazuo and Huang, Zichen and Zhuge, Yunzhi and Zhang, Lu and Hu, Ping and Wang, Dong and Lu, Huchuan and He, You},
  journal={IEEE Transactions on Pattern Analysis and Machine Intelligence},
  year={2025},
  publisher={IEEE}
}

@article{zhang2024vision,
  title={Vision-language models for vision tasks: A survey},
  author={Zhang, Jingyi and Huang, Jiaxing and Jin, Sheng and Lu, Shijian},
  journal={IEEE transactions on pattern analysis and machine intelligence},
  volume={46},
  number={8},
  pages={5625--5644},
  year={2024},
  publisher={IEEE}
}

@inproceedings{li2025llara,
  title={LLaRA: Supercharging Robot Learning Data for Vision-Language Policy},
  author={Li, Xiang and Mata, Cristina and Park, Jongwoo and Kahatapitiya, Kumara and Jang, Yoo Sung and Shang, Jinghuan and Ranasinghe, Kanchana and Burgert, Ryan D and Cai, Mu and Lee, Yong Jae and others},
  booktitle={The Thirteenth International Conference on Learning Representations},
  year={2025}
}

@article{fan2025diffusion,
  title={Diffusion trajectory-guided policy for long-horizon robot manipulation},
  author={Fan, Shichao and Yang, Quantao and Liu, Yajie and Wu, Kun and Che, Zhengping and Liu, Qingjie and Wan, Min},
  journal={IEEE Robotics and Automation Letters},
  year={2025},
  publisher={IEEE}
}

@inproceedings{ye2025latent,
  title={Latent Action Pretraining from Videos},
  author={Ye, Seonghyeon and Jang, Joel and Jeon, Byeongguk and Joo, Se June and Yang, Jianwei and Peng, Baolin and Mandlekar, Ajay and Tan, Reuben and Chao, Yu-Wei and Lin, Bill Yuchen and others},
  booktitle={The Thirteenth International Conference on Learning Representations},
  year={2025}
}

@article{bu2025learning,
  title={UniVLA: Learning to Act Anywhere with Task-centric Latent Actions},
  author={Bu, Qingwen and Yang, Yanting and Cai, Jisong and Gao, Shenyuan and Ren, Guanghui and Yao, Maoqing and Luo, Ping and Li, Hongyang},
  journal={arXiv preprint arXiv:2505.06111},
  year={2025}
}

@article{tharwat2025latent,
  title={Latent Action Pretraining Through World Modeling},
  author={Tharwat, Bahey and Nasser, Yara and Abouzeid, Ali and Reid, Ian},
  journal={arXiv preprint arXiv:2509.18428},
  year={2025}
}

@article{yang2025egovla,
  title={Egovla: Learning vision-language-action models from egocentric human videos},
  author={Yang, Ruihan and Yu, Qinxi and Wu, Yecheng and Yan, Rui and Li, Borui and Cheng, An-Chieh and Zou, Xueyan and Fang, Yunhao and Cheng, Xuxin and Qiu, Ri-Zhao and others},
  journal={arXiv preprint arXiv:2507.12440},
  year={2025}
}

@article{yoshida2025developing,
  title={Developing Vision-Language-Action Model from Egocentric Videos},
  author={Yoshida, Tomoya and Kurita, Shuhei and Nishimura, Taichi and Mori, Shinsuke},
  journal={arXiv preprint arXiv:2509.21986},
  year={2025}
}

@article{luo2025being,
  title={Being-h0: vision-language-action pretraining from large-scale human videos},
  author={Luo, Hao and Feng, Yicheng and Zhang, Wanpeng and Zheng, Sipeng and Wang, Ye and Yuan, Haoqi and Liu, Jiazheng and Xu, Chaoyi and Jin, Qin and Lu, Zongqing},
  journal={arXiv preprint arXiv:2507.15597},
  year={2025}
}

@article{jiang2025rynnvla,
  title={RynnVLA-001: Using Human Demonstrations to Improve Robot Manipulation},
  author={Jiang, Yuming and Huang, Siteng and Xue, Shengke and Zhao, Yaxi and Cen, Jun and Leng, Sicong and Li, Kehan and Guo, Jiayan and Wang, Kexiang and Chen, Mingxiu and others},
  journal={arXiv preprint arXiv:2509.15212},
  year={2025}
}

@inproceedings{wang2025unified,
  title={Unified vision-language-action model},
  author={Wang, Yuqi and Li, Xinghang and Wang, Wenxuan and Zhang, Junbo and Li, Yingyan and Chen, Yuntao and Wang, Xinlong and Zhang, Zhaoxiang},
  booktitle={International Conference on Learning Representations},
  volume={2026},
  pages={80929--80944},
  year={2026}
}

@article{tan2025anypos,
  title={Anypos: Automated task-agnostic actions for bimanual manipulation},
  author={Tan, Hengkai and Feng, Yao and Mao, Xinyi and Huang, Shuhe and Liu, Guodong and Hao, Zhongkai and Su, Hang and Zhu, Jun},
  journal={arXiv preprint arXiv:2507.12768},
  year={2025}
}

@article{dai2025prepare,
  title={Prepare Before You Act: Learning From Humans to Rearrange Initial States},
  author={Dai, Yinlong and Keyser, Andre and Losey, Dylan P},
  journal={arXiv preprint arXiv:2509.18043},
  year={2025}
}

@inproceedings{wang2025vla,
  title={Vla-adapter: An effective paradigm for tiny-scale vision-language-action model},
  author={Wang, Yihao and Ding, Pengxiang and Li, Lingxiao and Cui, Can and Ge, Zirui and Tong, Xinyang and Song, Wenxuan and Zhao, Han and Zhao, Wei and Hou, Pengxu and others},
  booktitle={Proceedings of the AAAI conference on artificial intelligence},
  volume={40},
  number={22},
  pages={18638--18646},
  year={2026}
}

@inproceedings{wen2025dexvla,
  title={DexVLA: Vision-Language Model with Plug-In Diffusion Expert for General Robot Control},
  author={Wen, Junjie and Zhu, Yichen and Li, Jinming and Tang, Zhibin and Shen, Chaomin and Feng, Feifei},
  booktitle={Conference on Robot Learning},
  pages={3094--3114},
  year={2025},
  organization={PMLR}
}

@article{ding2025humanoid,
  title={Humanoid-vla: Towards universal humanoid control with visual integration},
  author={Ding, Pengxiang and Ma, Jianfei and Tong, Xinyang and Zou, Binghong and Luo, Xinxin and Fan, Yiguo and Wang, Ting and Lu, Hongchao and Mo, Panzhong and Liu, Jinxin and others},
  journal={arXiv preprint arXiv:2502.14795},
  year={2025}
}

@article{li2025atomic,
  title={An atomic skill library construction method for data-efficient embodied manipulation},
  author={Li, Dongjiang and Peng, Bo and Li, Chang and Qiao, Ning and Zheng, Qi and Sun, Lei and Qin, Yusen and Li, Bangguo and Luan, Yifeng and Wu, Bo and others},
  journal={arXiv preprint arXiv:2501.15068},
  year={2025}
}

@inproceedings{wu2025momanipvla,
  title={Momanipvla: Transferring vision-language-action models for general mobile manipulation},
  author={Wu, Zhenyu and Zhou, Yuheng and Xu, Xiuwei and Wang, Ziwei and Yan, Haibin},
  booktitle={Proceedings of the Computer Vision and Pattern Recognition Conference},
  pages={1714--1723},
  year={2025}
}

@article{cui2025openhelix,
  title={Openhelix: A short survey, empirical analysis, and open-source dual-system vla model for robotic manipulation},
  author={Cui, Can and Ding, Pengxiang and Song, Wenxuan and Bai, Shuanghao and Tong, Xinyang and Ge, Zirui and Suo, Runze and Zhou, Wanqi and Liu, Yang and Jia, Bofang and others},
  journal={arXiv preprint arXiv:2505.03912},
  year={2025}
}

@inproceedings{li2025controlvla,
  title={ControlVLA: Few-shot Object-centric Adaptation for Pre-trained Vision-Language-Action Models},
  author={Li, Puhao and Wu, Yingying and Xi, Ziheng and Li, Wanlin and Huang, Yuzhe and Zhang, Zhiyuan and Chen, Yinghan and Wang, Jianan and Zhu, Song-Chun and Liu, Tengyu and others},
  booktitle={Conference on Robot Learning},
  pages={1898--1913},
  year={2025},
  organization={PMLR}
}

@inproceedings{sridhar2025ricl,
  title={RICL: Adding In-Context Adaptability to Pre-Trained Vision-Language-Action Models},
  author={Sridhar, Kaustubh and Dutta, Souradeep and Jayaraman, Dinesh and Lee, Insup},
  booktitle={9th Annual Conference on Robot Learning},
  year={2025}
}

@inproceedings{zhang2025align,
  title={Align-then-steer: Adapting the vision-language action models through unified latent guidance},
  author={Zhang, Yang and Wang, Chenwei and Zhao, Yuan and Ge, Yunfei and Sun, Zhenglong and Li, Xiu and Zhang, Chi and Bai, Chenjia and Li, Xuelong and others},
  booktitle={International Conference on Learning Representations},
  volume={2026},
  pages={14717--14749},
  year={2026}
}

@inproceedings{tan2025interactive,
  title={Interactive Post-Training for Vision-Language-Action Models},
  author={Tan, Shuhan and Dou, Kairan and Zhao, Yue and Kraehenbuehl, Philipp},
  booktitle={Workshop on Foundation Models Meet Embodied Agents at CVPR 2025},
  year={2025}
}

@article{lu2025vla,
  title={Vla-rl: Towards masterful and general robotic manipulation with scalable reinforcement learning},
  author={Lu, Guanxing and Guo, Wenkai and Zhang, Chubin and Zhou, Yuheng and Jiang, Haonan and Gao, Zifeng and Tang, Yansong and Wang, Ziwei},
  journal={arXiv preprint arXiv:2505.18719},
  year={2025}
}

@inproceedings{julg2025refined,
  title={Refined policy distillation: From vla generalists to rl experts},
  author={J{\"u}lg, Tobias and Burgard, Wolfram and Walter, Florian},
  booktitle={2025 IEEE/RSJ International Conference on Intelligent Robots and Systems (IROS)},
  pages={11677--11684},
  year={2025},
  organization={IEEE}
}

@article{jin2025dual,
  title={Dual-Actor Fine-Tuning of VLA Models: A Talk-and-Tweak Human-in-the-Loop Approach},
  author={Jin, Piaopiao and Wang, Qi and Sun, Guokang and Cai, Ziwen and He, Pinjia and You, Yangwei},
  journal={arXiv preprint arXiv:2509.13774},
  year={2025}
}

@article{xiao2025world,
  title={World-Env: Leveraging World Model as a Virtual Environment for VLA Post-Training},
  author={Xiao, Junjin and Yang, Yandan and Chang, Xinyuan and Chen, Ronghan and Xiong, Feng and Xu, Mu and Zheng, Wei-Shi and Zhang, Qing},
  journal={arXiv preprint arXiv:2509.24948},
  year={2025}
}

@article{chen2025conrft,
  title={Conrft: A reinforced fine-tuning method for vla models via consistency policy},
  author={Chen, Yuhui and Tian, Shuai and Liu, Shugao and Zhou, Yingting and Li, Haoran and Zhao, Dongbin},
  journal={arXiv preprint arXiv:2502.05450},
  year={2025}
}

@article{huang2025co,
  title={CO-RFT: Efficient Fine-Tuning of Vision-Language-Action Models through Chunked Offline Reinforcement Learning},
  author={Huang, Dongchi and Fang, Zhirui and Zhang, Tianle and Li, Yihang and Zhao, Lin and Xia, Chunhe},
  journal={arXiv preprint arXiv:2508.02219},
  year={2025}
}

@article{nakamoto2023cal,
  title={Cal-ql: Calibrated offline rl pre-training for efficient online fine-tuning},
  author={Nakamoto, Mitsuhiko and Zhai, Simon and Singh, Anikait and Sobol Mark, Max and Ma, Yi and Finn, Chelsea and Kumar, Aviral and Levine, Sergey},
  journal={Advances in Neural Information Processing Systems},
  volume={36},
  pages={62244--62269},
  year={2023}
}

@inproceedings{zhang2025balancing,
  title={Balancing signal and variance: Adaptive offline rl post-training for vla flow models},
  author={Zhang, Hongyin and Zhang, Shiyuan and Jin, Junxi and Zeng, Qixin and Qiao, Yifan and Lu, Hongchao and Wang, Donglin},
  booktitle={Proceedings of the AAAI Conference on Artificial Intelligence},
  volume={40},
  number={22},
  pages={18755--18763},
  year={2026}
}

@article{wang2025genie,
  title={Genie Centurion: Accelerating Scalable Real-World Robot Training with Human Rewind-and-Refine Guidance},
  author={Wang, Wenhao and Song, Jianheng and Liu, Chiming and Ma, Jiayao and Feng, Siyuan and Wang, Jingyuan and Jiang, Yuxin and Chen, Kylin and Zhan, Sikang and Wang, Yi and others},
  journal={arXiv preprint arXiv:2505.18793},
  year={2025}
}

@inproceedings{deng2025graspvla,
  title={GraspVLA: a Grasping Foundation Model Pre-trained on Billion-scale Synthetic Action Data},
  author={Deng, Shengliang and Yan, Mi and Wei, Songlin and Ma, Haixin and Yang, Yuxin and Chen, Jiayi and Zhang, Zhiqi and Yang, Taoyu and Zhang, Xuheng and Cui, Heming and others},
  booktitle={Conference on Robot Learning},
  pages={1004--1029},
  year={2025},
  organization={PMLR}
}

@inproceedings{argus2025cvla,
  title={cVLA: Towards Efficient Camera-Space VLAs},
  author={Argus, Max and Bratuli{\'c}, Jelena and Masnavi, Houman and Velikanov, Maksim and Heppert, Nick and Valada, Abhinav and Brox, Thomas},
  booktitle={CoRL 2025 Workshop RemembeRL},
  year={2025}
}

@article{chen2025robotwin,
  title={Robotwin 2.0: A scalable data generator and benchmark with strong domain randomization for robust bimanual robotic manipulation},
  author={Chen, Tianxing and Chen, Zanxin and Chen, Baijun and Cai, Zijian and Liu, Yibin and Li, Zixuan and Liang, Qiwei and Lin, Xianliang and Ge, Yiheng and Gu, Zhenyu and others},
  journal={arXiv preprint arXiv:2506.18088},
  year={2025}
}

@inproceedings{fang2025rebot,
  title={Rebot: Scaling robot learning with real-to-sim-to-real robotic video synthesis},
  author={Fang, Yu and Yang, Yue and Zhu, Xinghao and Zheng, Kaiyuan and Bertasius, Gedas and Szafir, Daniel and Ding, Mingyu},
  booktitle={2025 IEEE/RSJ International Conference on Intelligent Robots and Systems (IROS)},
  pages={11351--11358},
  year={2025},
  organization={IEEE}
}

@inproceedings{yu2025real2render2real,
  title={Real2Render2Real: Scaling Robot Data Without Dynamics Simulation or Robot Hardware},
  author={Yu, Justin and Fu, Letian and Huang, Huang and El-Refai, Karim and Ambrus, Rares Andrei and Cheng, Richard and Irshad, Muhammad Zubair and Goldberg, Ken},
  booktitle={Conference on Robot Learning},
  pages={547--577},
  year={2025},
  organization={PMLR}
}

@article{tai2025realmirror,
  title={RealMirror: A Comprehensive, Open-Source Vision-Language-Action Platform for Embodied AI},
  author={Tai, Cong and Zheng, Zhaoyu and Long, Haixu and Wu, Hansheng and Xiang, Haodong and Long, Zhengbin and Xiong, Jun and Shi, Rong and Zhang, Shizhuang and Qiu, Gang and others},
  journal={arXiv preprint arXiv:2509.14687},
  year={2025}
}

@article{li2025mimicdreamer,
  title={MimicDreamer: Aligning Human and Robot Demonstrations for Scalable VLA Training},
  author={Li, Haoyun and Zhang, Ivan and Ouyang, Runqi and Wang, Xiaofeng and Zhu, Zheng and Yang, Zhiqin and Zhang, Zhentao and Wang, Boyuan and Ni, Chaojun and Qin, Wenkang and others},
  journal={arXiv preprint arXiv:2509.22199},
  year={2025}
}

@article{yang2025beyond,
  title={Beyond Human Demonstrations: Diffusion-Based Reinforcement Learning to Generate Data for VLA Training},
  author={Yang, Rushuai and Wei, Hangxing and Zhang, Ran and Feng, Zhiyuan and Chen, Xiaoyu and Li, Tong and Zhang, Chuheng and Zhao, Li and Bian, Jiang and Su, Xiu and others},
  journal={arXiv preprint arXiv:2509.19752},
  year={2025}
}

@article{li2025vla,
  title={VLA-RFT: Vision-Language-Action Reinforcement Fine-tuning with Verified Rewards in World Simulators},
  author={Li, Hengtao and Ding, Pengxiang and Suo, Runze and Wang, Yihao and Ge, Zirui and Zang, Dongyuan and Yu, Kexian and Sun, Mingyang and Zhang, Hongyin and Wang, Donglin and others},
  journal={arXiv preprint arXiv:2510.00406},
  year={2025}
}

@inproceedings{song2025reconvla,
  title={Reconvla: Reconstructive vision-language-action model as effective robot perceiver},
  author={Song, Wenxuan and Zhou, Ziyang and Zhao, Han and Chen, Jiayi and Ding, Pengxiang and Yan, Haodong and Huang, Yuxin and Tang, Feilong and Wang, Donglin and Li, Haoang},
  booktitle={Proceedings of the AAAI Conference on Artificial Intelligence},
  volume={40},
  number={22},
  pages={18549--18557},
  year={2026}
}

@article{nie2025ermv,
  title={ERMV: Editing 4D Robotic Multi-view images to enhance embodied agents},
  author={Nie, Chang and Wang, Guangming and Lie, Zhe and Wang, Hesheng},
  journal={arXiv preprint arXiv:2507.17462},
  year={2025}
}

@article{dong2025emma,
  title={EMMA: Generalizing Real-World Robot Manipulation via Generative Visual Transfer},
  author={Dong, Zhehao and Wang, Xiaofeng and Zhu, Zheng and Wang, Yirui and Wang, Yang and Zhou, Yukun and Wang, Boyuan and Ni, Chaojun and Ouyang, Runqi and Qin, Wenkang and others},
  journal={arXiv preprint arXiv:2509.22407},
  year={2025}
}

@article{luo2025adathinkdrive,
  title={AdaThinkDrive: Adaptive Thinking via Reinforcement Learning for Autonomous Driving},
  author={Luo, Yuechen and Li, Fang and Xu, Shaoqing and Lai, Zhiyi and Yang, Lei and Chen, Qimao and Luo, Ziang and Xie, Zixun and Jiang, Shengyin and Liu, Jiaxin and others},
  journal={arXiv preprint arXiv:2509.13769},
  year={2025}
}

@article{zhou2025autovla,
  title={Autovla: A vision-language-action model for end-to-end autonomous driving with adaptive reasoning and reinforcement fine-tuning},
  author={Zhou, Zewei and Cai, Tianhui and Zhao, Seth and Zhang, Yun and Huang, Zhiyu and Zhou, Bolei and Ma, Jiaqi},
  journal={Advances in Neural Information Processing Systems},
  volume={38},
  pages={27920--27956},
  year={2026}
}

@article{jiang2025diffvla,
  title={Diffvla: Vision-language guided diffusion planning for autonomous driving},
  author={Jiang, Anqing and Gao, Yu and Sun, Zhigang and Wang, Yiru and Wang, Jijun and Chai, Jinghao and Cao, Qian and Heng, Yuweng and Jiang, Hao and Dong, Yunda and others},
  journal={arXiv preprint arXiv:2505.19381},
  year={2025}
}

@article{margadji2025hybrid,
  title={Hybrid reasoning for perception, explanation, and autonomous action in manufacturing},
  author={Margadji, Christos and Pattinson, Sebastian William},
  journal={Nature Communications},
  year={2026},
  publisher={Nature Publishing Group UK London}
}

@article{wu2026pragmatic,
  title={A Pragmatic VLA Foundation Model},
  author={Wu, Wei and Lu, Fan and Wang, Yunnan and Yang, Shuai and Liu, Shi and Wang, Fangjing and Zhu, Qian and Sun, He and Wang, Yong and Ma, Shuailei and others},
  journal={arXiv preprint arXiv:2601.18692},
  year={2026}
}

@inproceedings{tian2025interndata,
  title={Interndata-a1: Pioneering high-fidelity synthetic data for pre-training generalist policy},
  author={Tian, Yang and Yang, Yuyin and Xie, Yiman and Cai, Zetao and Shi, Xu and Gao, Ning and Liu, Hangxu and Jiang, Xuekun and Qiu, Zherui and Yuan, Feng and others},
  booktitle={Proceedings of the IEEE/CVF Conference on Computer Vision and Pattern Recognition},
  pages={976--985},
  year={2026}
}

@article{xu2026parameter,
  title={Parameter-efficient fine-tuning methods for pretrained language models: A critical review and assessment},
  author={Xu, Lingling and Xie, Haoran and Qin, S Joe and Tao, Xiaohui and Wang, Fu Lee},
  journal={IEEE Transactions on Pattern Analysis and Machine Intelligence},
  year={2026},
  publisher={IEEE}
}

@article{jacobs1991adaptive,
  title={Adaptive mixtures of local experts},
  author={Jacobs, Robert A and Jordan, Michael I and Nowlan, Steven J and Hinton, Geoffrey E},
  journal={Neural computation},
  volume={3},
  number={1},
  pages={79--87},
  year={1991},
  publisher={MIT Press}
}

@article{jiang2026vla,
  title={VLA-GSE: Boosting Parameter-Efficient Fine-Tuning in VLA with Generalized and Specialized Experts},
  author={Jiang, Yuhua and Lu, Junjie and Qin, Xinyao and Chen, Xiaoyu and Wang, Kaixin and Gao, Feifei and Zhao, Li},
  journal={arXiv preprint arXiv:2605.06175},
  year={2026}
}

@article{zhong2026vla,
  title={VLA-OPD: Bridging Offline SFT and Online RL for Vision-Language-Action Models via On-Policy Distillation},
  author={Zhong, Zhide and Yan, Haodong and Li, Junfeng and He, Junjie and Zhang, Tianran and Li, Haoang},
  journal={arXiv preprint arXiv:2603.26666},
  year={2026}
}

@article{zhang2026mvlad,
  title={Efficient and Explainable End-to-End Autonomous Driving via Masked Vision-Language-Action Diffusion},
  author={Zhang, Jiaru and Gagvani, Manav and Cui, Can and Peng, Juntong and Zhang, Ruqi and Wang, Ziran},
  journal={arXiv preprint arXiv:2602.20577},
  year={2026}
}

@article{tu2026sgvla,
  title={SG-VLA: Learning Spatially-Grounded Vision-Language-Action Models for Mobile Manipulation},
  author={Tu, Ruisen and Shukla, Arth and Yoo, Sohyun and Li, Xuanlin and Li, Junxi and Xie, Jianwen and Su, Hao and Tu, Zhuowen},
  journal={arXiv preprint arXiv:2603.22760},
  year={2026}
}

@inproceedings{li2025transferring,
  title={Transferring Vision-Language-Action Models to Industry Applications: Architectures, Performance, and Challenges},
  author={Li, Shuai and Chen, Yizhe and Li, Dong and Liu, Sichao and Lan, Dapeng and Liu, Yu and Pang, Zhibo},
  booktitle={2025 7th International Conference on Industrial Artificial Intelligence (IAI)},
  pages={1--6},
  year={2025},
  organization={IEEE}
}

@article{liu2026selfvla,
  title={SELF-VLA: A Skill Enhanced Agentic Vision-Language-Action Framework for Contact-Rich Disassembly},
  author={Liu, Chang and Tian, Sibo and Liang, Xiao and Zheng, Minghui},
  journal={arXiv preprint arXiv:2603.11080},
  year={2026}
}

@article{robotics2026habilis,
  title={Habilis-$\beta$: A Fast-Motion and Long-Lasting On-Device Vision-Language-Action Model},
  author={Robotics, Tommoro and Kang, Jesoon and Park, Taegeon and An, Jisu and Kimm, Soo Min and Kim, Jaejoon and Pahk, Jinu and Kim, Byungju and Lee, Junseok and Baek, Namheon and others},
  journal={arXiv preprint arXiv:2602.18813},
  year={2026}
}

@article{sun2025collabvla,
  title={CollabVLA: Self-Reflective Vision-Language-Action Model Dreaming Together with Human},
  author={Sun, Nan and Li, Yongchang and Wang, Chenxu and Li, Huiying and Liu, Huaping},
  journal={arXiv preprint arXiv:2509.14889},
  year={2025}
}

@article{wang2026vision,
  title={Vision-Language-Action in Robotics: A Survey of Datasets, Benchmarks, and Data Engines},
  author={Wang, Ziyao and Wang, Bingying and Zhang, Hanrong and Du, Tingting and Chen, Tianyang and Sun, Guoheng and He, Yexiao and Shen, Zheyu and Ye, Wanghao and Li, Ang},
  journal={Transactions on Machine Learning Research},
  year={2026}
}

@article{hui2026seeing,
  title={Seeing Realism from Simulation: Efficient Video Transfer for Vision-Language-Action Data Augmentation},
  author={Hui, Chenyu and Huang, Xiaodi and Xu, Siyu and Wang, Yunke and You, Shan and Wang, Fei and Huang, Tao and Xu, Chang},
  journal={arXiv preprint arXiv:2605.02757},
  year={2026}
}

@article{simple2026hifi,
  title={HiFi-UMI: Learning Deployable Manipulation Policies from High-Fidelity UMI Data Alone},
  author={{Simple AI} and Wei, Yuteng and Ma, Jinming and Wang, Jiawei and Zhou, Weitao and Zuo, Yushen and Rui, Ke and Li, Minglei and Zhang, Jinhao and Pan, Zhikang and Wang, Xiang and others},
  journal={arXiv preprint arXiv:2607.25895},
  year={2026}
}

@article{xie2026dynamicvla,
  title={DynamicVLA: A Vision-Language-Action Model for Dynamic Object Manipulation},
  author={Xie, Haozhe and Wen, Beichen and Zheng, Jiarui and Chen, Zhaoxi and Hong, Fangzhou and Diao, Haiwen and Liu, Ziwei},
  journal={arXiv preprint arXiv:2601.22153},
  year={2026}
}

@inproceedings{zhang2025robowheel,
  title={Robowheel: A data engine from real-world human demonstrations for cross-embodiment robotic learning},
  author={Zhang, Yuhong and Gao, Zihan and Li, Shengpeng and Chen, Ling-Hao and Liu, Kaisheng and Cheng, Runqing and Lin, Xiao and Liu, Junjia and Li, Zhuoheng and Feng, Jingyi and others},
  booktitle={Proceedings of the IEEE/CVF Conference on Computer Vision and Pattern Recognition},
  pages={6664--6674},
  year={2026}
}

@article{guo2026reflex,
  title={Reflex: Real-Time VLA Control through Streaming Inference},
  author={Guo, Yuanchun and Liu, Bingyan},
  journal={arXiv preprint arXiv:2607.14695},
  year={2026}
}

@inproceedings{ma2026blurr,
  title={BLURR: A Boosted Low-Resource Inference for Vision-Language-Action Model},
  author={Ma, Xiaoyu and Yuan, Zhengqing and Zhang, Zheyuan and Shi, Kaiwen and Sun, Lichao and Ye, Yanfang},
  booktitle={Companion Proceedings of the ACM Web Conference 2026},
  pages={225--228},
  year={2026}
}

@inproceedings{li2026semanticvla,
  title={Semanticvla: Semantic-aligned sparsification and enhancement for efficient robotic manipulation},
  author={Li, Wei and Zhang, Renshan and Shao, Rui and Fang, Zhijian and Zhou, Kaiwen and Tian, Zhuotao and Nie, Liqiang},
  booktitle={Proceedings of the AAAI Conference on Artificial Intelligence},
  volume={40},
  number={22},
  pages={18397--18405},
  year={2026}
}

@article{wang2026open,
  title={Open-loop planning, closed-loop verification: Speculative verification for vla},
  author={Wang, Zihua and Lin, Zhitao and Li, Ruibo and Zhang, Yu and Yang, Xu and Mi, Siya and Wei, Xiu-Shen},
  journal={arXiv preprint arXiv:2604.02965},
  year={2026}
}

@inproceedings{lin2026evo,
  title={Evo-1: Lightweight vision-language-action model with preserved semantic alignment},
  author={Lin, Tao and Zhong, Yilei and Du, Yuxin and Zhang, Jingjing and Liu, Jiting and Chen, Yinxinyu and Gu, Encheng and Liu, Ziyan and Cai, Hongyi and Zou, Yanwen and others},
  booktitle={Proceedings of the IEEE/CVF conference on computer vision and pattern recognition},
  pages={13397--13406},
  year={2026}
}

@inproceedings{williams2026lite,
  title={Lite vla: Efficient vision-language-action control on cpu-bound edge robots},
  author={Williams, Justin and Gupta, Kishor Datta and George, Roy and Sarkar, Mrinmoy},
  booktitle={2026 IEEE International Conference on Pervasive Computing and Communications Workshops and other Affiliated Events (PerCom Workshops)},
  pages={530--533},
  year={2026},
  organization={IEEE}
}

@article{zou2025asynchronous,
  title={Asynchronous fast-slow vision-language-action policies for whole-body robotic manipulation},
  author={Zou, Teqiang and Zeng, Hongliang and Nong, Yuxuan and Li, Yifan and Liu, Kehui and Yang, Haotian and Ling, Xinyang and Li, Xin and Ma, Lianyang},
  journal={arXiv preprint arXiv:2512.20188},
  year={2025}
}

@article{yan2026acting,
  title={Acting While Understanding: Asynchronous Semantic-Action Decoupling for Real-Time Vision-Language-Action Models},
  author={Yan, Shenhao and Wang, Ge and Liu, Qi and Meng, Weilin and Yang, Jiahao and Yao, Chengsi and Feng, Fan and Ma, Xiaoguang and Zhao, Yiming and Han, Yatong},
  journal={arXiv preprint arXiv:2606.15285},
  year={2026}
}

@article{sun2026unifs,
  title={UniFS: Unified Fast-to-Slow Hierarchical Architecture for Vision-Language-Action Models},
  author={Sun, Lin and Guan, Zhiwei and Wang, Conglin and Chen, Zihong and Yu, Jianhai and Li, Zongsheng and He, Boyong and Sun, Tao and Cao, Jiale and Liu, Lige},
  journal={arXiv preprint arXiv:2606.22794},
  year={2026}
}

@article{yang2026dysl,
  title={DySL-VLA: Efficient vision-language-action model inference via dynamic-static layer-skipping for robot manipulation},
  author={Yang, Zebin and Qi, Yijiahao and Xie, Tong and Yu, Bo and Liu, Shaoshan and Li, Meng},
  journal={arXiv preprint arXiv:2602.22896},
  year={2026}
}

@article{xu2026qvla,
  title={QVLA: Not All Channels Are Equal in Vision-Language-Action Model's Quantization},
  author={Xu, Yuhao and Yang, Yantai and Fan, Zhenyang and Liu, Yufan and Li, Yuming and Li, Bing and Zhang, Zhipeng},
  journal={arXiv preprint arXiv:2602.03782},
  year={2026}
}

@article{zhang2026quantvla,
  title={Quantvla: Scale-calibrated post-training quantization for vision-language-action models},
  author={Zhang, Jingxuan and Hsieh, Yunta and Wan, Zhongwei and Lin, Haokun and Wang, Xin and Wang, Ziqi and Lei, Yingtie and Zhang, Mi},
  journal={arXiv preprint arXiv:2602.20309},
  year={2026}
}

@article{gao2025compressor,
  title={Compressor-vla: Instruction-guided visual token compression for efficient robotic manipulation},
  author={Gao, Juntao and Ye, Feiyang and Zhang, Jing and Qian, Wenjing},
  journal={arXiv preprint arXiv:2511.18950},
  year={2025}
}

@article{liu2025bridging,
  title={Bridging the Semantic-Action Gap in Visual Token Pruning for Efficient VLA Inference},
  author={Liu, Ziyan and Chen, Yeqiu and Cai, Hongyi and Lin, Tao and Yang, Shuo and Liu, Zheng and Zhao, Bo},
  journal={arXiv preprint arXiv:2511.16449},
  year={2025}
}

@article{ye2026token,
  title={Token expand-merge: Training-free token compression for vision-language-action models},
  author={Ye, Yifan and Ma, Jiaqi and Cen, Jun and Lu, Zhihe},
  journal={IEEE Robotics and Automation Letters},
  year={2026},
  publisher={IEEE}
}

@article{qiu2026efficient,
  title={Efficient long-horizon vision-language-action models via static-dynamic disentanglement},
  author={Qiu, Weikang and Lei, Huashuo and Huang, Tinglin and Ying, Rex},
  journal={arXiv preprint arXiv:2602.03983},
  year={2026}
}

@inproceedings{im2026twinvla,
  title={TwinVLA: Data-efficient bimanual manipulation with twin single-arm vision-language-action models},
  author={Im, Hokyun and Jeong, Euijin and Kolobov, Andrey and Fu, Jianlong and Lee, Youngwoon},
  booktitle={International Conference on Learning Representations},
  volume={2026},
  pages={61969--61995},
  year={2026}
}

@article{zhang2026invertible,
  title={Invertible Neural Network Adapter for One-Step Flow Matching in Robot Manipulation},
  author={Zhang, Yu and Ji, Kangyi and Zou, Yongxiang and Xu, Rongtao and Zheng, Feng and Cheng, Long},
  journal={arXiv preprint arXiv:2606.19194},
  year={2026}
}

@article{zheng2026motionwam,
  title={MotionWAM: Towards Foundation World Action Models for Real-Time Humanoid Loco-Manipulation},
  author={Zheng, Jia and Ma, Teli and Fan, Yudong and Wang, Zifan and Yang, Shuo and Liang, Junwei},
  journal={arXiv preprint arXiv:2606.09215},
  year={2026}
}

@article{omaisan2025towards,
  title={Towards Accessible Physical AI: LoRA-Based Fine-Tuning of VLA Models for Real-World Robot Control},
  author={Omaisan, Abdullah Yahya Abdullah and Mohamed, Ibrahim Sheikh},
  journal={arXiv preprint arXiv:2512.11921},
  year={2025}
}

@article{kim2026adaptive,
  title={Adaptive capacity allocation for vision language action fine-tuning},
  author={Kim, Donghoon and Bae, Minji and Nam, Unghui and Kim, Gyeonghun and Lee, Suyun and Shim, Kyuhong and Shim, Byonghyo},
  journal={arXiv preprint arXiv:2603.07404},
  year={2026}
}

@article{xu2026rl,
  title={RL Token: Bootstrapping Online RL with Vision-Language-Action Models},
  author={Xu, Charles and Springenberg, Jost Tobias and Equi, Michael and Amin, Ali and Esmail, Adnan and Levine, Sergey and Ke, Liyiming},
  journal={arXiv preprint arXiv:2604.23073},
  year={2026}
}

@article{zhang2026survey,
  title={A Survey on Post-Training of Multimodal Large Language Models},
  author={Zhang, Haonan and Zeng, Pengpeng and Cao, Libin and Lai, Wenrui and Li, Jinlong and Peng, Duo and Bin, Yi and Wang, Xuanhan and Zhang, Ji and Song, Jingkuan and others},
  year={2026},
  publisher={Preprints}
}

@article{jiang2026wsa,
  title={WSA $ \_1 $: a 3D-Centric World-Spatial-Action Model for Generalizable Robot Control},
  author={Jiang, Jiahao and Zhang, Jianing and Yin, Zhenhan and Chen, Ruidong and Wang, Sen and Yu, Zhaoshu and Zeng, Pengpeng and Cao, Xiaofeng and Wang, Xuanhan and Song, Jingkuan and others},
  journal={arXiv preprint arXiv:2607.03941},
  year={2026}
}

@article{yuan2025more,
  title={More than one teacher: Adaptive multi-guidance policy optimization for diverse exploration},
  author={Yuan, Xiaoyang and Ding, Yujuan and Bin, Yi and Shao, Wenqi and Cai, Jinyu and Song, Jingkuan and Yang, Yang and Shen, Heng Tao},
  journal={arXiv preprint arXiv:2510.02227},
  year={2025}
}

@article{zhang2026omnicharacter++,
  title={OmniCharacter++: Towards Comprehensive Benchmark for Realistic Role-Playing Agents},
  author={Zhang, Haonan and Zeng, Pengpeng and Zhang, Ji and Song, Jingkuan and Sebe, Nicu and Shen, Heng Tao and Gao, Lianli},
  journal={IEEE Transactions on Pattern Analysis and Machine Intelligence},
  year={2026},
  publisher={IEEE}
}

@inproceedings{xiao2026rehearsevla,
  title={RehearseVLA: Simulated Post-Training for VLAs with Physically-Consistent World Model},
  author={Xiao, Junjin and Yang, Yandan and Chang, Xinyuan and Chen, Ronghan and Xiong, Feng and Xu, Mu and Zheng, Wei-Shi and Zhang, Qing},
  booktitle={Proceedings of the IEEE/CVF Conference on Computer Vision and Pattern Recognition},
  pages={20867--20877},
  year={2026}
}

\section{Biography Section}
\vspace{-35pt}

\begin{IEEEbiographynophoto}{Zhaoshu Yu} is currently pursuing the Ph.D. degree in Computer Science and Technology at Tongji University, Shanghai, China.
His research interests include vision-language-action models, embodied intelligence, and reinforcement learning.
\end{IEEEbiographynophoto}
\vspace{-35pt}

\begin{IEEEbiographynophoto}{Bo Wang} is currently a junior undergraduate student majoring in Computer Science and Technology at Tongji University, Shanghai, China.
His research interests include vision-language-action models, embodied intelligence, and multimodal learning.
\end{IEEEbiographynophoto}
\vspace{-35pt}

\begin{IEEEbiographynophoto}{Pengpeng Zeng} received the B.E. degree from Xi'an University of Technology in 2016, and the M.E. and Ph.D. degrees from University of Electronic Science and Technology of China in 2019 and 2023, respectively. He is now a Researcher in Tongji University, China.
His current research interests include visual understanding, machine learning, and reinforcement learning.
\end{IEEEbiographynophoto}
\vspace{-35pt}

\begin{IEEEbiographynophoto}{Haonan Zhang} is currently a Postdoctoral Researcher with the School of Computer Science and Technology, Tongji University, China. He obtained his Ph.D. degree from the University of Electronic Science and Technology of China in 2026. His research interests include multimodal learning, intelligent dialogue, and agent-based systems.
\end{IEEEbiographynophoto}
\vspace{-35pt}

\begin{IEEEbiographynophoto}{Ji Zhang} is an Assistant Professor with the School of Computing and Artificial Intelligence, Southwest Jiaotong University, China. He obtained his PhD degree from University of Electronic Science and Technology of China in 2024. His research interests include few-shot learning, transfer learning and robotics. 
\end{IEEEbiographynophoto}
\vspace{-35pt}

\begin{IEEEbiographynophoto}{Zheng Wang} received the B.E. and Ph.D. degrees from Zhejiang University, Hangzhou, China, in 2011 and 2017, respectively. He is currently with the School of Computer Science and Technology, Tongji University, Shanghai, China. 
His current research interests mainly focus on vision-language action, multimedia understanding, and computer vision.
\end{IEEEbiographynophoto}
\vspace{-35pt}

\begin{IEEEbiographynophoto}{Lianli Gao} is a professor with the School of Computer Science and Engineering, University of Electronic Science and Technology of China. She obtained her PhD degree in Information Technology from The University of Queensland (UQ), Australia, under the supervision of Prof. Jane Hunter and Prof. Michael Bruenig. Her research ranges from Semantic Web, Machine Learning, Deep Learning, Computer Vision (Images and Videos), NLP, Knowledge Reasoning, Knowledge and the related practical applications etc. Specifically, she is mainly focusing on integrating Natural Language for Visual Content Understanding. She has the winner of the IEEE Transactions on Multimedia 2020 Prize Paper Award, Best Student Paper Award in Australian Database Conference (2017, Australia), IEEE TCMC Rising Star Award 2020 and ALIBABA Academic Young Fellow. She is an Associate Editor of IEEE TMM.
\end{IEEEbiographynophoto}
\vspace{-35pt}

\begin{IEEEbiographynophoto}{Jingkuan Song} is a professor with the School of Computer
Science and Technology, Tongji University, China. 
He joined Columbia University as a Postdoc Research Scientist (2016-2017), and University of Trento as a Research Fellow (2014-2016). He obtained his PhD degree in 2014 from The University of Queensland (UQ), Australia. His research interest includes large-scale multimedia retrieval, LLMs and deep learning techniques. He was the winner of the Best Paper Award in ICPR (2016, Mexico), Best Student Paper Award in Australian Database Conference (2017, Australia), and Best Paper Honorable Mention Award (2017, Japan). He is an Associate Editor of IEEE TMM and ACM TOMM.
\end{IEEEbiographynophoto}
\vspace{-35pt}

\begin{IEEEbiographynophoto}{Nicu Sebe} is a professor with the Department of Information Engineering and
Computer Science, University of Trento, leading the research in the areas of multimedia information retrieval and human behavior understanding. He was the General CoChair of ACM Multimedia 2013 and 2022, and the Program Chair of ACM Multimedia 2007 and 2011, ECCV 2016, ICCV 2017 and ICPR 2020. He is a fellow of the International Association for Pattern Recognition (IAPr) and of the European Laboratory for Learning and Intelligent Systems (ELLIS). 
\end{IEEEbiographynophoto}
\vspace{-35pt}

\begin{IEEEbiographynophoto}{Heng Tao Shen} is a professor with the School of Computer Science and Technology, Tongji University, China. 
He obtained his BSc with 1st class Honours and PhD from Department of Computer Science, National University  of  Singapore in 2000 and 2004 respectively.
His current research interests include multimedia search, computer vision, artificial intelligence, and big data management. He has published 300+ peer-reviewed papers and received 7 best paper awards from international conferences, including the Best Paper Award from ACM Multimedia 2017 and Best Paper Award-Honourable Mention from ACM SIGIR 2017. He has served as General Co-chair for ACM Multimedia 2021 and TPC Co-Chair for ACM Multimedia 2015, and is an Associate Editor of ACM Trans. of Data Science (TDS), IEEE Trans. on Image Processing (TIP), IEEE Trans. on Multimedia (TMM), and IEEE Trans. on Knowledge and Data Engineering (TKDE). He is a Fellow of ACM/IEEE/OSA.
\end{IEEEbiographynophoto}
\vspace{-35pt}

\end{document}